\documentclass{article}

\PassOptionsToPackage{numbers, compress}{natbib}

\usepackage[preprint]{neurips_2026}

\usepackage[utf8]{inputenc} 
\usepackage[T1]{fontenc}    
\usepackage{hyperref}       
\usepackage{url}            
\usepackage{booktabs}       
\usepackage{amsfonts}       
\usepackage{nicefrac}       
\usepackage{microtype}      
\usepackage[table]{xcolor} 
\usepackage{amsmath}
\usepackage{amssymb}
\usepackage{graphicx}
\usepackage{subcaption}
\usepackage{multirow}
\usepackage{makecell}
\usepackage{tabularx}
\usepackage{algorithm}
\usepackage{algorithmic}
\usepackage{amsthm}
\usepackage{pifont}      
\usepackage{siunitx}
\usepackage{float}        
\usepackage{tikz}
\usetikzlibrary{positioning,arrows.meta,calc}
\usepackage{enumitem}    
\usepackage{longtable}   
\usepackage{wrapfig}     
\usepackage{needspace}   
\usepackage[skins,breakable]{tcolorbox}  
\usepackage{fancyvrb}    
\usepackage{fvextra}     
\usepackage[scaled=0.92]{inconsolata}     
\tcbset{
  cmpcard/.style={
    enhanced,
    colback=white,
    colframe=black!22,
    boxrule=0.4pt,
    arc=1.2mm,
    left=4pt, right=4pt, top=3pt, bottom=3pt,
    fonttitle=\footnotesize\bfseries,
    coltitle=black!75,
    colbacktitle=gray!8,
    titlerule=0pt,
    halign title=left,
    boxsep=2pt,
  },
  cmpcard inner/.style={
    cmpcard,
    colframe=black!15,
    colbacktitle=gray!4,
    boxrule=0.3pt,
  },
}
\hypersetup{
  pdfpagemode=UseOutlines,
  bookmarksopen=true,
  bookmarksopenlevel=2
}

\newcommand{\graydelta}[1]{{\color{gray}\scriptsize #1}}
\newcommand{\grayratio}[1]{{\color{gray}\scriptsize$\times$#1}}
\newcommand{\graydown}[1]{{\color{gray}\scriptsize$\downarrow$#1}}
\newtheorem{definition}{Definition}

\title{Draw2Think: Harnessing Geometry Reasoning through Constraint Engine Interaction}

\author{
Juncheng Hu\textsuperscript{1,2,3} \quad
Jiawei Du\textsuperscript{2,3} \quad
Xin Zhang\textsuperscript{2,3} \quad
Joey Tianyi Zhou\textsuperscript{2,3}\\
\textsuperscript{1}{\small National University of Singapore, Singapore}\\
\textsuperscript{2}{\small Centre for Frontier AI Research, Agency for Science, Technology and Research, Singapore}\\
\textsuperscript{3}{\small Institute of High Performance Computing, Agency for Science, Technology and Research, Singapore}\\
{\small\fontfamily{cmtt}\selectfont juncheng.hu@u.nus.edu}\hspace{1em}
{\small\fontfamily{cmtt}\selectfont{\{dujw, zhangx7, Joey\_Zhou\}@a-star.edu.sg}}\\
}

\begin{document}

\maketitle

\begin{abstract}
  \linespread{1.08}\selectfont
  Vision-language models solve geometry problems with rising accuracy, yet their intermediate states remain latent and unverifiable: a relation expressed in textual reasoning or drawing code carries no guarantee that a constraint-satisfying configuration realizes it. We observe that existing externalization methods based on rendered pixels or one-shot scripts fail to provide exact, per-action geometric guarantees. Enforcing geometric relations by algebraic definition closes this gap: the workspace becomes a constraint-checked evolving canvas. We present \textbf{Draw2Think}, a framework that recasts geometric reasoning from latent spatial inference into agentic interaction with the GeoGebra constraint engine. In a \textbf{Propose-Draw-Verify} loop, Draw2Think externalizes hypotheses onto an executable canvas, measures exact geometric quantities, and feeds structured observations back to the model, so subsequent reasoning proceeds from checked canvas state grounded by the shared workspace. This externalization makes two properties separately auditable: model-level \emph{Construction Fidelity} (whether the canvas realizes the intended configuration) and engine-level \emph{Measurement Faithfulness} (exact values and relations from canvas constraints). Across construction, outcome, and rendering evaluations, Draw2Think builds canvases that pass 95.9\% predicate-level and 84.0\% strict problem-level construction checks on \textit{GeoGoal}, improves outcome accuracy by up to 4.1\%/16.4\% on planar/solid benchmarks, and attains 68.2\%/90.5\% strict/relaxed rendering scores on \textit{GenExam-math}. Project page is available at \small\url{https://draw2think.github.io/}.

\end{abstract}

\section{Introduction}

Geometry problem solving pairs visual perception with multi-step deductive reasoning, making it a canonical stress test for vision-language model (VLM) capability~\cite{DL4GPS,GPSSurveyLMEra}. Benchmarks span planar, analytic, and solid regimes~\cite{Geometry3K,PGPS9K,MathVerse,SolidGeo,OlympiadBench}; progress is measured almost exclusively by end-to-end outcome accuracy, which conflates perception, construction, and deduction into a single signal.

Geometry is particularly well-structured for agentic interaction with an external workspace: actions are typed, state transitions are deterministic, intermediate quantities are exactly measurable, and invalid operations surface as errors rather than silent approximations. Yet most VLMs still solve it without such a workspace; they reason in latent tokens, where an $89^\circ$ angle is visually indistinguishable from $90^\circ$ despite encoding a different relation~\cite{MathVerse}. Outcome accuracy therefore hides an accountability gap: mid-reasoning claims such as angles, auxiliary lines, and derived lengths become premises for later steps without any engine confirming the underlying configuration.

Existing systems address fragments of this gap without closing it: formal provers certify proof steps but not the per-action construction states benchmark answers depend on~\cite{AlphaGeometry2,InternGeometry}; trained VLM solvers improve outcome accuracy while leaving intermediate states approximate~\cite{GeoVLMath,GeometryZero,SocraticGeo}; code- and render-externalization systems execute or re-perceive external artifacts but offer no engine-exact per-action check~\cite{GGBench,GeoSketch,GeoGramBench}. The missing piece is a \emph{verifiable intermediate state}: constructions that are executable, revisable, and independently checkable against geometric ground truth~\cite{GeoLaux}. When a model can directly measure an angle from a verified construction, it sidesteps fragile theorem-selection entirely.

\begin{figure}[H]
    \centering
    {\hypersetup{pdfnewwindow=true}%
    \href{https://draw2think.github.io/\#focus=paradigms}{%
        \includegraphics[width=\linewidth]{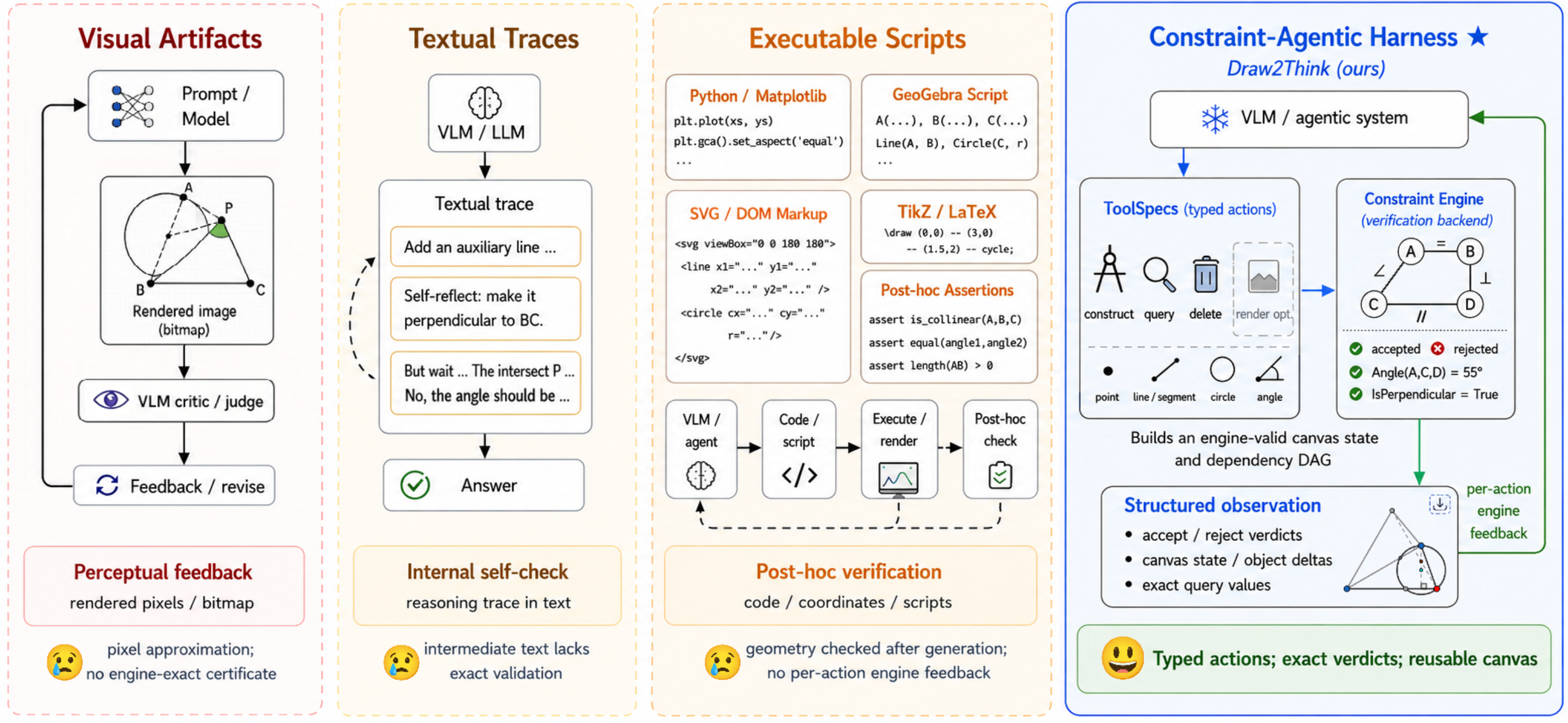}}}
    \vspace{-1\baselineskip}
    \caption{\textbf{Paradigms for externalizing intermediate geometry.} Prior routes externalize intermediate geometry as visual artifacts, textual traces, or executable scripts. \textbf{Draw2Think} adds a constraint-agentic harness: a frozen VLM selects typed ToolSpecs, the GeoGebra engine updates an engine-valid canvas state, and structured observations return after each action. The distinction is less about externalizing state than about when verification enters the loop.}
    \label{fig:overview}
\end{figure}
\vspace{-0.5ex}
The root cause is the timing and authority of feedback: perception, self-check, and post-hoc execution expose artifacts without certifying construction-time validity. \textbf{Draw2Think} takes a fourth route: a \emph{constraint-agentic harness} that wraps a frozen VLM around a dynamic-geometry engine through typed ToolSpecs. In dynamic geometry systems such as GeoGebra~\cite{GeoGebra}, construction commands enforce geometric relationships algebraically rather than by coordinate approximation, making the model's geometric priors executable and each accepted action checkable. The VLM proposes typed actions, the engine executes or rejects them on a shared canvas, and structured observations ground the next step; trajectory-level strategy remains with the model.

We evaluate this design \emph{training-free}, instantiating Draw2Think on Gemini 3 Flash Preview~\cite{Gemini3Flash} throughout: frontier VLMs already carry geometric priors strong enough to benefit from an external workspace, while their intermediate claims remain unreliable without grounded verification. Controlled comparisons on eight planar and two solid benchmarks show \emph{selective} outcome gains up to 4.1\% on planar and 16.4\% on solid: gains are largest on visually grounded benchmarks (\textit{GeoSketch}~\cite{GeoSketch}, \textit{MathVerse}~\cite{MathVerse}) and become negative where the baseline is already saturated (\textit{Geo3K}~\cite{Geometry3K}, \textit{PGPS9K}~\cite{PGPS9K}); ablations (\S5) isolate which components drive these gains.

Three contributions thread through the paper:
\begin{itemize}[nosep,leftmargin=1.2em]
    \item \textbf{Formulation.} We recast geometric reasoning as an inference-time closed loop around a constraint engine: one substrate serves as executor, verifier, and generator while the frozen VLM perceives, reasons, and dispatches typed tools over a verifiable, revisable canvas.
    \item \textbf{Evidence.} We validate the approach along three axes: outcome gains concentrate where visual measurement or spatial construction is the bottleneck (Table~\ref{tab:main-results}); constructed canvases pass $95.9\%$ predicate-level and $84.0\%$ strict problem-level checks on GeoGoal~\cite{GeoGoal}; and the same framework transfers to solid geometry and GenExam-math rendering without changing the loop.
    \item \textbf{Mechanism.} We localize when grounding changes behavior, through failure taxonomy, answer provenance, and graduated ablations (\S\ref{sec:analysis}).
\end{itemize}

\section{Related Work}
\label{sec:related-work}

Geometry problem solving requires jointly perceiving spatial relations from diagrams and performing multi-step deductive reasoning over them~\cite{DL4GPS,GPSSurveyLMEra}. A defining difficulty is that intermediate states such as auxiliary lines, derived measurements, and construction hypotheses are reused as premises by later steps, so an unverified local hallucination becomes a hidden global premise. Formal theorem proving~\cite{AlphaGeometry2,InternGeometry,Newclid} addresses this failure mode by certifying entire proof chains against axioms. Its guarantees are proof-targeted rather than measurement-targeted: the certified objects are derivation steps rather than the per-action numerical states that downstream readouts depend on. For solving and rendering, the complementary guarantee is measurement faithfulness at query time; this motivates inference-time verification with per-step engine feedback.

\paragraph{Textual and visual reasoning traces.}
A contrasting line records intermediate state as textual or visual traces, leaving geometric validity uncertified. Recent diagnostics localize the fragility: VLMs show weak fine-grained grounding on geometry diagrams~\cite{MathBlind}, and applying identified geometric principles within the visual context remains a dominant bottleneck~\cite{GeoSense}. Textual auxiliary-line systems expose the same validation gap: GeoVLMath~\cite{GeoVLMath} finds incorrect auxiliary lines worse than omitting them, and GeometryZero~\cite{GeometryZero} shows that unconditional auxiliary construction can be counterproductive; Socratic-Geo~\cite{SocraticGeo} instead validates generated training triplets, leaving inference-time object validity outside the loop. Visual-trace systems fare similarly: MathCanvas~\cite{MathCanvas} and ThinkMorph~\cite{ThinkMorph} emit pixel intermediates for mode-switched chain-of-thought (CoT), leaving rendered geometry without a constraint-level certificate. Across this line, intermediate states may guide later reasoning without receiving certification from an exact engine query.

\paragraph{Executable scripts and rendered feedback.}
Externalized artifacts make state visible, while per-action geometric exactness remains unguaranteed. \emph{Coordinate-first} systems commit numerical $(x,y)$ values in code: GeoCode~\cite{GeoCode} generates drawing programs, GGBench~\cite{GGBench} reports $79\%$ executability against $57\%$ geometric correctness, and GeoSDF~\cite{GeoSDF} exposes a final diagram after signed-distance optimization. Rendered-feedback loops return the artifact to a visual channel: Visual Sketchpad~\cite{VisualSketchpad} sketches on matplotlib renders, and Canvas-CoT~\cite{CanvasCoT} pairs mutable document object model (DOM) state with a VLM critic against a \emph{reference image}. Across these systems, verification remains code-level, visual, or post-hoc; what is missing is an engine source of exact intermediate measurements.

\paragraph{GeoGebra before inference-time operation.}
Recent geometry work uses GeoGebra as infrastructure for building resources, executing scripts, evaluating outputs, or shaping rewards. We-Math~2.0~\cite{WeMath2} places GeoGebra in its MathBook resource-building pipeline, and Newclid~\cite{Newclid} routes a graphical interface into a symbolic backend. GGBench~\cite{GGBench} freezes GeoGebra code as an evaluation oracle, while Faire~\cite{Faire} executes generated GeoGebra programs and model-synthesized boolean assertions as post-generation verification for reinforcement-learning (RL) reward. What remains underexplored is inference-time operation: the model issues actions one at a time, receives engine verdicts immediately, and conditions later calls on the verified canvas state. Draw2Think wraps the engine in a typed constraint-agentic harness~\cite{Toolformer,ToolStar,AHE}, with the embedded Giac computer algebra system (CAS) serving as a per-action verifier inside the Propose-Draw-Verify loop.

\section{Method}

\subsection{Problem Formulation: Construction as State-Space Search}
\label{sec:state-space}

We consider a geometry problem $\mathcal{P} = (I, T, Q, C)$ with diagram $I$, text $T$, question $Q$, and optional candidate set $C$. Direct VLM inference maps $\mathcal{P}$ to an answer in a single forward pass. Draw2Think instead searches for a \emph{canvas state} $\mathcal{S}_K$ from which the answer is determined by a bounded readout $h(Q,\mathcal{S}_K)$.

We formalize the canvas as a deterministic environment $\mathcal{E} = (\mathcal{S}, \mathcal{A}, \mathcal{T}, \mathcal{O})$:

\begin{definition}[Geometric Construction Environment]
\label{def:env}
A canvas environment is a tuple $\mathcal{E} = (\mathcal{S}, \mathcal{A}, \mathcal{T}, \mathcal{O})$ where
$\mathcal{S}$ is a \textbf{state space} of typed geometric objects whose dependencies induce a Directed Acyclic Graph (DAG);
$\mathcal{A}$ is an \textbf{action space} of 92 typed ToolSpecs (\S\ref{sec:toolspec});
$\mathcal{O}: \mathcal{S} \times \mathcal{A} \to \mathcal{Y}$ is the \textbf{observation} function (where $\mathcal{Y}$ comprises structured responses: object deltas, exact scalar values, or error messages); and $\mathcal{T}: \mathcal{S} \times \mathcal{A} \to \mathcal{S}'$ is a \textbf{deterministic transition}.
Construction actions extend $\mathcal{S}$ with new objects; query actions return exact values ($v \in \mathbb{R}$ or $b \in \{\textsc{true}, \textsc{false}\}$) without modifying $\mathcal{S}$; delete actions remove an object and all its dependents. Invalid actions are rejected with an explicit error whenever the engine detects the failure.
\end{definition}

\noindent We instantiate $\mathcal{E}$ with GeoGebra~\cite{GeoGebra}, a dynamic geometry system in which objects are defined by construction commands rather than free-form coordinates. The key modeling choice is therefore constraint-first: the model declares \emph{which} geometric relation should hold, and the engine commits an object satisfying that relation by construction. The algebraic guarantee behind this execution model is detailed in \S\ref{sec:toolspec}.

Solving is then a state-reachability problem. At the \emph{action} level, the model constructs a grounded state through a trajectory $\tau = (a_1, \mathcal{O}_1, \ldots, a_K, \mathcal{O}_K)$ of interleaved actions and observations:
\begin{equation}
\label{eq:state-reachability}
    \mathcal{S}_k = \mathcal{T}(\mathcal{S}_{k-1}, a_k), \quad
    \mathcal{O}_k = \mathcal{O}(\mathcal{S}_k, a_k), \quad
    \mathcal{S}_0 = \varnothing,
    \qquad
    \hat{y} = h(Q, \mathcal{S}_K),
\end{equation}
where $k$ indexes individual tool calls. Algorithm~\ref{alg:pdv} groups these calls into \emph{turns}: at each turn $t$, the model may emit multiple actions $\{a_t^{(i)}\}$ before receiving the next observation; Eq.~\ref{eq:state-reachability} applies sequentially within the turn. For numerical targets, $h=f_{\mathrm{query}}$ reads an engine measurement; for multiple-choice targets, $h=g(\mathrm{text}_t,C,\mathcal{S}_K)$ combines the final text with engine-grounded values. The same final state also supports rendering via $\hat{I}=\rho(\mathcal{S}_K;\eta)$, with task differences captured by the active tool subset and readout.

Eq.~\ref{eq:state-reachability} leaves three burdens to the model: choosing valid high-arity actions, preserving DAG consistency, and deciding when query-supported readout is sufficient. Draw2Think handles these burdens with a training-free controller: a typed action interface with auditable guarantees (\S\ref{sec:toolspec}), composed by the Propose-Draw-Verify loop (Algorithm~\ref{alg:pdv}).



\subsection{Action Interface and Auditable Guarantees}
\label{sec:toolspec}

\paragraph{Action space.}
The predefined ToolSpecs group forms a typed interface to the canvas, following the structured function-calling view of tool descriptions and typed parameter bindings~\cite{TAFC}: each specification combines a natural-language description (selection cue), typed parameter signatures (invocation contract), and preconditions jointly enforced by schema and backend. Construction tools extend $\mathcal{S}$ with algebraically-defined objects; query tools return exact scalars without modifying $\mathcal{S}$; delete tools prune subtrees via cascading removal (Definition~\ref{def:env}). Full catalog and representative descriptions in Appendix~\ref{app:toolspec}. For 3D benchmarks (\S\ref{sec:3d-rendering}), this set is extended with 21 solid-geometry tools (Table~\ref{tab:toolspec-catalog-3d}).

At inference time, each action call decomposes as $a_k=(u_k,\varphi_k)$, where $u_k$ is the ToolSpec choice and $\varphi_k$ assigns its typed parameters. This yields the causal factorization
{\fontsize{9pt}{10pt}\selectfont
\begin{equation}
\label{eq:action-decomp}
P(a_k \mid \mathcal{P}, \mathcal{A}, \mathcal{H}_{k-1})
=
P(u_k,\varphi_k \mid \mathcal{P}, \mathcal{A}, \mathcal{H}_{k-1})
=
\underbrace{P(u_k \mid \mathcal{P}, \mathcal{A}, \mathcal{H}_{k-1})}_{\text{tool selection}}
\cdot
\underbrace{P(\varphi_k \mid u_k, \mathcal{P}, \mathcal{A}, \mathcal{H}_{k-1})}_{\text{parameter generation}} .
\end{equation}
}
Here $\mathcal{P}$ and $\mathcal{A}$ are fixed for the problem instance, while $\mathcal{H}_{k-1}=((a_j,\mathcal{O}_j))_{j<k}$ is the evolving verified trace history before action $k$. This decomposition makes the ToolSpec layers operational: natural-language descriptions primarily shape tool selection, while typed signatures and preconditions constrain parameter generation. The description ablation in \S\ref{sec:ablation} preserves the second channel while removing the first, isolating selection-layer guidance.

\begin{table}[t]
\centering
\caption{Action types and their state semantics. The 55 construction tools include \texttt{delete\_object}, which removes the target together with its dependents via cascading delete (Definition~\ref{def:env}).}
\label{tab:action-types}
\small
\begin{tabular}{@{}llcl@{}}
\toprule
\textbf{Type} & \textbf{State Transition} & \textbf{N} & \textbf{Example} \\
\midrule
Construction (\texttt{add\_*}) & $\mathcal{S}' = \mathcal{S} \cup \{o_{\mathrm{new}}\}$ & 55 & \texttt{add\_perpendicular\_line} \\
Query (\texttt{query\_*}) & $f_{\mathrm{query}}(\mathcal{S}, q) \to v \in \mathbb{R}$ & 24 & \texttt{query\_angle} \\
Render (\texttt{render\_*}) & visual style only; $\mathcal{S}$ unchanged & 13 & \texttt{render\_set\_color} \\
\bottomrule
\end{tabular}
\end{table}

\paragraph{Construction Fidelity (model-level, empirical).}
Construction Fidelity is a model-level property: the final canvas satisfies the externally specified geometric predicates of the problem, regardless of whether the final answer is correct. This is not guaranteed by the engine and must be audited empirically; we evaluate it in \S\ref{sec:process-fidelity}. The distinction matters for training-free operation: ToolSpecs expose textbook operations such as ``perpendicular,'' ``bisector,'' and ``tangent'' as verified primitives, but the model must still choose the right primitives and parameters. Removing natural-language ToolSpec cues while preserving typed signatures raises LLM-bypass $4.6\times$ (\S\ref{sec:ablation}), confirming that selection-layer guidance is part of the method rather than cosmetic documentation.

\paragraph{Measurement Faithfulness (engine-level, algebraic).}
Measurement Faithfulness is the complementary engine-level property: because accepted objects are stored as algebraic relations and resolved by GeoGebra's embedded Giac CAS~\cite{GiacGeoGebra} using Gr\"obner-basis elimination~\cite{GeoGebraATP}, each query $f_{\mathrm{query}}(\mathcal{S}, q)$ returns the deterministic value $v^*(\mathcal{S}, q)$ implied by the committed canvas. The guarantee is \emph{semantic-free}: it says what follows from $\mathcal{S}$, not whether $\mathcal{S}$ matches the problem's intended configuration. Correct constructions therefore yield exact readouts; incorrect constructions yield exact measurements of the wrong canvas, exposing discrepancies for replanning.

In practice, degenerate inputs account for a small fraction of all failed calls (${\sim}$5\%, Appendix~\ref{app:failure-taxonomy}) and may produce undefined objects without an error message, which propagate as downstream reference failures.


\subsection{Propose-Draw-Verify Loop}

\begin{wrapfigure}[14]{r}{0.45\linewidth}
\vspace{-2.8\baselineskip}
\begin{minipage}{\linewidth}
\refstepcounter{algorithm}
\label{alg:pdv}
\hrule height 0.8pt
\vspace{2pt}
\noindent\textbf{Algorithm \thealgorithm}\ Draw2Think Inference
\par\vspace{2pt}
\hrule
\vspace{3pt}
\footnotesize
\begin{algorithmic}[1]
\REQUIRE $\mathcal{P}, \mathcal{A}, \mathcal{E}$
\STATE $\mathcal{S}_0 \!\leftarrow\! \varnothing$; $\mathcal{H} \!\leftarrow\! \langle\,\rangle$
\FOR{$t = 1, 2, \ldots$}
    \STATE $\mathrm{text}_t, \{a_t^{(i)}\} \!\leftarrow\! \pi_\theta(\mathcal{P}, \mathcal{A}, \mathcal{H})$ \hfill\textcolor{gray}{[Propose]}
    \IF{$\textsc{Answer} \!\in\! \mathrm{text}_t$}
        \RETURN $h(\mathrm{text}_t, \mathcal{S}_{t-1})$
    \ENDIF
    \IF{$\{a_t^{(i)}\} \!=\! \varnothing$}
        \RETURN $\mathrm{text}_t$
    \ENDIF
    \STATE $\mathcal{S} \!\leftarrow\! \mathcal{S}_{t-1}$
    \FOR{$i = 1$ \textbf{to} $|a_t|$}
        \STATE $\mathcal{S}, \mathcal{O}_t^{(i)} \!\leftarrow\! \mathcal{E}.\textsc{Step}(\mathcal{S}, a_t^{(i)})$ \hfill\textcolor{gray}{[Draw]}
    \ENDFOR
    \STATE $\mathcal{S}_t \!\leftarrow\! \mathcal{S}$;\ \ $\mathcal{H} \!\leftarrow\! \mathcal{H} \cdot \big\langle (a_t^{(i)}, \mathcal{O}_t^{(i)}) \big\rangle_{i=1}^{|a_t|}$ \hfill\textcolor{gray}{[Verify]}
\ENDFOR
\end{algorithmic}
\vspace{2pt}
\hrule height 0.8pt
\end{minipage}
\vspace{-1.1\baselineskip}
\end{wrapfigure}

The Propose-Draw-Verify (PDV) loop (Algorithm~\ref{alg:pdv}) externalizes geometric hypotheses onto the constraint engine and feeds back structured state after each action, so that each reasoning step is grounded by an engine-verified observation instead of a purely autoregressive continuation. The loop exploits a fundamental asymmetry: proposal is a high-entropy decision over 92 typed tools and their parameters, whereas verification reduces to a deterministic algebraic check. This asymmetry is what makes the closed loop productive: judgment is delegated to the external oracle while the model remains responsible for proposal.

Two properties of the observation distinguish this loop from visual-feedback alternatives:

\textbf{Structured canvas memory}: each observation $\mathcal{O}_t$ is engine-grounded; construction calls expose object-level deltas, and query calls return exact values (Measurement Faithfulness, \S\ref{sec:toolspec}). In contrast to visual-feedback loops that re-perceive rendered output, these observations are algebraically exact and directly conditioned upon without a second round of visual interpretation that would compound generation with perception error (Appendix~\ref{app:agentic-landscape}). The model conditions on the verified history $a_{k+1} \sim \pi_\theta(\cdot \mid \mathcal{P}, \mathcal{A}, \mathcal{H}_k)$ where $\mathcal{H}_k = \big((a_j, \mathcal{O}_j)\big)_{j=1}^{k}$ is the ordered trace of actions and observations and every $\mathcal{O}_j$ is engine-supplied rather than self-attested.

\textbf{Revisable state}: the engine preserves a consistent partial canvas under failure. Invalid actions are rejected explicitly, failed calls roll back locally while valid objects remain available, and the model may additionally invoke cascading delete when it chooses to retract a committed branch (\S\ref{sec:recovery}). This recovery regime is unavailable to open-loop code generation~\cite{GGBench}.


\section{Experiments}
\label{sec:experiments}

\subsection{Setup}
\label{sec:setup}

We evaluate along three axes: \emph{construction fidelity} on GeoGoal~\cite{GeoGoal} under an engine-exact predicate verifier (\S\ref{sec:process-fidelity}); \emph{outcome accuracy} on eight planar and two solid benchmarks under ground-truth (GT) match (\S\ref{sec:main-results}, \S\ref{sec:cross-model}, \S\ref{sec:3d-rendering}); and \emph{output utility} on GenExam-math~\cite{GenExam} under VLM-judge scoring (\S\ref{sec:3d-rendering}).

\begin{figure}[H]
  \centering
  \includegraphics[width=\linewidth]{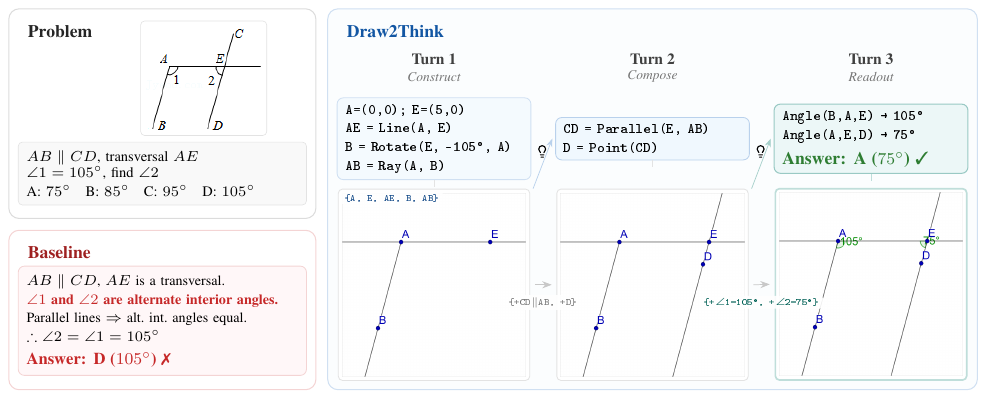}
  \caption{Mechanism comparison on MathVista/290. Baseline applies alternate-interior-angles and concludes $105^\circ$; Draw2Think runs \emph{Construct} $\to$ \emph{Compose} $\to$ \emph{Readout}, with canvas state accumulating across turns, and the readout returns $\angle 2 = 75^\circ$.}
  \label{fig:qualitative}
\end{figure}

We compare two inference modes of Gemini-3-Flash~\cite{Gemini3Flash}. \textbf{BL} is a direct vision-language \textbf{\underline{B}ase\underline{L}ine} that answers in one pass from internal state. \textbf{CT} is our closed-loop \textbf{\underline{C}ons\underline{T}ruction} mode: the model issues a ToolSpec call, the engine executes it on the shared canvas, and the returned structured feedback conditions the next model step. We define one such model--engine round trip as a turn. Within each benchmark, BL and CT share temperature~0 and the same thinking level (\texttt{medium}, \textit{OlympiadBench} and \textit{SolidGeo-hard} at \texttt{high}); CT is capped at 30 turns and 120 seconds per turn. All results are single-attempt Pass@1.

\subsection{Selective Outcome Gains under a Fixed Backbone}
\label{sec:main-results}

Table~\ref{tab:main-results} summarizes the controlled same-model comparison. CT gains are \emph{selective} and track BL internal-simulation burden: the CT/BL token ratio shifts from $>1$ in low-BL-thinking regimes to $<1$ in high-BL-thinking regimes. Gains are largest on visually grounded tasks (\textit{MathVerse}, \textit{GeoSketch}) and on the officially hard 3D split (\textit{SolidGeo-hard}), where exact external measurement replaces expensive latent simulation. They shrink on already-strong textbook benchmarks and reverse on the saturated cluster (\textit{Geo3K}, \textit{PGPS9K}), where forced construction adds overhead without resolving a real ambiguity. Within \textit{GeoLaux}, gains increase with annotated reference-solution length.
Engine grounding matters most when the bottleneck is deriving exact quantities from visual input, whether through perceptual estimation or theorem selection, and becomes overhead when the model already has a cheap internal route.

\begin{table}[H]
  \vspace{-1.2\baselineskip}
  \centering
  \caption{Same-model Pass@1 comparison. Rows are ordered by BL thinking tokens, exposing how CT gains depend on the baseline's observed internal-simulation burden. Save/Break are paired outcome transitions; the right inset profiles the BL traces behind planar Save/Break cases (Cross-system in Appendix~\ref{app:cross-system}; Wall-time analysis in ~\ref{app:walltime-profile}; Full Save/Break in Appendix~\ref{app:outcome-transition}.)} 
  \label{tab:main-results}
  \scriptsize
  \begin{minipage}[t]{0.75\linewidth}
  \centering
  \setlength\tabcolsep{1.7pt}
  \begin{tabular}{@{}lr @{\hspace{0.6em}} r@{$\,\rightarrow\,$}c@{}l @{\hspace{0.6em}} ccc @{\hspace{1.0em}} r@{\hspace{0.6em}}c@{}l@{}}
  \toprule
  & & \multicolumn{6}{c}{\textbf{Outcome accuracy (Pass@1)}} & \multicolumn{3}{c}{\textbf{Think tokens / prob.}} \\
  \cmidrule(lr){3-8} \cmidrule(lr){9-11}
  \textbf{Benchmark} & \textbf{N} & \hspace{0.8em}\textbf{BL} & \textbf{CT} & & \makecell{\textbf{Save}\\{\tiny BL\ding{55}CT\checkmark}} & \makecell{\textbf{Break}\\{\tiny BL\checkmark CT\ding{55}}} & \makecell{\textbf{Win}\\{\scriptsize S/B}} & \hspace{2em}\textbf{BL} & \textbf{CT} & \\
  \midrule
  GeoQA~\cite{GeoQA} / UniGeo~\cite{UniGeo} & 754 & 93.6 & \textbf{96.9} & \graydelta{+3.3} & 40 & 15 & \textbf{2.7} & 808 & 1\,785 & \grayratio{2.21} \\
  PGPS9K~\cite{PGPS9K} & 1000 & \textbf{94.5} & 90.5 & \graydelta{-4.0} & 13 & 53 & 0.2 & 1\,043 & 1\,566 & \grayratio{1.50} \\
  MathVista~\cite{MathVista} GPS & 208 & 97.1 & \textbf{97.6} & \graydelta{+0.5} & 4 & 3 & \textbf{1.3} & 1\,174 & 1\,629 & \grayratio{1.39} \\
  Geo3K~\cite{Geometry3K} & 601 & 98.2 & 95.3 & \graydelta{-2.9} & 6 & 23 & 0.3 & 1\,485 & 1\,878 & \grayratio{1.26} \\
  GeoLaux~\cite{GeoLaux} & 221 & 93.2 & \textbf{94.1} & \graydelta{+0.9} & 11 & 9 & \textbf{1.2} & 2\,301 & 2\,872 & \grayratio{1.25} \\
  MathVerse~\cite{MathVerse} Plane & 510 & 87.5 & \textbf{91.0} & \graydelta{+3.5} & 29 & 11 & \textbf{2.6} & 2\,637 & 2\,810 & \grayratio{1.07} \\
  GeoSketch~\cite{GeoSketch} & 390 & 81.8 & \textbf{85.9} & \graydelta{+4.1} & 26 & 10 & \textbf{2.6} & 5\,498 & \textbf{3\,545} & \grayratio{0.64} \\
  OlympiadBench~\cite{OlympiadBench} & 112 & 89.3 & 89.3 & \graydelta{+0.0} & 4 & 4 & \textbf{1.0} & 5\,884 & \textbf{5\,063} & \grayratio{0.86} \\
  \midrule
  MathVerse-solid & 119 & 82.4 & \textbf{88.2} & \graydelta{+5.8} & 10 & 3 & \textbf{3.3} & 2\,121 & \textbf{1\,897} & \grayratio{0.89} \\
  SolidGeo-hard~\cite{SolidGeo} (Lv.3) & 177 & 59.9 & \textbf{76.3} & \graydelta{\textbf{+16.4}} & 36 & 7 & \textbf{5.1} & 9\,000 & \textbf{8\,114} & \grayratio{0.90} \\
  \bottomrule
  \end{tabular}
  \end{minipage}
  \hfill
  \begin{minipage}[p]{0.23\linewidth}
  \vspace{1.7ex}
  \centering
  
  \noindent\textit{Save/Break reasoning effort comparison: higher $\rightarrow$ harder}
  \setlength\tabcolsep{2pt}
  \begin{tabular}{@{}lrr@{}}
  \toprule
  \makecell[l]{\textbf{BL think trace}} & \textbf{Save} & \textbf{Break} \\
  \midrule
  Empty BL Log & 51\% & 0\% \\
  BL Median & 5206.5 & 1503.0 \\
  BL Mean & 6004.7 & 3166.9 \\
  \bottomrule
  \end{tabular}

  \vspace{2.8ex}

  \noindent{\itshape Wall time by BL-thinking regime.}
  \setlength\tabcolsep{1.5pt}
  \begin{tabular}{@{}ccrr@{}}
  \toprule
  \makecell[l]{\textbf{Thinking}\\\textbf{burden group}} & \textbf{BL} & \makecell{\textbf{CT}\\\textbf{tot.}} & \makecell{\textbf{CT}\\\textbf{/turn}} \\
  \midrule
  Low BL-think & 10.2 & 21.3 & 9.2 \\
  High BL-think & 35.7 & 35.9 & 12.6 \\
  \bottomrule
  \end{tabular}
  \end{minipage}
\end{table}
\vspace{-1.8\baselineskip}

\subsection{Canvas-Level Construction Fidelity}
\label{sec:process-fidelity}

\begin{figure}[H]
\centering
\begin{minipage}[c]{0.36\linewidth}
\centering
\captionof{table}{Construction fidelity on GeoGoal ($N\!=\!256$, 13{,}254 predicates). \textbf{SR}: predicate-level pass rate; \textbf{SC}: strict problem-level predicate pass rate; \textbf{CR}: non-empty canvases rate.}
\label{tab:process-fidelity}
\footnotesize\setlength\tabcolsep{4pt}
\begin{tabular}{@{}l ccc@{}}
\toprule
\textbf{Metric}      & \multicolumn{3}{c}{\textbf{Rate\%}} \\
\midrule
\textbf{SR overall}  & \multicolumn{3}{c}{\textbf{95.94}} \\
\hspace{0.4em}\itshape by tier & \makecell{\itshape premise\\96.85} & \makecell{\itshape num-check\\97.33} & \makecell{\itshape derived\\94.78} \\
\midrule
\textbf{SC}          & \multicolumn{3}{c}{\textbf{84.0}} \\
\textbf{CR}          & \multicolumn{3}{c}{100.0} \\
\bottomrule
\end{tabular}
\end{minipage}%
\hfill
\begin{subfigure}[c]{0.24\linewidth}
\centering
\includegraphics[width=\linewidth]{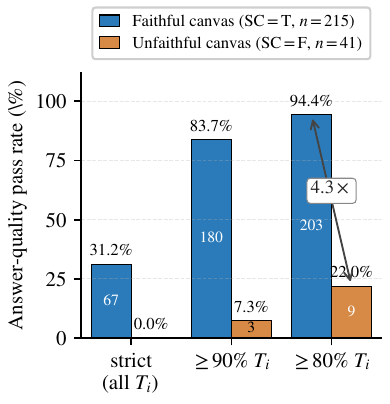}
\caption{Fidelity-conditional answer quality.}
\label{fig:fidelity-bars}
\end{subfigure}%
\hfill
\begin{subfigure}[c]{0.35\linewidth}
\centering
\includegraphics[width=\linewidth]{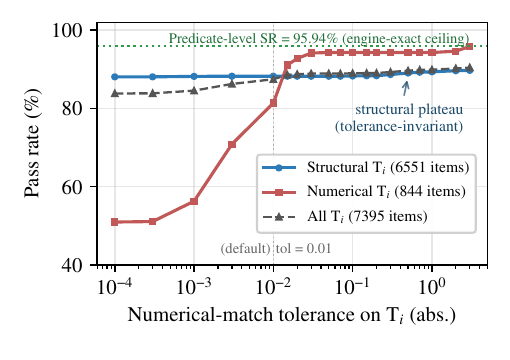}
\caption{Tolerance sweep on $T_i$ match.}
\label{fig:tol-sweep}
\end{subfigure}
\caption{\textbf{(a)} Faithful canvases ($n{=}215$) dominate unfaithful ($n{=}41$) at three $T_i$-match thresholds. \textbf{(b)} Structural $T_i$ plateau at $\sim\!88\%$ (tolerance-invariant); numerical $T_i$ climb to the engine-exact SR (dotted) as tolerance approaches $\sim\!0.1\%$ rel.}
\label{fig:fidelity-precision}
\end{figure}

Accuracy validates the \emph{outcome} axis; \emph{Construction Fidelity} (\S\ref{sec:toolspec}) requires a separate audit. We therefore audit each final canvas on GeoGoal~\cite{GeoGoal} against ground-truth predicates from TrustGeoGen's~\cite{TrustGeoGen} formal engine. The audit is \emph{offline and deterministic}: Newclid~\cite{Newclid} checks each predicate only from named-point coordinates, without access to the model's reasoning trace or final answer (13{,}254 checks; Table~\ref{tab:process-fidelity}). SR remains stable across premise, numerical-check, and derived predicates, indicating that typed tool composition does not erode with derivation depth.

Figure~\ref{fig:fidelity-precision} decomposes the residual. The gap between faithful and unfaithful canvases widens as the $T_i$-match threshold tightens because structural errors propagate globally across derived readouts, whereas precision limits affect only the specific numerical queries they involve. The two failure modes also split cleanly by expression type: angle $T_i$ concentrate structural errors (97\%), length $T_i$ concentrate precision errors (100\%; Appendix~\ref{app:geogoal-failure-decomp}). Engine-exact verification separates them: an LLM-judge would collapse both failure modes into a single aggregate signal, while per-predicate checking splits them for independent mitigation.

\subsection{Backbone Portability}
\label{sec:cross-model}

\begin{wraptable}{r}{0.47\linewidth}
\vspace{-1.5\baselineskip}
\centering
\caption{Backbone portability on \textit{OlympiadBench} (N = 112; BL correct = 100 / 112 for all three models; CT correct = 100 / 103 / 106).}
\label{tab:cross-model}
\vspace{-0.4\baselineskip}
\footnotesize\setlength\tabcolsep{4pt}
\begin{tabular}{@{}l ccc@{}}
\toprule
\textbf{Model} & \textbf{BL} & \multicolumn{1}{c}{\textbf{CT}} & \textbf{Cost} (US\$) \\
\midrule
Gemini 3 Flash@high         & 89.3 & 89.3~\graydelta{+0.0}          & \$4.8 \\
Claude Sonnet 4.6@high      & 89.3 & 92.0~\graydelta{$\uparrow$2.7} & \$26.0 \\
GPT-5.4@high                & 89.3 & \textbf{94.6}~\graydelta{$\uparrow$5.4} & \$40.5 \\
\bottomrule
\end{tabular}
\vspace{-1\baselineskip}
\end{wraptable}

The main study measures the engine-interaction effect under a fixed backbone (Gemini-3-Flash). Table~\ref{tab:cross-model} probes portability across providers under the same prompt and ToolSpec harness: Claude / GPT deltas are positive on this N = 112 subset, supporting interface portability without establishing broad cross-model generalization (Appendix~\ref{app:cross-model}; Appendix~\ref{app:construction-gallery}).

\subsection{Formulation Transfer to 3D and Rendering}
\label{sec:3d-rendering}

\noindent\textit{Solid geometry.} The PDV loop transfers to 3D by adding 21 solid-geometry ToolSpecs, without changing the prompt structure or interaction protocol. Gains hold on \textit{MathVerse-solid} (+5.8\%) and \textit{SolidGeo-hard} Level 3 (+16.4\%), with thinking tokens decreasing in both cases (Table~\ref{tab:main-results}).

\begin{wraptable}{r}{0.53\linewidth}
\vspace{-1.2\baselineskip}
\centering
\caption{\textit{GenExam}-math: top closed-source systems (full leaderboard \href{https://github.com/OpenGVLab/GenExam\#-leaderboard}{\underline{online}} and in Appendix~\ref{app:genexam-detail}).}
\label{tab:genexam-utility}
\vspace{-0.4\baselineskip}
\footnotesize\setlength\tabcolsep{3pt}
\begin{tabular}{@{}l l rr@{}}
\toprule
\textbf{Method} & \textbf{Base / date} & \textbf{Strict} & \textbf{Relaxed} \\
\midrule
\textbf{Draw2Think} & Gemini 3 Flash {\fontsize{4.6pt}{3pt}\selectfont Dec.2025} & \textbf{68.2} & \textbf{90.5} \\
Nano Banana 2 & Gemini 3.1 Flash Image {\fontsize{4.6pt}{3pt}\selectfont Feb.2026} & 56.3 & 87.8 \\
GPT-Image-2 & OpenAI native T2I {\fontsize{4.6pt}{3pt}\selectfont Apr.2026} & 50.3 & 85.2 \\
\bottomrule
\end{tabular}
\vspace{-1.5\baselineskip}
\end{wraptable}

\noindent\textit{Diagram rendering.} On \textit{GenExam-math}~\cite{GenExam}, where leading closed-source text-to-image (T2I) systems struggle, Draw2Think attains 68.2/90.5 strict/relaxed accuracy, improving strict score by 11.9 points over the leading T2I generation system (Appendix~\ref{app:genexam-detail}). The gain traces to the externalized constraint engine: stated geometric relations are enforced algebraically instead of approximated in pixels.

\section{Analysis: Mechanisms of Grounded Interaction}
\label{sec:analysis}

The experiments isolate three mechanisms: grounding reshapes the model's reasoning trace under specific bottlenecks (\S\ref{sec:selectivity}), verified readout offers an explicit extraction channel for engine-grounded quantities (\S\ref{sec:ablation}), and the remaining failures are model-level errors in perception and strategy (\S\ref{sec:recovery}).

\subsection{Selective Effects of Grounding}
\label{sec:selectivity}
CT effects are bottleneck-dependent: gains concentrate when the missing resource is executable state, including vision-heavy measurement, auxiliary construction, and hard solid geometry, and can reverse on saturated tasks where strong internal priors already supply a low-cost route (Table~\ref{tab:main-results}). That bottleneck profile is consistent with cognitive offloading~\cite{ActingLess, SMART}: tool calls replace fragile spatial simulation when simulation is expensive and add overhead when external state is unnecessary.

On \textit{GeoSketch} and \textit{OlympiadBench}, CT cuts thinking tokens by 36\% and 14\% respectively while raising accuracy (Table~\ref{tab:main-results}). In these regimes, the engine substitutes for internal simulation precisely where simulation is most expensive. Figure~\ref{fig:think-difficulty} (Appendix~\ref{app:process}) visualizes this transition across the difficulty gradient. Removing query tools reverses this effect, reducing accuracy despite shorter context (\S\ref{sec:ablation}). Per-problem Save/Break transitions reveal the same asymmetry: CT rescues many more BL failures than it breaks on vision-heavy or construction-heavy benchmarks, while regressions concentrate where BL already has a low-cost internal route (Table~\ref{tab:main-results}; Appendix~\ref{app:outcome-transition}).

\subsection{Readout Mechanism: Source Distribution and Ablation}
\label{sec:ablation}

\begin{wraptable}{r}{0.47\linewidth}
\vspace{-2\baselineskip}
\centering
\caption{Graduated ablations on matched subsets. Query ablation (N=379) removes measurement (meas.) tools or all query tools; description ablation (N=396) uses bare-signature ToolSpecs. Fail\,=\,tool error rate; Bypass\,=\,tool-using problems whose final answers lack an explicit readout anchor.}
\label{tab:ablation}
\vspace{-0.5\baselineskip}
\footnotesize\setlength\tabcolsep{4pt}
\begin{tabular}{@{}l|l r@{\,}l r r@{}}
\toprule
\textbf{Type} & \textbf{Setting} & \multicolumn{2}{c}{\textbf{Acc.}} & \textbf{Fail} & \textbf{Bypass} \\
\midrule
\multirow{3}{*}{Query} & Full CT       & 96.3 &                 & 5.4\%  &   7 \\
                       & w/o meas.     & 92.1 & \graydown{4.2}  & 7.6\%  & 256 \\
                       & w/o query     & 91.3 & \graydown{5.0}  & 8.9\%  & 286 \\
\midrule
\multirow{2}{*}{Desc.} & Full CT       & 95.2 &                 & 5.7\%  &   7 \\
                       & Bare signature& 93.4 & \graydown{1.8}  & 8.3\%  &  32 \\
\bottomrule
\end{tabular}
\vspace{1.5ex}

\noindent\textit{Graded query ablation answer-path migration (\%).\\ablation intensity $\uparrow$ from left $\rightarrow$ right.}
\vspace{0.2ex}

\setlength\tabcolsep{3pt}
\begin{tabular}{@{}lrrr@{}}
\toprule
\textbf{Answer path} & \textbf{Full CT} & \textbf{w/o meas.} & \textbf{w/o all query} \\
\midrule
Query oracle & 44.6 & 0.3 & 1.1 \\
Hybrid derived & 31.9 & 16.9 & 10.3 \\
Resilient fallback & 11.1 & 33.0 & 29.3 \\
Construction return & 1.3 & 43.5 & 52.2 \\
\midrule
Tool-domain total & 98.7 & 94.0 & 93.7 \\
\bottomrule
\end{tabular}
\vspace{-2\baselineskip}
\end{wraptable}

Across benchmarks, 74--91\% of correct CT answers use at least one engine-returned value; pure LLM-bypass stays below 9\% everywhere and below 4\% on six of eight benchmarks (Appendix~\ref{app:mechanism}). Table~\ref{tab:ablation} isolates the interface channels. Removing measurement tools drops accuracy by 4.2 points and drives bypass from 7 to 256 problems: once verified readout is unavailable, more final answers lose an explicit engine anchor.
The lower inset shows the corresponding migration: explicit query-oracle answers nearly vanish, while most answers remain inside the tool domain through resilient fallback and construction-return paths.

Stripping ToolSpec descriptions to bare signatures preserves the invocation interface but degrades tool selection; bypass rises from 7 to 32 because more final answers lack an explicit readout anchor despite preceding tool use. The largest loss occurs on vision-only inputs, where text provides little cue for selecting among construction tools (Appendix~\ref{app:desc-ablation}). This description ablation identifies the ToolSpec surface as a behavioral lever. A per-parameter micro-ablation further shows that non-mutating ToolSpec overlays can support source-preserving harness refinement (Appendix~\ref{app:harness-micro}), consistent with concurrent \emph{Agentic Harness Engineering} findings that harness components beyond prompt wording can drive measured gains~\cite{AHE}.

\subsection{Residual Errors Rooted in Perception and Strategy}
\label{sec:recovery}
Per-step engine verdicts already sit near the ceiling: construction fidelity holds at 95.94\% predicate-level, and outcome accuracy exceeds 90\% on seven of ten benchmarks. Residual failures therefore arise outside the per-action verifier. Some precede the loop: when text and diagram are ambiguous or inconsistent, Measurement Faithfulness can only certify the canvas that the model has chosen to build.

The larger residual is strategic. The engine certifies each action, while the plan threading those actions together lies outside its scope. This step-to-strategy gap accounts for the harder derived predicates (\S\ref{sec:process-fidelity}), long-chain failures (Appendix~\ref{app:trajectory}), the moderate-complexity sweet spot (Table~\ref{tab:main-results}), and the regression on the saturated cluster (\textit{Geo3K}, \textit{PGPS9K}) under forced tool use~\cite{ActingLess,SMART,ToolLight}. Both benchmarks reach a high BL ceiling and are linked by training-data overlap: 78 of 1000 \textit{PGPS9K}-test items are direct \textit{Geo3K}-test problems. Their per-problem transitions cluster at the same low Win ratio (\textit{Geo3K}: 6 saves vs.\ 23 breaks; \textit{PGPS9K}: 13 vs.\ 53; Appendix~\ref{app:outcome-transition}), a regime where the baseline already has a cheap internal route and extra construction mainly introduces avoidable execution overhead. The next lever is therefore policy-level: deciding when engine invocation is warranted (\S\ref{sec:future}).

\section{Conclusion}

Draw2Think makes VLM geometry solving more accountable by moving intermediate geometric claims onto a constraint-first GeoGebra canvas. The experiments show that this external state gives measurements engine-level faithfulness while leaving construction fidelity auditable: the construction harness improves outcome accuracy most when visual measurement or spatial construction is the bottleneck, passes \textbf{95.9\%} predicate-level construction checks, and attains \textbf{68.2}/\textbf{90.5} strict/relaxed rendering accuracy on \textit{GenExam-math}. The same evidence also bounds the claim: grounding adds overhead when the model has an efficient internal route, whereas engine faithfulness applies to the model-selected canvas, with semantic alignment to the target configuration left to construction fidelity.

\paragraph{Limitations.}\label{sec:limitations}
The engine certifies individual actions, leaving trajectory-level strategy to the model; failures therefore concentrate in ambiguous perception, long construction chains, and cases where tool use is unnecessary. The current framework targets single numeric Euclidean instances and assumes sufficient tool-use competence.

\paragraph{Future Work.}\label{sec:future}

\noindent\textbf{Toward strategy-level verification.}\quad
The next harness layer should make the strategy easier to maintain: expose the construction-DAG topology so the planner can see dependencies, progressively disclose ToolSpec groups to reduce the active action space, and return symbolic query forms that later calls can reuse.

\noindent\textbf{Richer intermediate state and proximal twins.}\quad
The present observation channel is deliberately narrow: typed objects, object deltas, and exact scalar readouts. Rendered feedback could expose layout and region-selection errors, while DAG summaries could make long traces easier to assimilate. Beyond geometry, a workflow can support the same loop once it establishes a \emph{proximal twin}: an efficient executable surrogate that rejects invalid actions and returns local state changes. Such a twin need not model the full deployment environment; it only needs to make intermediate actions checkable.

\noindent\textbf{Trajectories as reusable process records.}\quad
A verified construction trajectory records more process structure than textual or visual chain-of-thought transcripts~\cite{CoT, VisualCoT, Gemma4, DeepSeek_TwVP}: every accepted step carries typed dependencies, an engine verdict, and a concrete canvas effect. This record preserves per-action ground truth, exposes tool-selection and orchestration statistics, and supports controlled problem generation by replaying or perturbing verified subgraphs. These trajectories can also close the loop back into training~\cite{Faire}: strong general-purpose models can run the harness at scale, retain engine-verified traces, and use the resulting process labels to adapt smaller domain models. Draw2Think points to a broader design principle for tool-mediated agents: reliable task delivery is governed by which state is externalized, which operations are typed, which verdicts are exact, and which traces remain reusable.

\newpage
\bibliographystyle{plainnat}
\bibliography{references}  

\appendix  
\numberwithin{table}{section}
\numberwithin{figure}{section}
\numberwithin{equation}{section}


\clearpage

\section{Positioning in the Broader Agentic Landscape}
\label{app:agentic-landscape}

The Propose-Draw-Verify loop in Draw2Think instantiates a broader agent-engine pattern: a frontier agentic model selects actions, an external engine executes them and returns feedback, and the model refines its plan accordingly. Similar action-execution-feedback loops appear across scientific computing and general agentic reasoning. Table~\ref{tab:comparison} anchors the geometry comparison; the following notes position Draw2Think relative to post-generation verification, modular agents, harness engineering, tool-use decomposition, and training-time adaptation.

\newcommand{\rwyes}{\(\checkmark\)}
\newcommand{\rwno}{\(\times\)}
\newcommand{\rwpart}{\(\circ\)}
\begin{table}[H]
  \centering
  \setlength{\abovecaptionskip}{2pt}
  \setlength{\belowcaptionskip}{1pt}
  \caption{Representative geometry systems by verification channel; \(\checkmark\)/\(\circ\)/\(\times\) = full / partial / absent support at inference. DOM = document object model. Verdicts apply to each system's inference-time loop, not its training pipeline.}
  \label{tab:comparison}
  \scriptsize
  \setlength{\tabcolsep}{3pt}
  \renewcommand{\arraystretch}{0.9}
  \begin{tabularx}{\linewidth}{@{}>{\raggedright\arraybackslash}p{0.16\linewidth}>{\raggedright\arraybackslash}X *{6}{>{\centering\arraybackslash}p{0.075\linewidth}}@{}}
  \toprule
  \textbf{System} & \textbf{Substrate} &
  \makecell{\textbf{Step}\\\textbf{feedback}} &
  \makecell{\textbf{Invalid}\\\textbf{signal}} &
  \makecell{\textbf{Action}\\\textbf{exact}} &
  \makecell{\textbf{Value}\\\textbf{readout}} &
  \textbf{Revisable} &
  \makecell{\textbf{No ref.}\\\textbf{image}} \\
  \midrule
  AlphaGeo2~\cite{AlphaGeometry2}   & Formal language + prover & \rwpart & \rwpart & \rwpart & \rwno   & \rwpart & \rwyes \\
  InternGeo~\cite{InternGeometry}   & Interactive prover & \rwyes & \rwyes  & \rwpart & \rwno   & \rwpart & \rwyes \\
  GeoVLMath~\cite{GeoVLMath}        & Aux-line text     & \rwno  & \rwno   & \rwno   & \rwno   & \rwno   & \rwyes  \\
  Socratic-Geo~\cite{SocraticGeo}   & Synthetic code/data & \rwno & \rwno   & \rwno   & \rwno   & \rwno   & \rwyes  \\
  GeoCode~\cite{GeoCode}            & Image-to-code     & \rwno  & \rwpart & \rwno   & \rwno   & \rwpart & \rwno   \\
  Visual Sketchpad~\cite{VisualSketchpad} & Python sketch & \rwpart & \rwpart & \rwno   & \rwno   & \rwpart & \rwyes \\
  GeoSketch~\cite{GeoSketch}        & Logic form + render & \rwpart & \rwpart & \rwno & \rwno & \rwpart & \rwyes \\
  Canvas-CoT~\cite{CanvasCoT}       & DOM + render      & \rwyes & \rwpart & \rwno   & \rwno   & \rwyes  & \rwno  \\
  \midrule
  \textbf{Draw2Think}               & \textbf{Constraint canvas} & \rwyes & \rwyes & \rwyes & \rwyes & \rwyes & \rwyes \\
  \bottomrule
  \end{tabularx}
\end{table}

\paragraph{Notes on Table~\ref{tab:comparison} verdicts.}
\emph{Action exact} requires per-action engine certification: formal-language provers~\cite{AlphaGeometry2,InternGeometry} verify proof chains rather than each LLM-proposed primitive, and so receive \(\circ\). \emph{Step feedback} requires a substrate-side signal; VLM re-perception of a rendered sketch~\cite{VisualSketchpad,GeoSketch} is \(\circ\), while structured DOM mutations with a typed critique~\cite{CanvasCoT} are \(\checkmark\). \emph{Revisable} likewise separates per-step rollback from trajectory-level exploration: AlphaGeometry-style sibling search~\cite{AlphaGeometry2}, long-horizon trial-and-error~\cite{InternGeometry}, and re-rendering loops~\cite{VisualSketchpad,GeoSketch} all receive \(\circ\). Cross-modal reward models~\cite{GeoVLMath} and offline data-synthesis generators~\cite{SocraticGeo} consume ground-truth diagrams during training but not during solving, so the corresponding \emph{No ref. image} entries are \(\checkmark\); Canvas-CoT's critic, by contrast, evaluates each step against a target diagram and is marked \(\times\).

\paragraph{Relation to post-generation GeoGebra verification.}
Faire~\cite{Faire} also uses GeoGebra constructions as executable objects. Its task formulation casts geometric interleaving as generation of textual reasoning and GeoGebra code, followed by post-generation verification and reinforcement-learning (RL) reward. In that setting, syntactic and geometric errors can accumulate inside a generated code block before the verifier exposes them. Draw2Think makes a narrower decomposition: the initial problem perception remains model-dependent; after that, intermediate geometric state, exact measurement, and rendering are externalized through typed ToolSpec calls to the constraint engine. This shifts the bottleneck from continuous visual reasoning to discrete action selection over a verified state space.

\paragraph{Single agent vs.\ modular agentic system.}
AgentFlow~\cite{AgentFlow} decomposes agentic reasoning into four trained modules (planner, executor, verifier, generator) coordinated through shared memory, and trains only the planner via on-policy RL (Flow-GRPO, a group-relative policy-optimization variant). Draw2Think collapses three of those roles into a single deterministic engine: GeoGebra runs the typed construction (\emph{executor}), returns exact algebraic values or errors (\emph{verifier}), and emits the resulting object deltas, scalar readouts, and rendered canvases (\emph{generator}), leaving planning, tool selection, and parameter binding to a frozen LLM. Verification \emph{and} generation are delegated rather than learned.

\paragraph{Curated vs.\ auto-discovered harnesses.}
A parallel line treats the harness itself as the optimization target, consistent with broader accounts of externalized memory, skills, protocols, and harness engineering in LLM agents~\cite{ExterLLMAgent}. AutoHarness~\cite{AutoHarness} searches over code harnesses, while AHE~\cite{AHE} reports component-level evidence that tools, middleware, and memory can matter more than prompt wording alone. Draw2Think occupies a constraint-domain-specific point on this spectrum. The substrate is a pre-existing engine with a symbolic verifier, while the ToolSpec layer (79 planar-solving tools, plus 21 solid-geometry extensions and 13 rendering tools when needed) was bootstrapped from GeoGebra's official command manual and then curated across roughly eight rounds in response to pilot failures. The verifier remained symbolic throughout; the ToolSpec surface owes its shape to LLM-authored drafts plus error-driven curation. These regimes are complementary: auto-discovery scales where harness construction is cheap, whereas substrate-fixed designs can deliver deterministic correctness guarantees when the environment is already a constraint engine. In this geometry setting, surface curation around a fixed verifier lets a frozen frontier model close gaps otherwise addressed by retraining.

\paragraph{Decoupling reasoning, tool search/selection, and tool execution.}
Recent work decomposes reasoning, tool selection, and execution at different granularities. AgentFlow~\cite{AgentFlow} separates tool selection (planner) from execution (executor). Pattern-Aware TIR~\cite{PatternAwareTIR} identifies two distinct tool-use strategies, the \emph{calculator-pattern} (text reasoning plus spot computation) and the \emph{algorithmic-pattern} (encoding the full problem as a program), and shows that strategy--problem mismatch is the primary cause of tool-use errors. Draw2Think's ToolSpec descriptions perform an analogous role at prompt time: query tools (e.g., \texttt{Angle}, \texttt{Distance}) correspond to the calculator pattern, while construction tools (e.g., \texttt{Point}, \texttt{Line}, \texttt{Circle}) correspond to the algorithmic pattern. The description ablation in \S\ref{sec:analysis} supports this view: removing semantic guidance degrades both valid tool routing and readout anchoring.

\paragraph{Verification fidelity and token efficiency.}
Systems vary in how they verify intermediate state, forming a fidelity gradient: no feedback (open-loop code generation~\cite{GGBench}), block-level DSL execution with numerical checks~\cite{GeoBuildBench}, binary outcome reward~\cite{AgentFlow,DeepSeekMath}, approximate visual re-perception~\cite{GeoSketch,CanvasCoT}, and exact continuous measurement (Draw2Think). As argued by~\citet{UnderstandingTIR}, tool calls compress token-expensive manual simulation into structured I/O. In geometry, an \texttt{Angle(A,B,C)} query replaces coordinate extraction, dot-product computation, and inverse-trigonometric readout in one verified step. Human-readable ToolSpec names (e.g., \texttt{query\_angle}, \texttt{add\_perpendicular\_line}) keep that compressed trace interpretable: each token-saving call remains a named geometric operation.

\paragraph{From inference-time to training-time.}
Draw2Think operates training-free: capability improvements come from better base models and prompt design rather than task-specific fine-tuning. This avoids the reasoning--tool-use gradient tension identified by DART~\cite{DART}, while leaving further gains to training-time adaptation. Recent training approaches already broadcast trajectory-level reward to all turns~\cite{AgentFlow} or align tool-use strategy with problem structure~\cite{PatternAwareTIR}. Every Draw2Think trajectory exposes structured step-level signals (accept/reject per construction and exact values per query) that could serve as process-dense reward in an analogous \emph{GeometryGym} environment, bridging the gap from inference-time to training-time capability.


\section{Framework Details}

\subsection{ToolSpec Catalog}
\label{app:toolspec}

\subsubsection{Catalog}

The wrapper exposes three disjoint ToolSpec sets per evaluation context: 79 \emph{solving} tools (Table~\ref{tab:toolspec-catalog}; 55 construction + 24 query, all planar benchmarks); 21 solid-geometry primitives (Table~\ref{tab:toolspec-catalog-3d}; activated for \textit{MathVerse-solid}, \textit{SolidGeo}, GenExam-Solid); and 13 rendering tools (Table~\ref{tab:toolspec-render}; restricted to the GenExam rendering pipeline). Each ToolSpec carries a description, typed parameters, and preconditions — the catalog tables use summaries; full descriptions appear in Table~\ref{tab:toolspec-full}.

\paragraph{Provenance and licensing.}
Every ToolSpec wraps a GeoGebra command from the official manual\footnote{\url{https://github.com/geogebra/manual/tree/main/en/modules/ROOT/pages}}; the manual's six command families form a much larger pool from which ours is a curated subset. The current surface was drafted with LLM assistance, then audited against pilot-run failures for per-call determinism. GeoGebra is open source under EUPL v1.2\footnote{\url{https://github.com/geogebra/geogebra}}, so the stack is reproducible from open components; the wrapper will be released with the paper.

{\footnotesize
\setlength{\tabcolsep}{4pt}
\begin{longtable}{@{}llp{6.0cm}@{}}
\caption{Solving ToolSpec catalog (79 tools: 55 construction + 24 query, exposed for all solving benchmarks). C = Construction, Q = Query, T = Transform, U = Utility.}
\label{tab:toolspec-catalog} \\
\toprule
\textbf{Type} & \textbf{ToolSpec Name} & \textbf{Summary Description} \\
\midrule
\endfirsthead
\multicolumn{3}{@{}l}{\small\itshape (Table~\ref{tab:toolspec-catalog} continued)} \\
\toprule
\textbf{Type} & \textbf{ToolSpec Name} & \textbf{Summary Description} \\
\midrule
\endhead
\midrule
\multicolumn{3}{r}{\footnotesize\itshape Continued on next page} \\
\endfoot
\bottomrule
\endlastfoot
\multicolumn{3}{@{}l}{\textbf{Points (4)}} \\
C & \texttt{add\_point} & Place a free point at $(x, y)$ \\
C & \texttt{add\_point\_on} & Point constrained to a curve or path \\
C & \texttt{add\_intersect} & Intersection of two objects (indexed) \\
C & \texttt{add\_midpoint} & Midpoint of two points \\
\midrule
\multicolumn{3}{@{}l}{\textbf{Lines \& Segments (10)}} \\
C & \texttt{add\_segment} & Bounded segment between two points \\
C & \texttt{add\_line} & Infinite line through two points \\
C & \texttt{add\_ray} & Ray from a point through another \\
C & \texttt{add\_vector} & Vector from point to point \\
C & \texttt{add\_perpendicular\_line} & Line perpendicular to $L$ through $P$ \\
C & \texttt{add\_perpendicular\_bisector} & Perpendicular bisector of a segment \\
C & \texttt{add\_parallel\_line} & Line parallel to $L$ through $P$ \\
C & \texttt{add\_angle\_bisector} & Bisector of angle $\angle ABC$ \\
C & \texttt{add\_tangent} & Tangent line to a conic through a point \\
C & \texttt{add\_tangent\_conic\_conic} & Common tangent of two conics \\
\midrule
\multicolumn{3}{@{}l}{\textbf{Circles \& Conics (9)}} \\
C & \texttt{add\_circle} & Circle by center and radius/point \\
C & \texttt{add\_arc} & Circular arc \\
C & \texttt{add\_sector} & Circular sector \\
C & \texttt{add\_semicircle} & Semicircle on a diameter \\
C & \texttt{add\_circle\_3\_points} & Circumscribed circle through 3 points \\
C & \texttt{add\_incircle} & Inscribed circle of a triangle \\
C & \texttt{add\_ellipse} & Ellipse by foci and point \\
C & \texttt{add\_parabola} & Parabola by focus and directrix \\
C & \texttt{add\_hyperbola} & Hyperbola by foci and point \\
\midrule
\multicolumn{3}{@{}l}{\textbf{Polygons \& Centers (5)}} \\
C & \texttt{add\_polygon} & Polygon from vertex list \\
C & \texttt{add\_regular\_polygon} & Regular $n$-gon \\
C & \texttt{add\_vertex} & Extract vertex from polygon \\
C & \texttt{add\_center} & Center of a circle or conic \\
C & \texttt{add\_triangle\_center} & Centroid, incenter, circumcenter, etc. \\
\midrule
\multicolumn{3}{@{}l}{\textbf{Measurements (4)}} \\
C & \texttt{add\_angle} & Construct and display an angle object \\
C & \texttt{add\_distance} & Construct a distance object \\
C & \texttt{add\_area} & Construct an area object \\
C & \texttt{add\_slope} & Construct a slope object \\
\midrule
\multicolumn{3}{@{}l}{\textbf{Functions \& Calculus (8)}} \\
C & \texttt{add\_function} & Define $f(x) = \text{expr}$ \\
C & \texttt{add\_derivative} & Derivative of a function \\
C & \texttt{add\_integral\_function} & Antiderivative \\
C & \texttt{add\_inflection\_point} & Inflection point(s) of a function \\
C & \texttt{add\_asymptote} & Asymptote(s) of a function \\
C & \texttt{add\_curve} & Parametric curve \\
C & \texttt{add\_roots} & Root(s) of a function \\
C & \texttt{add\_turning\_point} & Local extrema \\
\midrule
\multicolumn{3}{@{}l}{\textbf{Other Construction (5)}} \\
C & \texttt{add\_slider} & Numeric slider \\
C & \texttt{add\_best\_fit\_line} & Linear regression line \\
C & \texttt{add\_inequality} & Region defined by inequality \\
C & \texttt{add\_integral\_shade} & Shaded definite integral region \\
C & \texttt{add\_text} & Text label on canvas \\
\midrule
\multicolumn{3}{@{}l}{\textbf{Transforms (5)}} \\
T & \texttt{transform\_reflect\_line} & Reflect object across a line \\
T & \texttt{transform\_reflect\_point} & Reflect object across a point \\
T & \texttt{transform\_rotate} & Rotate object by angle around center \\
T & \texttt{transform\_translate} & Translate by vector \\
T & \texttt{transform\_dilate} & Dilate from center by factor \\
\midrule
\multicolumn{3}{@{}l}{\textbf{Utility (4)}} \\
U & \texttt{delete\_object} & Cascading delete from DAG \\
U & \texttt{set\_value} & Set slider/free numeric value \\
U & \texttt{rename\_object} & Rename an object \\
U & \texttt{set\_label\_visible} / \texttt{set\_object\_visible} & Toggle visibility \\
\midrule
\multicolumn{3}{@{}l}{\textbf{Geometric Measurement (9)}} \\
Q & \texttt{query\_angle} & Angle measure at vertex (degrees) \\
Q & \texttt{query\_distance} & Shortest distance between two objects \\
Q & \texttt{query\_length} & Length of segment, arc, or vector \\
Q & \texttt{query\_perimeter} & Perimeter of polygon or circle \\
Q & \texttt{query\_area} & Area of polygon, circle, or sector \\
Q & \texttt{query\_slope} & Slope of a line \\
Q & \texttt{query\_radius} & Radius of circle/arc \\
Q & \texttt{query\_x\_coord} / \texttt{query\_y\_coord} & Point coordinates \\
\midrule
\multicolumn{3}{@{}l}{\textbf{Geometric Verification (8)}} \\
Q & \texttt{query\_are\_parallel} & Test parallelism \\
Q & \texttt{query\_are\_perpendicular} & Test perpendicularity \\
Q & \texttt{query\_is\_tangent} & Test tangency \\
Q & \texttt{query\_is\_in\_region} & Point-in-region test \\
Q & \texttt{query\_are\_equal} & Geometric identity test \\
Q & \texttt{query\_are\_collinear} & Collinearity test \\
Q & \texttt{query\_are\_concyclic} & Concyclicity test \\
Q & \texttt{query\_are\_congruent} & Congruence test \\
\midrule
\multicolumn{3}{@{}l}{\textbf{CAS \& Canvas Inspection (7)}} \\
Q & \texttt{query\_solve} & CAS symbolic equation solver \\
Q & \texttt{query\_nsolve} & CAS numeric equation solver \\
Q & \texttt{query\_definite\_integral} & Numeric definite integral \\
Q & \texttt{query\_function\_max} / \texttt{query\_function\_min} & Function extrema in interval \\
Q & \texttt{query\_is\_defined} & Object existence check \\
Q & \texttt{query\_dependents} & List dependent objects (DAG inspection) \\
\end{longtable}
}

{\footnotesize
\setlength{\tabcolsep}{4pt}
\begin{longtable}{@{}l l p{8cm}@{}}
\caption{Solid-geometry ToolSpec extensions (21 tools, activated for 3D benchmarks only). C = Construction, Q = Query, R = Render.}
\label{tab:toolspec-catalog-3d} \\
\toprule
\textbf{Type} & \textbf{ToolSpec Name} & \textbf{Summary Description} \\
\midrule
\endfirsthead
\multicolumn{3}{@{}l}{\small\itshape (Table~\ref{tab:toolspec-catalog-3d} continued)} \\
\toprule
\textbf{Type} & \textbf{ToolSpec Name} & \textbf{Summary Description} \\
\midrule
\endhead
\midrule
\multicolumn{3}{r}{\footnotesize\itshape Continued on next page} \\
\endfoot
\bottomrule
\endlastfoot
\multicolumn{3}{@{}l}{\textbf{3D Points \& Vectors (2)}} \\
C & \texttt{add\_point3d} & Place a free point at 3D coordinates $(x,y,z)$ \\
C & \texttt{add\_vector3d} & 3D vector from origin or between two points \\
\midrule
\multicolumn{3}{@{}l}{\textbf{Planes (4)}} \\
C & \texttt{add\_plane} & Infinite mathematical plane \\
C & \texttt{add\_finite\_plane} & Bounded rectangular plane (visual) \\
C & \texttt{add\_perpendicular\_plane} & Plane through point, perpendicular to line/vector \\
C & \texttt{add\_plane\_bisector} & Perpendicular bisector plane of two points \\
\midrule
\multicolumn{3}{@{}l}{\textbf{Solids (7)}} \\
C & \texttt{add\_pyramid} & Pyramid from base polygon + apex \\
C & \texttt{add\_prism} & Prism from base polygon + translation \\
C & \texttt{add\_cone} & Cone from base circle + apex \\
C & \texttt{add\_cylinder} & Cylinder (filled solid, not lateral surface only) \\
C & \texttt{add\_sphere} & Sphere by center and radius/point \\
C & \texttt{add\_tetrahedron} & Regular tetrahedron \\
C & \texttt{add\_cube} & Cube; three points must form a square \\
\midrule
\multicolumn{3}{@{}l}{\textbf{Derived \& Auxiliary (4)}} \\
C & \texttt{add\_cross\_section} & Intersection of plane with a solid \\
C & \texttt{add\_net} & Unfolded net of a convex polyhedron \\
C & \texttt{add\_text\_3d} & 3D-positioned text or LaTeX label \\
C & \texttt{add\_surface\_revolution} & Rotate function around $x$-axis to form surface \\
\midrule
\multicolumn{3}{@{}l}{\textbf{3D Measurement (3)}} \\
Q & \texttt{query\_volume} & Volume of pyramid/prism/cone/cylinder/sphere \\
Q & \texttt{query\_surface\_area} & Total surface area of a 3D solid \\
Q & \texttt{query\_coords3d} & 3D coordinates $(x,y,z)$ of a point \\
\midrule
\multicolumn{3}{@{}l}{\textbf{3D Render (1)}} \\
R & \texttt{render\_set\_3d\_view} & Configure viewport: rotation, zoom, axes \\
\end{longtable}
}

\begin{table}[H]
\centering
\footnotesize
\setlength{\tabcolsep}{3.5pt}
\caption{Rendering ToolSpec catalog (13 tools, restricted to the GenExam rendering pipeline). R = Render. Used to style canvas objects for the rendered output; excluded from solving benchmarks.}
\label{tab:toolspec-render}
\begin{tabular}{@{}l l p{6.0cm}@{}}
\toprule
\textbf{Type} & \textbf{ToolSpec Name} & \textbf{Summary Description} \\
\midrule
R & \texttt{render\_set\_color} & Set object color \\
R & \texttt{render\_set\_line\_style} & Set line style (solid, dashed, dotted) \\
R & \texttt{render\_set\_line\_thickness} & Set line thickness \\
R & \texttt{render\_set\_point\_style} & Change point marker shape \\
R & \texttt{render\_set\_point\_size} & Change point marker size \\
R & \texttt{render\_set\_filling} & Set fill opacity of closed shapes \\
R & \texttt{render\_set\_decoration} & Add tick marks or arrows to segments \\
R & \texttt{render\_show\_axes} & Show or hide coordinate axes \\
R & \texttt{render\_show\_grid} & Show or hide coordinate grid \\
R & \texttt{render\_set\_caption} & Set custom label caption \\
R & \texttt{render\_set\_label\_mode} & Control label display mode \\
R & \texttt{render\_set\_coord\_system} & Set visible viewport bounds \\
R & \texttt{render\_add\_right\_angle\_mark} & Add right-angle square marker \\
\bottomrule
\end{tabular}
\end{table}

\subsubsection{Display Style Presets}
\label{app:toolspec-style}

Beyond the per-call \texttt{render\_*} tools (Table~\ref{tab:toolspec-render}), a single preset \texttt{apply\_textbook\_style()} flips the canvas from GeoGebra defaults to a publication-grade textbook style in one call. Both presets share the same engine and primitives; the visual defaults differ (Table~\ref{tab:display-style}).

\begin{table}[H]
  \centering
  \footnotesize
  \setlength{\tabcolsep}{6pt}
  \caption{Display-style presets in the wrapper. The default preset preserves GeoGebra's stock interactive appearance; the textbook preset overrides it for paper-grade rendering and is applied to all rendered galleries shown in this paper. The rightmost column flags whether the LLM can override that dimension per-object during the GenExam rendering pipeline via the corresponding \texttt{render\_*} tool (Table~\ref{tab:toolspec-render}); \emph{preset only} dimensions are set by the preset switch alone.}
  \label{tab:display-style}
  \begin{tabular}{@{}llll@{}}
  \toprule
  \textbf{Dimension} & \textbf{Default} & \textbf{Textbook} & \textbf{LLM-adjustable} \\
  \midrule
  Object color        & GeoGebra palette (blue points, etc.)  & Pure black                       & per-call    \\
  Line thickness      & 2--3\,pt                              & 1\,pt                            & per-call    \\
  Point marker        & Blue circle, size 5                   & Black filled circle, size 3      & per-call    \\
  Axes                & Visible                               & Hidden                           & per-call    \\
  Grid                & Visible                               & Hidden                           & per-call    \\
  Polygon fill        & Translucent palette color             & None (wireframe only)            & per-call    \\
  Hidden 3D edges     & Solid lines (engine default)          & Dashed (textbook convention)     & preset only \\
  Background          & Light gray                            & White                            & preset only \\
  Label rendering     & Upright sans-serif                    & Italic NAME mode                 & per-call    \\
  \bottomrule
  \end{tabular}
\end{table}

The preset is opt-in and idempotent: applying it after construction repaints existing objects and registers a listener for subsequent ones, so the same ToolSpec sequence preserves the construction graph while changing the visual policy. The textbook conventions match concurrent benchmark work~\cite{WeMath2}.

\paragraph{Engine state as typed XML.}
GeoGebra's native \texttt{.ggb} format is a ZIP wrapping \texttt{geogebra.xml}: a typed tree of \texttt{<construction>} → \texttt{<element>} / \texttt{<expression>} / \texttt{<command>} nodes (\texttt{Point}, \texttt{LineBisector}, \texttt{Intersect}, \texttt{Circle}, ...), with style attributes attached per-element. ToolSpec calls and engine observations serialize as typed records over this same graph, mirroring Anthropic's XML-tag prompt convention\footnote{\url{https://docs.anthropic.com/en/docs/build-with-claude/prompt-engineering/use-xml-tags}.}; the textbook-style preset is its visual analogue.

\subsubsection{Representative Full Descriptions}
\label{app:toolspec-full}

The catalog tables summarise; the model actually sees preconditions, parameter constraints, and usage guidance. We reproduce six \emph{complete} ToolSpec descriptions verbatim as passed to the model, chosen to illustrate error-prevention guardrails, geometric-convention encoding, and the description-accuracy link validated in \S\ref{sec:toolspec}.

\small
\begin{longtable}{@{}p{0.22\linewidth}p{0.74\linewidth}@{}}
\caption{Six representative ToolSpecs with full descriptions. \textbf{Bold} highlights design-critical phrases. Parameters listed below each description.}
\label{tab:toolspec-full} \\
\toprule
\textbf{ToolSpec} & \textbf{Full Description \& Parameters} \\
\midrule
\endfirsthead
\toprule
\textbf{ToolSpec} & \textbf{Full Description \& Parameters} \\
\midrule
\endhead
\bottomrule
\endlastfoot

{\footnotesize\texttt{add\_intersect}}
\newline\textit{(highest failure rate)}
&
Create intersection point(s) of two geometric objects. Both obj1 and obj2 must be lines, segments, rays, circles, conics, or functions---\textbf{NOT points or numerics}. Omit index to get ALL intersections as a single point (best for segment-segment or line-line which have exactly one). \textbf{Use index=1 or index=2 when two objects have multiple intersections} (e.g.\ line-circle). The result is stored under the exact `name' you provide---\textbf{do NOT append \_1 or \_2 suffixes} to access it.
\newline\textit{Params:} \texttt{name} (string), \texttt{obj1} (string, NOT a point), \texttt{obj2} (string, NOT a point), \texttt{index} (integer, optional).
\\
\midrule

{\footnotesize\texttt{add\_perpendicular\_line}}
\newline\textit{(constructive axiom)}
&
Create a line through a point, perpendicular to a reference line/segment/ray. The `point' must be a point object, and `line' must be a line, segment, or ray---\textbf{NOT a point}. \textbf{If you have two points A and B and want a perpendicular at A, first create the line: L=Line(A,B), then PerpendicularLine(A, L).}
\newline\textit{Params:} \texttt{name} (string), \texttt{point} (string, must be point), \texttt{line} (string, must be line/segment/ray).
\\
\midrule

{\footnotesize\texttt{add\_tangent}}
\newline\textit{(strategy guidance)}
&
Create tangent line(s) from an external point to a conic (circle, ellipse, hyperbola, parabola). \textbf{ALWAYS use this tool for tangent lines}---do NOT manually construct tangents with add\_segment or add\_line. May return one or two lines stored under the exact `name' you provide. \textbf{Do NOT append \_1 or \_2 suffixes}---use the name directly. For a reliable single tangent contact point, \textbf{prefer the Thales-circle construction} (Midpoint + Circle + Intersect).
\newline\textit{Params:} \texttt{name} (string), \texttt{point} (string), \texttt{conic} (string).
\\
\midrule

{\footnotesize\texttt{add\_semicircle}}
\newline\textit{(convention encoding)}
&
Create a semicircle with the given segment as diameter. The arc is drawn on the \textbf{LEFT side when walking from p1 to p2}. Direction rules: (1) p1=left, p2=right (horizontal) $\to$ arc ABOVE; (2) p1=right, p2=left $\to$ arc BELOW; (3) p1=bottom, p2=top (vertical) $\to$ arc LEFT; (4) p1=top, p2=bottom $\to$ arc RIGHT. \textbf{To flip the arc, SWAP p1 and p2.} Useful for Thales' theorem: any point on the semicircle sees the diameter at $90^\circ$.
\newline\textit{Params:} \texttt{name} (string), \texttt{p1} (string), \texttt{p2} (string).
\\
\midrule

{\footnotesize\texttt{query\_angle}}
\newline\textit{(measurement convention)}
&
Measure the angle at vertex b (the apex), sweeping counter-clockwise from ray $b \!\to\! a$ to ray $b \!\to\! c$. Returns degrees. \textbf{IMPORTANT: point order matters}---Angle(A,B,C) $\neq$ Angle(C,B,A). For the interior angle of triangle ABC at vertex B, \textbf{use Angle(C,B,A)} (go from one side to the other in the direction that gives the interior angle). All three must be existing point objects.
\newline\textit{Params:} \texttt{name} (string), \texttt{a} (string, first arm), \texttt{b} (string, vertex), \texttt{c} (string, second arm).
\\
\midrule

{\footnotesize\texttt{render\_set\_label\_mode}}
&
Control what is displayed as an object's label \textbf{and automatically make the label visible}. 0 = Name only (e.g.\ `A'), 1 = Name + Value (e.g.\ `A = (1,2)'), 2 = Value only (e.g.\ `(1,2)'), 3 = Caption (must set caption first).
\newline\textit{Params:} \texttt{obj} (string), \texttt{mode} (integer: 0/1/2/3).
\\

\end{longtable}

\clearpage
\subsection{Construction DAG and ToolSpec Coverage}
\label{app:dag-proof}

Each construction sequence induces a dependency DAG that is inspectable, prunable, and search-traversable. Table~\ref{tab:dag-proof} shows a 3-4-5 right-triangle build (perpendicular line + circle, then a $BP{=}5$ measurement).

\begin{table}[H]
\centering
\caption{Example construction DAG showing dependency structure. \texttt{deps} = what this step requires, \texttt{used by} = what depends on this step. Cascading delete of any node removes all downstream dependents.}
\label{tab:dag-proof}
\small
\setlength{\tabcolsep}{4pt}
\begin{tabular}{@{}r>{\fontsize{8pt}{8pt}\selectfont\ttfamily}lllll@{}}
\toprule
\textbf{Step} & \normalfont\textbf{Tool Call} & \textbf{Deps} & \textbf{Used by} & \textbf{Geometric Interpretation} & \textbf{Semantic} \\
\midrule
1 & add\_point(A, 0, 0) & $\varnothing$ & \{3,4,5\} & Point existence & $A = (0, 0)$ \\
2 & add\_point(B, 4, 0) & $\varnothing$ & \{3,7\} & Point existence & $B = (4, 0)$ \\
3 & add\_segment(AB, A, B) & \{1,2\} & \{4\} & Two points $\to$ segment & $|AB| = 4$ \\
4 & add\_perp\_line(L, A, AB) & \{1,3\} & \{6\} & Perpendicular through a point & $L \perp AB$ \\
5 & add\_circle(c, A, 3) & \{1\} & \{6\} & Center + radius $\to$ circle & $r = 3$ \\
6 & add\_intersect(P, L, c, 1) & \{4,5\} & \{7\} & Line--circle intersection (1st) & $P = (0, -3)$ \\
7 & query\_distance(B, P) & \{2,6\} & --- & Measurement readout & $d = 5.0$ \checkmark \\
\bottomrule
\end{tabular}
\end{table}

\vspace{-1\baselineskip}
\begin{figure}[H]
\centering
\includegraphics[width=0.54\linewidth]{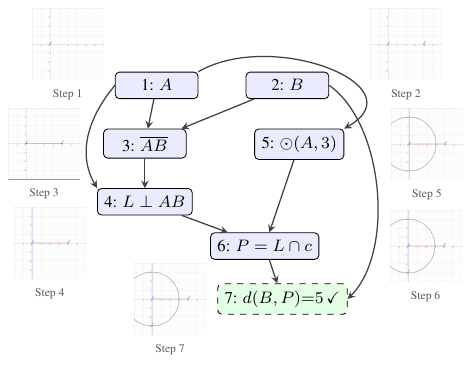}
\caption{Dependency DAG for the construction in Table~\ref{tab:dag-proof}, with per-step engine-canvas thumbnails embedded around the corresponding nodes. Blue~$=$ construction step (engine-verified); dashed green $=$ query (measurement). Each thumbnail shows the canvas immediately after the labelled step (e.g.\ Step~3 shows $\overline{AB}$ on top of points $A,B$, Step~6 adds intersection $P=(0,-3)$). Removing any node cascades to all downstream dependents.}
\label{fig:dag-example}
\end{figure}

\vspace{-1\baselineskip}
\textbf{Edge semantics.}\;Nodes are typed construction calls (\texttt{add\_*}, blue) and read-only queries (\texttt{query\_*}, dashed green); directed edges encode \emph{parent$\to$child} dependencies, built dynamically as the LLM emits tools. Three properties hold by construction:
\begin{itemize}[leftmargin=*,nosep,topsep=2pt,itemsep=1pt]
\item \emph{Proof-like dependency.}\;Step~6 requires steps~4 and~5; deleting either cascades and invalidates step~6 and below, mirroring premise retraction.
\item \emph{Instance-level verification.}\;Step~7 certifies $d(B,P){=}5$ for \emph{this} 3-4-5 triangle through exact engine arithmetic; universal quantification ($\forall a,b$) sits beyond the harness (\S6).
\item \emph{Repair under deps.}\;If step~4 fails, the engine preserves \{1,2,3,5\}; the model replans the failed branch alone (\S\ref{sec:recovery}).
\end{itemize}

\paragraph{Construction as graph search.}
Geometry solving is a construction-state search from known objects to a target state, recorded post-hoc by the DAG. The canvas is currently exposed as flat JSON; surfacing DAG topology (parent--child edges, frontier reachability) during planning is left to future work.


\paragraph{Construction DAGs as proof-relevant traces.}
Each successful tool call admits a proof-relevant geometric reading, so the DAG is a \emph{structural analogy} of theorem search rather than functional equivalence with formal provers, giving instance-level causal justification rather than outcome-only reward. GeoLaux~\cite{GeoLaux} reports up to 50\% gap between answer (ACS) and process (PCS) correctness for top models; our engine guarantees step-level exactness for \emph{this instance}, while universally quantified proofs and DAG-based proof-oriented search remain open.

\clearpage


\section{Extended Benchmark Results}
\label{app:cross-system}

Table~\ref{tab:cross-system} compares Draw2Think against the strongest previously reported results on each benchmark.
Prior results use heterogeneous metrics, models, and evaluation protocols; direct comparison should be interpreted with the caveats below.

\begin{table}[H]
\centering
\caption{Cross-system comparison. All Draw2Think results are single-attempt Pass@1, temperature~0.
Blank cells = not evaluated on that benchmark.
Neural-symbolic methods use parsed diagram annotations rather than raw images.
SFT = supervised fine-tuning; RL = reinforcement learning; GRPO = group relative policy optimization.}
\label{tab:cross-system}
\newcommand{\tb}{\fontsize{8}{9.6}\selectfont}
\newcommand{\ts}{\fontsize{5.5}{6.6}\selectfont}
\newcommand{\mcell}[2]{\makecell[l]{#1\\[-3pt]\textcolor{gray}{\ts #2}}}
\tb
\setlength\tabcolsep{3pt}
\begin{tabularx}{\linewidth}{@{}l*{6}{c}X@{}}
\toprule
\textbf{Method} &
  \makecell{\textbf{Geo3K}\,{\ts\cite{Geometry3K}}\\[-1pt]{\ts N{=}601}} &
  \makecell{\textbf{GeoQA/}\\\textbf{UniGeo}\,{\ts\cite{UniGeo}}\\[-1pt]{\ts N{=}754}} &
  \makecell{\textbf{PGPS9K}\,{\ts\cite{PGPS9K}}\\[-1pt]{\ts N{=}1000}} &
  \makecell{\textbf{MathVerse}\\[-1pt]\textbf{Vision-Only}\,{\ts\cite{MathVerse}}\\[-1pt]{\ts N{=}510}} &
  \makecell{\textbf{MathVista}\\[-1pt]\textbf{GPS}\,{\ts\cite{MathVista}}\\[-1pt]{\ts N{=}208}} &
  \makecell{\textbf{Olympiad}\\[-1pt]\textbf{Bench}\,{\ts\cite{OlympiadBench}}\\[-1pt]{\ts N{=}112}} & \\
\midrule
\multicolumn{8}{@{}l}{{\textit{Neural-symbolic (diagram parsing)}}} \\
\mcell{InterGPS~\cite{Geometry3K}}{formal solver} & 57.5 &      &      &      &      &      & \\
\mcell{GeoDRL~\cite{GeoDRL}}{RL + formal solver} & 68.4 &      & 66.7 &      &      &      & \\
\mcell{PGPSNet-v2~\cite{PGPS9K}}{neural program synthesis} & 76.4 &      & 69.2 &      &      &      & \\
\mcell{Pi-GPS~\cite{PiGPS}}{PGDP + \texttt{o3-mini} solver} & 77.8 &      & 69.8 &      &      &      & \\
\midrule
\multicolumn{8}{@{}l}{{\textit{Open-source VLM (zero-shot)}}} \\
\mcell{Qwen2.5-VL-7B~\cite{Qwen25VL}}{reported by GeoGen~\cite{GeoGen}} & 44.8 & 64.2 & 43.0 & 44.1 & 54.3 &      & \\
\midrule
\multicolumn{8}{@{}l}{{\textit{Open-source fine-tuned (SFT\,/\,RL)}}} \\
\mcell{AVAR~\cite{AVAR}}{\texttt{qwen2.5-vl-7b} + RL} &      &      &      & 50.4 &      &      & \\
\mcell{SocraticGeo~\cite{SocraticGeo}}{\texttt{qwen2.5-vl-7b} + GRPO} &      & 49.2 &      & 45.1 & 63.6 &      & \\
\mcell{NeSyGeo~\cite{NeSyGeo}}{\texttt{qwen2.5-vl-7b} + SFT} &      & 71.8 &      & 46.7 &      &      & \\
\mcell{GeoGen-SFT~\cite{GeoGen}}{\texttt{qwen2.5-vl-7b} + SFT} & 58.4 & 78.0 & 54.3 &      & 74.0 &      & \\
\mcell{GeoSketch-RL~\cite{GeoSketch}}{\texttt{qwen2.5-vl-7b} + SFT+RL} & 28.8 & 72.5 &      &      &      &      & \\
\mcell{MathCanvas~\cite{MathCanvas}}{\texttt{bagel-7b} + SFT} &      &      &      &      & 79.3 &      & \\
\midrule
\multicolumn{8}{@{}l}{{\textit{Closed-source API (zero-shot)}}} \\
\mcell{GPT-4o~\cite{GPT4o}}{reported by TrustGeoGen~\cite{TrustGeoGen}} & 31.5 & 42.3 &      &      &      & 13.4 & \\
\mcell{Gemini-2.5-Pro~\cite{Gemini25Pro}}{reported by TrustGeoGen~\cite{TrustGeoGen}} & 80.7 & 79.6 &      &      &      & 75.0 & \\
\mcell{OpenAI-o3~\cite{OpenAIo3}}{reported by TrustGeoGen~\cite{TrustGeoGen}} & 81.0 & 79.3 &      &      &      & 77.7 & \\
\midrule
\multicolumn{8}{@{}l}{{\textit{Ours (training-free)}}} \\
\mcell{Draw2Think BL}{\texttt{gemini-3-flash-preview}; direct VLM} & \textbf{98.2} & 93.6 & \textbf{94.5} & 87.5 & 97.1 & 89.3 & \\
\mcell{\textbf{Draw2Think CT}}{\textbf{\texttt{gemini-3-flash-preview}; executable environment}} & \makecell{95.3\\{\ts\textcolor{gray}{$-$2.9}}} & \makecell{\textbf{96.9}\\{\ts\textcolor{gray}{+3.3}}} & \makecell{90.5\\{\ts\textcolor{gray}{$-$4.0}}} & \makecell{\textbf{91.0}\\{\ts\textcolor{gray}{+3.5}}} & \makecell{\textbf{97.6}\\{\ts\textcolor{gray}{+0.5}}} & \makecell{\textbf{89.3}\\{\ts\textcolor{gray}{+0.0}}} & \\
\midrule
Human expert & \makecell{90.9\\{\ts\cite{Geometry3K}}} &      &      & \makecell{66.7\\{\ts\cite{MathVerse}}} &      &      & \\
\bottomrule
\end{tabularx}
\end{table}

\noindent\textbf{Caveats.}
(1)~Table rows use Top-1 (single-attempt) accuracy; methods using BoN@K or non-standard test splits are excluded.
(2)~Our BL already exceeds most prior methods, largely due to Gemini-3-Flash's capabilities; the CT--BL gap in Table~\ref{tab:main-results} isolates the framework's contribution.
(3)~For Geo3K and PGPS9K, Table~\ref{tab:cross-system} uses Choice accuracy when prior papers report both Completion and Choice; our numbers follow the same multiple-choice protocol. Closed-source API rows reported by TrustGeoGen~\cite{TrustGeoGen} use Completion accuracy and are included as context rather than direct format-matched comparisons.
(4)~Our GeoQA/UniGeo numbers use the UniGeo~\cite{UniGeo} calculation test split; prior methods report varying GeoQA protocols, so this column should be read as split-matched only where explicitly stated.
(5)~MathVerse human accuracy (66.7\%) is from 10 college students on Vision-Only testmini~\cite{MathVerse}.


\section{Process Fidelity on GeoGoal}
\label{app:geogoal-sgvr}

\begin{figure}[H]
  \centering
  \includegraphics[width=0.92\linewidth]{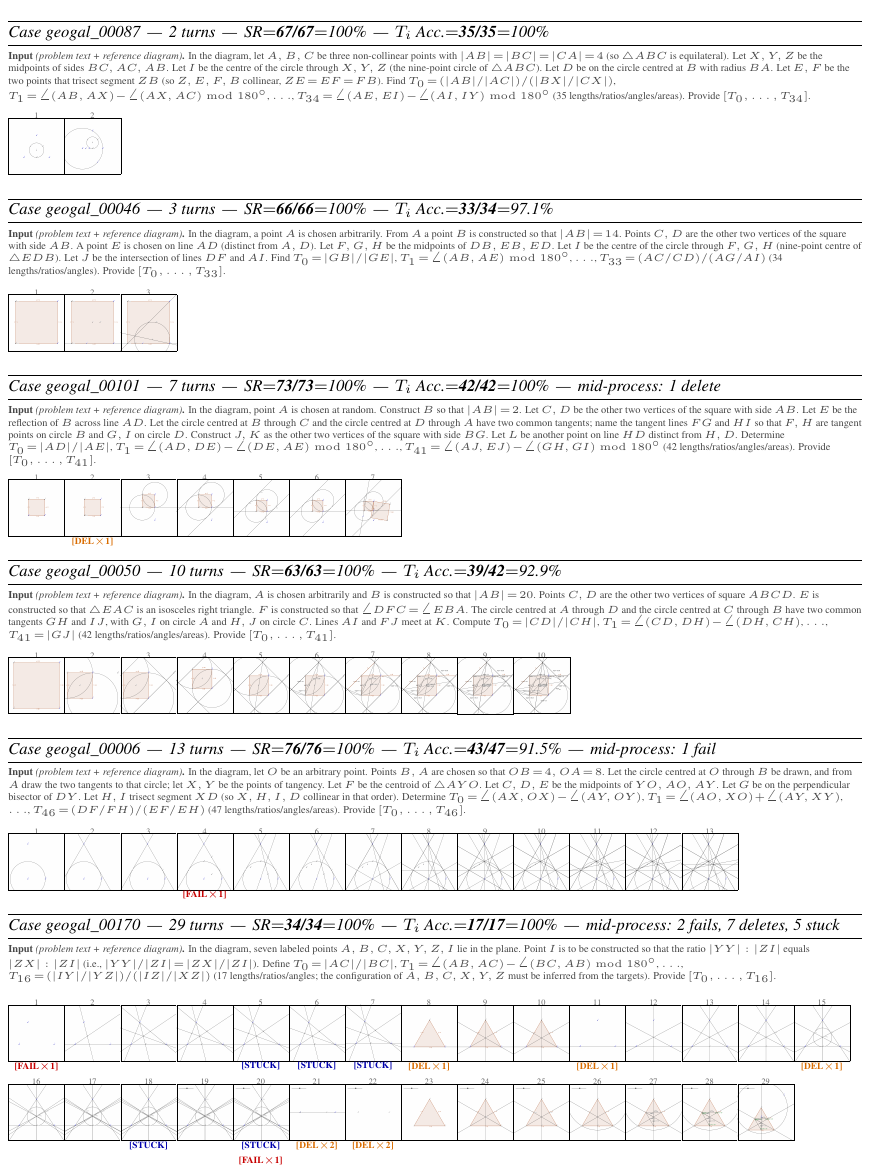}
  \caption{Progressive canvas-construction staircase on GeoGoal. Six representative trajectories (2--29 construction turns) reach $\mathrm{SR}=100\%$; subtitles show predicate pass counts. Badges mark engine errors \textcolor{red!78!black}{\textbf{[FAIL$\times n$]}}, cascading deletions \textcolor{orange!85!black}{\textbf{[DEL$\times n$]}}, and query-only stuck turns \textcolor{blue!68!black}{\textbf{[STUCK]}}.}
  \label{fig:geogoal-progress}
\end{figure}
\vspace{-2pt}
\subsection{Benchmark, Verification, and Metrics}
\label{app:geogoal-setup}

\textbf{Benchmark.}\;GeoGoal~\cite{GeoGoal} is an externally authored, formally verified benchmark of 256 problems synthesized by TrustGeoGen~\cite{TrustGeoGen}: each provides a natural-language construction, $n_i \in [11,66]$ query expressions, and a formal skeleton of $m_i \in [20,113]$ predicates over 20 types (\texttt{cong}, \texttt{coll}, \texttt{perp}, \texttt{para}, \texttt{eqangle}, \texttt{eqratio}, \texttt{midp}, \texttt{cyclic}, \ldots; 7{,}395 queries, 13{,}254 predicates, mean 52/problem).

\textbf{Verification pipeline.}\;On CT completion (\texttt{CONSTRUCTION\_DONE} or two-turn no-tool streak), named-point coordinates from the final canvas are passed to Newclid's~\cite{Newclid,AlphaGeometry2} \texttt{check\_numerical()} (absolute tol.\ $4\!\times\!10^{-7}$ or relative $0.1\%$); a predicate passes iff the relation holds under this tolerance.

\textbf{Metrics.}\;Under SGVR's metric names~\cite{GeoGoal}, with $n_i,\,p_i$ the GT and passing predicate counts for problem $i$:
\begin{itemize}[nosep,leftmargin=1.4em]
  \item \textbf{CR} (Completion Rate): fraction of problems with a non-empty canvas.
  \item \textbf{SR} (Skeleton Rate): predicate-level pass rate $\sum_i p_i / \sum_i n_i$, decomposed into three tiers (\emph{premise}, \emph{numerical check}, \emph{derived}) over SGVR's 30 source tags.
  \item \textbf{SC} (Skeleton Completion): problem-level all-pass rate $|\{i:p_i{=}n_i\}|/N$; a single failure disqualifies the problem.
\end{itemize}
SGVR's Consistency Ratio $\mathrm{SC}/\mathrm{SR}$ is dropped in favour of reporting SR and SC directly.

\subsection{Main Results and Diagnostic Cases}

\begin{figure}[H]
\centering
\includegraphics[width=0.85\linewidth]{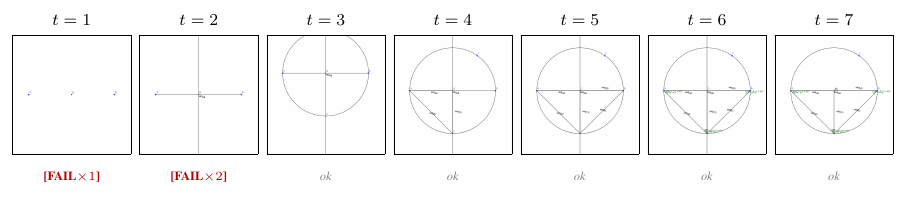}
\caption{Partial-failure case: \texttt{geogal\_00033}, $\mathrm{SR}=43\%$. Seven turns shown; three early failures are rejected, later turns run fail-free, and the final valid canvas still misses 57\% of GT predicates.}
\label{fig:geogoal-failure}
\end{figure}

Figure~\ref{fig:geogoal-progress} shows canvas richness scaling with turn count (emit-only tails omitted, pixel-identical to predecessor). The 29-turn row exposes long-tail behaviours legible from traces: stuck query-only phases, cascade-pruned restarts, fallback to canonical scaffolds (\texttt{add\_regular\_polygon}). Figure~\ref{fig:geogoal-failure} gives the complementary mode: \texttt{geogal\_00033} runs fail-free after three early rejected calls yet only reaches $\mathrm{SR}=43\%$ because the plan omits GT-required relations, locating residual burden at policy-level construction rather than execution.

\vspace{-2pt}
\vspace{-2pt}

Evaluating SGVR's \texttt{solution\_FL} predicates with local Newclid~\cite{Newclid} gives $\mathrm{CR}{=}100.0\%$, $\mathrm{SR}{=}95.94\%$ (premise / num.\ / derived $=96.85/97.33/94.78$), and $\mathrm{SC}{=}84.0\%$ (Table~\ref{tab:process-fidelity}); SGVR's Gemini-2.5-Pro baseline under a gpt-5-nano LLM judge reports $\mathrm{SR}{=}88.7\%$, $\mathrm{SC}{=}44.5\%$~\cite{GeoGoal}. Under the stricter engine verifier, Draw2Think's SC rises by $\sim$40 points, showing that typed-operator construction survives exact predicate checking rather than relying on semantic lenience.

\subsection{Tolerance and Residual Error Decomposition}
\label{app:geogoal-tol-sweep}
\label{app:geogoal-failure-decomp}

Sweeping query-level tolerance separates structure from coordinate precision: structural targets ($T_i\in\{0,1,90,180\}$, 6{,}551 items) hold an $\sim$88\% plateau across five orders of magnitude; numerical targets (844 items) rise from 51\% at $10^{-4}$ to 94\% at $3{\times}10^{-2}$, approaching Newclid's $\sim$0.1\% relative tolerance (Figure~\ref{fig:tol-sweep}). Table~\ref{tab:geogoal-failure-decomp} classifies residual failures by error scale and expression type.

\vspace{-2pt}
\begin{table}[h]
\centering
\caption{$T_i$ failure decomposition on GeoGoal (7{,}395 queries). C1 = structural-scale error; C2 = coordinate-precision error; C3 = negligible residual.}
\label{tab:geogoal-failure-decomp}
\small
\setlength\tabcolsep{5pt}
\begin{tabular}{@{}l r r r l@{}}
\toprule
& \textbf{All} & \textbf{Faithful} & \textbf{Unfaithful} & \textbf{Dominant expr type} \\
& (N${=}$7395) & (IR${=}$T, 6216) & (IR${=}$F, 1179) & among failures \\
\midrule
OK                              & 87.48\% & 92.87\% & 59.03\% & --- \\
\midrule
C1 structural                   &  6.45\% &  3.06\% & 24.34\% & angle expr (97\%) \\
C2 precision                    &  2.46\% &  1.59\% &  7.04\% & length expr (100\%) \\
C3 tight                        &  0.00\% &  0.00\% &  0.00\% & --- \\
NA (undefined point)            &  3.61\% &  2.48\% &  9.58\% & --- \\
\bottomrule
\end{tabular}
\end{table}

\noindent C1 failures in faithful canvases are mostly angle expressions (184/190) implying collinearity / parallelism beyond SGVR's explicit predicates (structural-understanding errors); C2 are all pure length queries (99/99) at $\sim$0.04\% relative error, consistent with 2--3 sig.-digit coordinates from the model. The absent C3 regime locates precision loss at coordinate input, not float64 engine arithmetic. Engine-exact verification thus separates structural from precision errors that an LLM judge collapses into one sub-goal failure.

\clearpage
\section{Mechanism Analysis}
\label{app:mechanism}

\subsection{Answer Source Distribution}
\label{app:answer-source}

\begin{figure}[H]
\centering
\includegraphics[width=\textwidth]{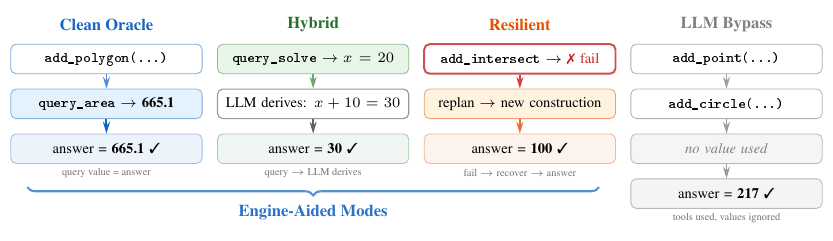}
\caption{Answer provenance taxonomy in Draw2Think. Clean Oracle reads an engine value; Hybrid derives from an engine-returned intermediate; Resilient recovers after a tool failure; LLM Bypass invokes tools while the final answer lacks an explicit anchor to returned values. The first three columns are grouped as engine-aided modes.}
\label{fig:answer-provenance}
\end{figure}

Figure~\ref{fig:answer-provenance} schematises the taxonomy and Table~\ref{tab:answer-source} classifies correct CT answers by answer provenance visible in the traces, separating observable anchors from internal causal attribution. \textbf{Clean Oracle}: the final answer is directly anchored to an engine-returned value. \textbf{Hybrid}: the answer combines engine-returned values with model reasoning. \textbf{Resilient}: the answer is recovered after intermediate tool failures. \textbf{LLM Bypass}: the final answer lacks a textual anchor to engine-returned values despite tool use.

\begin{table}[H]
\centering
\caption{Answer source distribution among answered problems (CT mode). Engine-Involved counts answers whose provenance includes at least one textual anchor to an engine-returned value.}
\label{tab:answer-source}
\footnotesize
\begin{tabular}{@{}lrrrrrr@{}}
\toprule
\textbf{Dataset} & \textbf{Pass/N} & \makecell{\textbf{Clean}\\\textbf{Oracle}} & \textbf{Hybrid} & \textbf{Resilient} & \makecell{\textbf{LLM}\\\textbf{Bypass}} & \makecell{\textbf{Engine-}\\\textbf{Involved}} \\
\midrule
\textit{PGPS9K}        & 905\,/\,1000 & 50.2 & 34.3 & 13.0 & 2.5  & \textbf{91.5} \\
\textit{UniGeo}        & 731\,/\,754  & 52.9 & 25.5 & 19.8 & 1.8  & \textbf{91.3} \\
\textit{Geo3K}         & 573\,/\,601  & 39.5 & 42.6 & 14.6 & 3.3  & \textbf{88.0} \\
\textit{MathVista}     & 204\,/\,208  & 25.5 & 51.0 & 19.6 & 3.9  & \textbf{82.4} \\
\textit{MathVerse}$^\dagger$  & 569\,/\,629  & 23.0 & 54.9 & 16.2 & 5.9  & \textbf{81.6} \\
\textit{GeoSketch}     & 335\,/\,390  & 19.9 & 56.4 & 21.3 & 2.4  & \textbf{80.9} \\
\textit{GeoLaux}       & 208\,/\,221  & 27.9 & 38.1 & 31.2 & 2.8  & \textbf{77.7} \\
\textit{OlympiadBench} & 100\,/\,112  & 23.8 & 41.9 & 25.7 & 8.6  & \textbf{74.3} \\
\bottomrule
\multicolumn{7}{@{}l}{\footnotesize $^\dagger$MathVerse combines Plane (510) and Solid (119) subsets.}
\end{tabular}
\end{table}

\subsection{Process Statistics}
\label{app:process}

\begin{table}[H]
\centering
\caption{Turn distribution across benchmarks (CT mode). Percentages are row-wise.}
\label{tab:turn-dist}
\small
\begin{tabular}{@{}lrrrrrrr@{}}
\toprule
\textbf{Dataset} & \textbf{N} & \textbf{1} & \textbf{2} & \textbf{3} & \textbf{4} & \textbf{5+} & \textbf{Mean} \\
\midrule
GeoSketch & 390  & 5\% & 33\% & 30\% & 18\% & 14\% & 3.3 \\
MathVerse Plane & 510  & 0\% & 66\% & 22\% & 8\% & 4\% & 2.5 \\
GeoLaux   & 221 & 1\% & 72\% & 20\% & 5\% & 2\% & 2.4 \\
MathVista GPS & 208 & 0\% & 80\% & 14\% & 4\% & 2\% & 2.3 \\
GeoQA     & 754  & 3\% & 77\% & 15\% & 4\% & 1\% & 2.2 \\
PGPS9K    & 1000 & 7\% & 71\% & 18\% & 4\% & 1\% & 2.2 \\
Geo3K     & 601  & 2\% & 85\% & 10\% & 2\% & 1\% & 2.1 \\
\midrule
SolidGeo-hard & 177 & 9\% & 26\% & 35\% & 19\% & 11\% & 3.0 \\
MathVerse Solid & 119  & 2\% & 77\% & 18\% & 3\% & 0\% & 2.2 \\
\bottomrule
\end{tabular}
\end{table}

Table~\ref{tab:turn-dist} reports the turn distribution across benchmarks. On the textbook-level benchmarks (\textit{PGPS9K}, \textit{Geo3K}, \textit{GeoQA}/\textit{UniGeo}, \textit{MathVista}, \textit{MathVerse}), 66--85\% of problems complete in exactly two turns. \textit{GeoSketch} (avg 3.3 turns) and \textit{SolidGeo-hard} (avg 3.0 turns) shift the distribution toward longer trajectories, reflecting auxiliary construction and 3D competition-level reasoning respectively.

\begin{figure}[h]
\centering
\includegraphics[width=\linewidth]{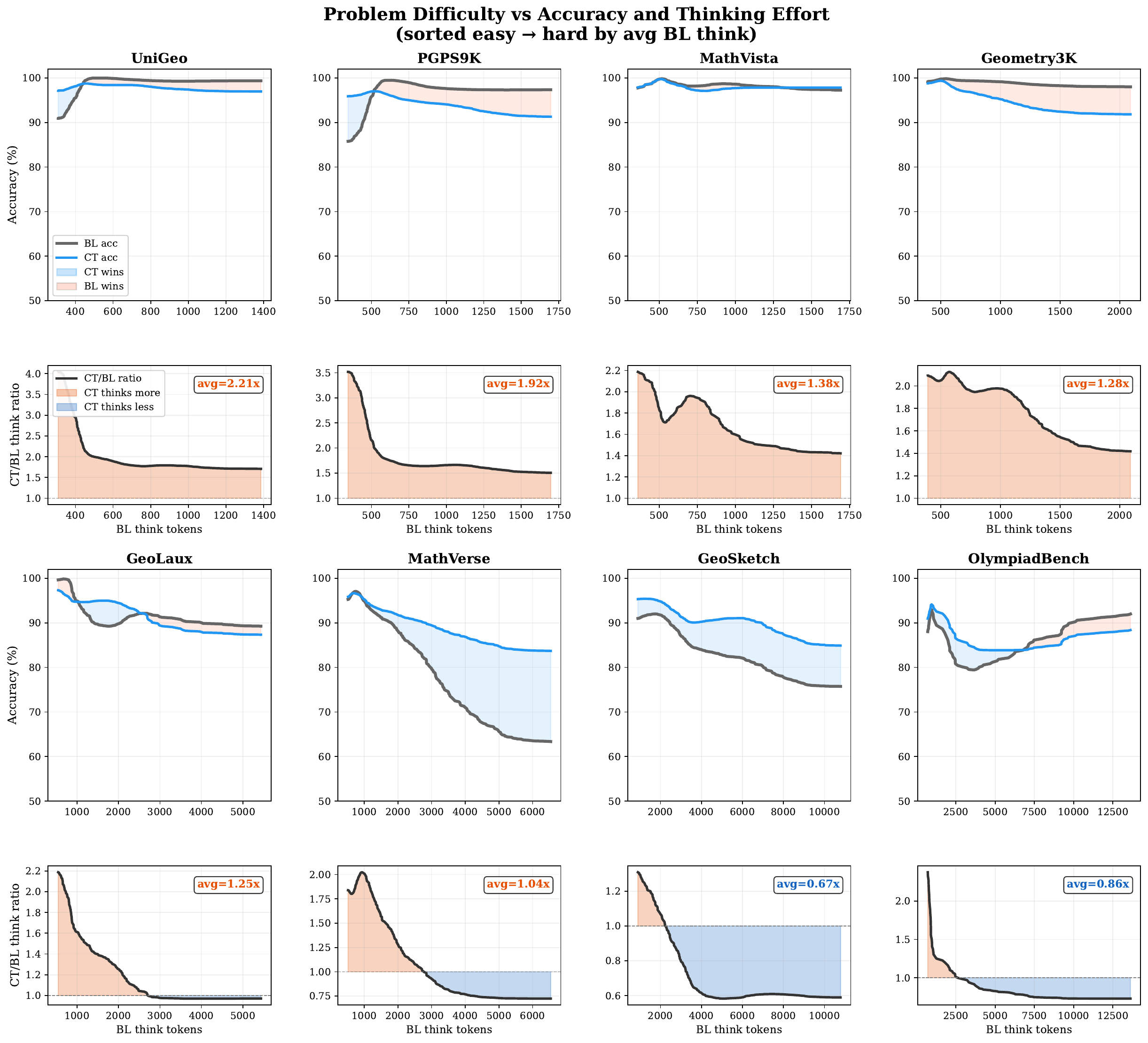}
\caption{Per-problem accuracy and thinking effort vs.\ BL thinking budget across eight benchmarks (sorted by average BL thinking tokens). \textbf{Top}: rolling BL/CT accuracy; blue fill = CT wins, red = BL wins. \textbf{Bottom}: CT/BL thinking ratio; orange fill = CT thinks more, blue fill = CT thinks less. The ratio shifts from $>$1 on low-BL-thinking benchmarks to $<$1 on high-BL-thinking benchmarks, consistent with a larger thinking-substitution effect as baseline thinking budget rises.}
\label{fig:think-difficulty}
\end{figure}

\paragraph{Token efficiency and wall time.}
BL is one direct model call; CT sums all engine-mediated turns in one construction trajectory. Table~\ref{tab:token-efficiency} shows generated tokens and wall time. CT thinking overhead shrinks as BL thinking budget rises: from $+121\%$ on \textit{GeoQA} to $-36\%$ on \textit{GeoSketch} and $-14\%$ on \textit{OlympiadBench}. Output tokens often drop because measurement and computation move into the engine. The remaining overhead is mostly repeated ToolSpec/canvas exposure (${\sim}$35--64K input tokens per problem), which motivates adaptive schema disclosure (\S\ref{sec:future}). Figure~\ref{fig:think-difficulty} gives the per-problem view.

\paragraph{Wall-time profile under the 120\,s per-call cap.}
\label{app:walltime-profile}

\begin{table}[H]
    \centering
    \caption{Per-problem generated tokens and wall time (mean, excluding timeouts). Format: BL\,$\rightarrow$\,CT\;\grayratio{ratio}; bold = CT uses fewer resources. Model: \texttt{gemini-3-flash-preview@medium} ($^\ddagger$\textit{OlympiadBench}: \texttt{@high}); $^\dagger$\textit{MathVerse} combines Plane+Solid.}
    \label{tab:token-efficiency}
    \small
    \setlength\tabcolsep{2.5pt}
    \begin{tabular}{@{}l r@{$\;\rightarrow\;$}r@{\;}l @{\quad} r@{$\;\rightarrow\;$}r@{\;}l @{\quad} r r r @{}}
    \toprule
    & \multicolumn{3}{c}{\textbf{Thinking tokens}} & \multicolumn{3}{c}{\textbf{Output tokens}} & \multicolumn{3}{c}{\textbf{Wall time (s)}} \\
    \cmidrule(lr){2-4} \cmidrule(lr){5-7} \cmidrule(lr){8-10}
    \textbf{Dataset} & \multicolumn{1}{r}{\scriptsize BL} & \multicolumn{1}{r}{\scriptsize CT} & & \multicolumn{1}{r}{\scriptsize BL} & \multicolumn{1}{r}{\scriptsize CT} & & \multicolumn{1}{r}{\scriptsize BL} & \multicolumn{1}{r}{\scriptsize CT-tot} & \multicolumn{1}{r}{\scriptsize CT-pt} \\
    \midrule
    \textit{GeoQA/UniGeo} &   808 & 1{,}785 & \grayratio{2.21}  & 335 & 425 & \grayratio{1.27}                              &  8.8 & 23.5 &  8.6 \\
    \textit{PGPS9K}     & 1{,}043 & 1{,}566 & \grayratio{1.50}  & 365 & 301 & {\color{gray}\scriptsize\textbf{$\times$0.82}} & 10.5 & 22.5 & 12.9 \\
    \textit{MathVista}  & 1{,}174 & 1{,}629 & \grayratio{1.39}  & 373 & 399 & \grayratio{1.07}                              & 10.2 & 18.5 &  6.7 \\
    \textit{Geo3K}      & 1{,}485 & 1{,}878 & \grayratio{1.26}  & 374 & 306 & {\color{gray}\scriptsize\textbf{$\times$0.82}} & 11.2 & 20.5 &  8.4 \\
    \textit{GeoLaux}    & 2{,}301 & 2{,}872 & \grayratio{1.25}  & 439 & 538 & \grayratio{1.23}                              & 17.0 & 28.4 &  9.7 \\
    \textit{MathVerse}$^\dagger$ & 2{,}540 & 2{,}637 & \grayratio{1.04}  & 392 & 327 & {\color{gray}\scriptsize\textbf{$\times$0.83}} & 17.7 & 25.6 & 10.3 \\
    \textit{GeoSketch}  & 5{,}498 & 3{,}545 & {\color{gray}\scriptsize\textbf{$\times$0.64}} & 648 & 692 & \grayratio{1.07} & 32.8 & 32.5 & 10.1 \\
    \textit{OlympiadBench}$^\ddagger$ & 5{,}884 & 5{,}063 & {\color{gray}\scriptsize\textbf{$\times$0.86}} & 632 & 934 & \grayratio{1.48} & 38.5 & 39.3 & 15.0 \\
    \bottomrule
    \end{tabular}
\end{table}

\begin{figure}[H]
    \centering
    \includegraphics[width=0.8\linewidth]{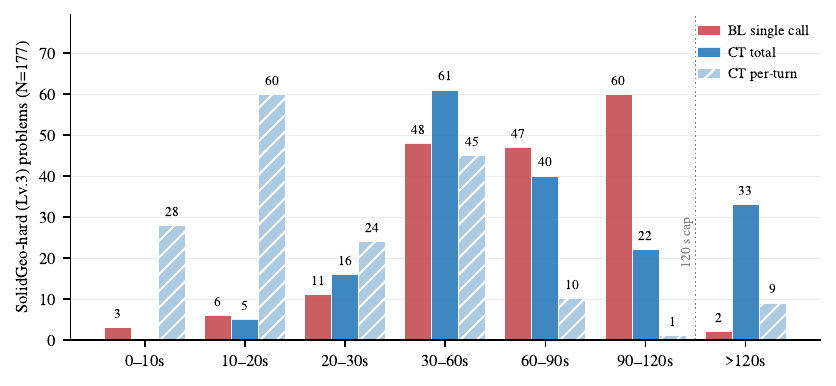}
    \caption{Wall-time distribution for Gemini-3-Flash@high on \textit{SolidGeo-hard} ($N{=}177$). BL (single call, hard-capped at 120\,s), CT total (multi-turn sum), and CT per-turn ($t_{\text{final}}/\text{eff}_T$). Median: BL $72.5$\,s, CT total $63.1$\,s, CT per-turn $20.1$\,s.}
    \label{fig:walltime-lv3}
\end{figure}

Figure~\ref{fig:walltime-lv3} reports \textit{SolidGeo-hard} under the same 120\,s per-call cap. CT's median total time (63.1\,s) is below BL's (72.5\,s); the 33 CT attempts exceeding 120\,s reflect multi-turn accumulation at the problem level, while every individual turn stays within the cap.

\paragraph{Tool usage temporal structure.}

\begin{figure}[H]
    \centering
    \includegraphics[width=\linewidth]{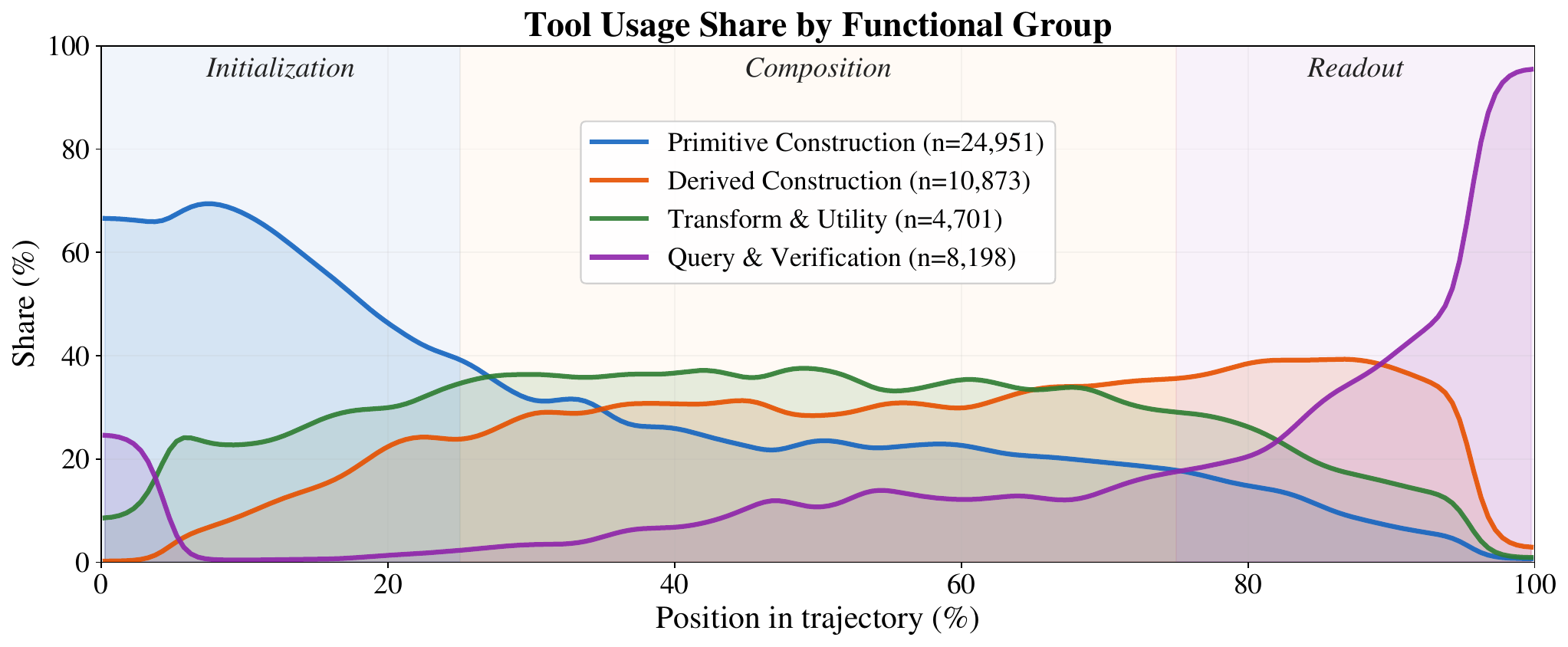}
\caption{Tool usage share by functional group over normalized trajectory position. Early calls concentrate on primitive construction; the final quarter concentrates on query and verification tools.}
    \label{fig:tool-trajectory}
\end{figure}

Figure~\ref{fig:tool-trajectory} groups 50{,}025 tool calls into Primitive Construction, Derived Construction, Transform \& Utility, and Query \& Verification, then plots each group's share over normalized call position $p=i/(K{-}1)$.

The phase split motivates progressive tool-schema disclosure. Initialization (0--25\%) is primitive-heavy (${\sim}$65\%); readout (75--100\%) is query-heavy (${\sim}$80\%). A staged schema could reduce the active action space during early construction while preserving the tools needed for that phase.\footnote{Google Gemini recommends keeping active tools to 10--20 where possible; OpenAI supports deferred loading via \texttt{tool\_search}; Anthropic recommends consolidating related operations.}

\subsection{Cross-Model Validation}
\label{app:cross-model}

Table~\ref{tab:cross-model-detail} reports BL and CT accuracy on \textit{OlympiadBench} across three frontier VLMs released in the past six months (December 2025 -- present). Generated-token columns report thinking and output tokens for the matched BL and CT runs; costs are reported separately for the two modes. The token pattern is model-specific: Gemini reduces thinking under CT ($5{,}884{\rightarrow}5{,}063$), whereas Claude Sonnet 4.6 and GPT-5.4 spend more generated tokens in exchange for higher accuracy. CT gains range from $+0.0\%$ to $+5.4\%$, providing a benchmark-level portability check across providers while showing that the cost--accuracy tradeoff varies with model tool-use competence.

\begin{table}[H]
\centering
\caption{Cross-model BL and CT comparison on \textit{OlympiadBench} ($N{=}112$, Pass@1). Think, Out, and Turns are per-problem averages; CT Turns counts engine-mediated turns in the construction trajectory. Costs are for the matched BL and CT runs: Gemini is computed from logged tokens at list price, Claude from platform billing, and GPT as an upper-bound estimate.}
\label{tab:cross-model-detail}
\footnotesize
\setlength{\tabcolsep}{1.5pt}
\begin{tabular}{@{}lrrrr@{\hspace{0.8em}}rrrrrr@{}}
\toprule
\textbf{Model} & \multicolumn{4}{c}{\textbf{BL}} & \multicolumn{6}{c}{\textbf{CT}} \\
\cmidrule(lr){2-5} \cmidrule(lr){6-11}
& \textbf{Acc} & \textbf{Think} & \textbf{Out} & \textbf{Cost} & \multicolumn{2}{c}{\textbf{Acc}} & \textbf{Think} & \textbf{Out} & \makecell{\textbf{Turns}\\\textbf{/prob.}} & \textbf{Cost} \\
\midrule
Gemini-3-Flash@high  & 89.3\% & 5{,}884 &   632 & \$2.3 & 89.3\% & \graydelta{+0.0} & 5{,}063 &   934 & 2.7 & \$4.8\textsuperscript{a} \\
Claude-Sonnet-4.6@high       & 89.3\% & 1{,}523 & 5{,}274 & \$9.1 & 92.0\% & \graydelta{$\uparrow$2.7} & 2{,}455 & 7{,}528 & 6.4 & \$26.0\textsuperscript{b} \\
GPT-5.4@high                 & 89.3\% & 1{,}123 & 1{,}437 & \$4.6 & 94.6\% & \graydelta{$\uparrow$5.4} & 2{,}206 & 2{,}842 & 8.2 & \$40.5\textsuperscript{c} \\
\bottomrule
\multicolumn{11}{@{}l}{\scriptsize\textsuperscript{a}Logged-token list-price estimate; BL+CT total US\$7.0; actual billing may be lower with cache.}\\
\multicolumn{11}{@{}l}{\scriptsize\textsuperscript{b}Platform bill with \texttt{cache\_control}; BL+CT total US\$35.1.}\\
\multicolumn{11}{@{}l}{\scriptsize\textsuperscript{c}Upper-bound estimate with \texttt{previous\_response\_id}; BL+CT total US\$45.1.}
\end{tabular}
\end{table}

\subsection{Outcome Transition Analysis}
\label{app:outcome-transition}

Table~\ref{tab:outcome-transition} decomposes accuracy into per-problem Save/Break transitions. Overall, CT rescues 133 BL failures and introduces 128 regressions (net $+5$). Saves dominate on construction-heavy or visually grounded benchmarks (\textit{GeoQA/UniGeo}, \textit{MathVerse}, \textit{GeoSketch}); breaks dominate on saturated benchmarks (\textit{PGPS9K}, \textit{Geo3K}), consistent with the ceiling-effect interpretation in Table~\ref{tab:main-results}. Save cases are often no-response or high-BL-thinking failures; Break cases concentrate where BL already has a cheap internal route.

\begin{table}[H]
\centering
\caption{Per-problem outcome transition between BL and CT (same model, same problem). Sorted by win rate (Save\,/\,Break) descending.}
\label{tab:outcome-transition}
\small
\begin{tabular}{@{}lrrrrrrr@{}}
\toprule
\textbf{Dataset} & \textbf{N} & \textbf{BL\checkmark CT\checkmark} & \makecell{\textbf{Save}\\{\scriptsize BL\ding{55}$\to$CT\checkmark}} & \makecell{\textbf{Break}\\{\scriptsize BL\checkmark$\to$CT\ding{55}}} & \textbf{BL\ding{55} CT\ding{55}} & \textbf{Net} & \makecell{\textbf{Win}\\{\scriptsize Save/Break}} \\
\midrule
\textit{GeoQA/UniGeo} & 754  & 691 & 40 & 15 &   8 & $+$25 & 2.7 \\
\textit{MathVerse}    & 510  & 435 & 29 & 11 &  35 & $+$18 & 2.6 \\
\textit{GeoSketch}    & 390  & 309 & 26 & 10 &  45 & $+$16 & 2.6 \\
\textit{MathVista}    & 208  & 199 &  4 &  3 &   2 & $+$1  & 1.3 \\
\textit{GeoLaux}      & 221  & 197 & 11 &  9 &   4 & $+$2  & 1.2 \\
\textit{OlympiadBench}& 112  &  96 &  4 &  4 &   8 & $+$0  & 1.0 \\
\textit{Geo3K}        & 601  & 567 &  6 & 23 &   5 & $-$17 & 0.3 \\
\textit{PGPS9K}       & 1000 & 892 & 13 & 53 &  42 & $-$40 & 0.2 \\
\midrule
\textbf{Total}        & 3796 & 3386& 133& 128& 149 & $+$5  & 1.0 \\
\bottomrule
\end{tabular}
\end{table}


\subsection{Tool Failure Taxonomy}
\label{app:failure-taxonomy}

The tool layer succeeds on ${\sim}$95\% of calls. We categorize the remaining failed calls from the five textbook-level solving benchmarks by root cause; percentages in Table~\ref{tab:failure-taxonomy} are over failed calls, not problems. Many affected problems still recover through construction-path diversity and state-preserving replanning (\S\ref{sec:recovery}).

\begin{table}[h]
\centering
\caption{Taxonomy of intermediate tool-call failures observed during the solving process. These are \emph{process-level} events; problems with such failures frequently still reach correct final answers via recovery (\S\ref{sec:recovery}). \emph{Layer} indicates root cause origin. Percentages are relative to total failed calls, not total problems.}
\label{tab:failure-taxonomy}
\footnotesize
\begin{tabularx}{\linewidth}{@{}l c r >{\raggedright\arraybackslash}X@{}}
\toprule
\textbf{Category} & \textbf{Layer} & \textbf{Share} & \textbf{Typical manifestation} \\
\midrule
Entity not found          & Model   & ${\sim}$35\% & References \texttt{segment\_AB} before it exists (only A, B defined) \\
Geometric precondition    & Model   & ${\sim}$19\% & \texttt{Intersect(L1, L2)} on non-intersecting lines \\
Degenerate input          & Model   & ${\sim}$5\%  & \texttt{Line(P, P)}: identical points to a 2-point constructor \\
Type mismatch             & Model   & ${\sim}$3\%  & Point passed where a curve is expected \\
\midrule
\textbf{Model-layer total} &         & ${\sim}$\textbf{62\%} & \\
\cmidrule(r){1-4}
Silent cascade            & Interface & ${\sim}$37\% & Downstream call fails on undefined parent; no signal returned \\
\cmidrule(r){1-4}
Runtime error             & Engine  & ${\sim}$1\%  & GeoGebra JS exception or CAS solver failure \\
\bottomrule
\end{tabularx}
\end{table}

\paragraph{Main failure modes.}
\label{app:entity-tracking}
Entity-not-found errors dominate model-layer failures: the model assumes an object exists when the canvas lacks it, e.g., treating points A,B as if \texttt{segment\_AB} had been created, predicting names that differ from GeoGebra's internal labels, or continuing after a failed \texttt{add\_intersect}. Geometric-precondition errors are second: the object references are valid, while the requested relation is absent from the current canvas. A small subset reflects engine tolerance, such as a line--circle intersection recovered by indexed retry; most reflect unsupported assumptions about parallelism, tangency, or intersection. Degenerate-input and type errors isolate parameter generation: the model selects the right operation but binds invalid operands, as in \texttt{Line(F,F)} or \texttt{add\_circle(C,)}.

\paragraph{Cascade and recovery.}
\label{app:cascade-recovery}
We observe 55 cascade chains, usually after a failed \texttt{add\_intersect}: the model issues 2--8 dependent calls to the undefined object, then replans once the turn-level observation exposes the missing parent. Despite these cascades, 70\% of affected problems still reach the correct answer because previously valid objects remain on the canvas.

\paragraph{Implications for Harness Design.}
\label{app:failure-implications}

The engineering targets are direct: existence checks for missing objects, feasibility checks before geometric constructors, and explicit diagnostics for silent cascades. Since 70\% of cascade-affected problems already recover, better diagnostics mainly reduce wasted turns.

\subsection{Mechanism Ablations}
\label{app:mechanism-ablation}

\subsubsection{Delete Ablation: Conditional Value of Branch Retraction}
\label{app:delete-ablation}

We remove \texttt{delete\_object} from the action space and rerun the 86 problems across the five textbook-level benchmarks that originally invoked it. Table~\ref{tab:delete-cat} classifies the original delete usage into four patterns; Table~\ref{tab:delete-result} reports per-category ablation results.

\begin{table}[h]
\centering
\caption{Original delete usage patterns on the 86 problems. Category is determined by delete count and co-occurrence with tool failures.}
\label{tab:delete-cat}
\small
\begin{tabular}{@{}llrrr@{}}
\toprule
\textbf{Category} & \textbf{Characterization} & \textbf{N} & \textbf{Avg del} & \textbf{Avg tools} \\
\midrule
A: Light cleanup & $\leq$2 deletes, 0 tool failures & 38 & 1.4 & 16 \\
B: Error repair & Delete after a tool failure & 18 & 1.9 & 23 \\
C: Strategy reset & 3--4 deletes, no prior failure & 19 & 3.3 & 21 \\
D: Heavy rollback & $\geq$5 deletes & 11 & 7.9 & 32 \\
\bottomrule
\end{tabular}
\end{table}

\begin{table}[h]
\centering
\caption{Per-category ablation results after removing \texttt{delete\_object}. $\Delta$tools is the relative change in average tool calls.}
\label{tab:delete-result}
\small
\begin{tabular}{@{}lrrrrr@{}}
\toprule
\textbf{Category} & \textbf{N} & \textbf{Orig acc} & \textbf{Ablat acc} & \textbf{$\Delta$acc} & \textbf{$\Delta$tools} \\
\midrule
A: Light cleanup & 38 & 95\% & 95\% & 0 & $-$17\% \\
B: Error repair & 18 & 89\% & 94\% & +1 & $-$20\% \\
C: Strategy reset & 19 & 95\% & 89\% & $-$1 & $-$34\% \\
D: Heavy rollback & 11 & 91\% & 100\% & +1 & $-$46\% \\
\midrule
\textbf{Overall} & \textbf{86} & \textbf{93.0\%} & \textbf{94.2\%} & \textbf{+1.2\%} & \textbf{$-$29\%} \\
\bottomrule
\end{tabular}
\end{table}

\paragraph{Key findings.}

\noindent\emph{Delete has conditional value.} Removing \texttt{delete\_object} slightly raises aggregate accuracy (93.0\%$\to$94.2\%) and reduces tool calls in every category (17--46\%), showing that many deletes are cleanup or over-exploration rather than necessary repair. Seven of 86 problems flip outcome (3 correct$\to$wrong, 4 wrong$\to$correct), and 63 of the 79 stable cases use fewer tools. The boundary appears in the paired cases: deletion repairs one paper-folding construction and recovers $110^\circ$, whereas removing deletion prevents a rotation-area rebuild loop and restores the correct answer. Delete is therefore an aggressive revisability primitive. It retracts canvas branches more cleanly than textual self-correction, while invalid-action rejection and state-preserving replanning account for much of the PDV loop's practical revisability.

\subsubsection{Query Ablation: Answer Source and Trajectory Migration}
\label{app:ablation-details}

Table~\ref{tab:ablation}a in the main text reports aggregate accuracy and cost changes under two query ablation levels. Here we break down the adaptation mechanism along three dimensions on the same N\,=\,379 matched subset.

\paragraph{Answer source migration.}
Table~\ref{tab:ablation-source} classifies each problem's final answer by its derivation channel.
Under Full CT, 44.6\% of correct answers are \emph{clean oracle}: the final value matches a query return directly.
Under ablation, this channel nearly vanishes; the model compensates primarily through \emph{construction-based derived readout} (classified as ``constr.\ readout'' below): tools such as \texttt{add\_angle} and \texttt{add\_distance} return exact values as a side effect of construction, and the model uses these instead of the blocked query interface.

\begin{table}[H]
\centering
\caption{Answer source distribution under query ablation (N\,=\,379). ``Constr.\ readout'' = the model invoked construction tools but obtained values from their return fields rather than explicit query calls.}
\label{tab:ablation-source}
\small
\begin{tabular}{@{}lrrr@{}}
\toprule
\textbf{Answer source} & \textbf{Full CT} & \textbf{w/o meas} & \textbf{w/o query} \\
\midrule
Clean oracle         & 44.6\% &  0.3\% &  1.1\% \\
Hybrid               & 31.9\% & 16.9\% & 10.3\% \\
Resilient (oracle)   &  9.8\% &  0.3\% &  0.8\% \\
Resilient (fallback) & 11.1\% & 33.0\% & 29.3\% \\
Constr.\ readout     &  1.3\% & 43.5\% & 52.2\% \\
\bottomrule
\end{tabular}
\end{table}

\paragraph{Trajectory structure migration.}
Table~\ref{tab:ablation-trajectory} reports the average share of tool calls by functional group (matching the four groups in Figure~\ref{fig:tool-trajectory}).
Query-phase calls drop from 16.5\% to 4.6\%, while derived-construction calls rise from 21.9\% to 33.2\%. Readout functionality migrates from the dedicated query phase into the construction trajectory.

\begin{table}[H]
\centering
\caption{Tool-group share (\% of tool calls per problem) under query ablation.}
\label{tab:ablation-trajectory}
\small
\begin{tabular}{@{}lrrrr@{}}
\toprule
\textbf{Tool group} & \textbf{Full CT} & \textbf{w/o meas} & \textbf{w/o query} \\
\midrule
Primitive Construction & 53.4\% & 48.7\% & 49.7\% \\
Derived Construction   & 21.9\% & 30.5\% & 33.2\% \\
Transform \& Utility   &  7.9\% &  8.9\% &  8.5\% \\
Query \& Verification  & 16.5\% &  6.8\% &  4.6\% \\
\bottomrule
\end{tabular}
\end{table}

\paragraph{Tool substitution pattern.}
Table~\ref{tab:ablation-substitution} shows the dominant 1:1 substitution pairs.
The near-perfect mirror between blocked query tools and their derived counterparts (\texttt{query\_angle}\,$\leftrightarrow$\,\texttt{add\_angle}, etc.) indicates a systematic substitution pattern in the model's adaptation.
Under \textbf{w/o meas}, \texttt{query\_solve} (CAS) surges from 13 to 226 calls as an additional escape channel; under \textbf{w/o query}, inspection via \texttt{query\_is\_defined} rises from 0 to 295 as the model uses object-existence checks in place of value queries.

\begin{table}[H]
\centering
\caption{Tool call counts on the matched 379-problem subset. Blocked tools are marked with \textdagger.}
\label{tab:ablation-substitution}
\small
\begin{tabular}{@{}lrrrr@{}}
\toprule
\textbf{Tool} & \textbf{Full CT} & \textbf{w/o meas} & \textbf{w/o query} \\
\midrule
\texttt{query\_angle}\textsuperscript{\textdagger}    & 318 &  59 &  47 \\
\texttt{query\_distance}\textsuperscript{\textdagger}  & 206 &  41 &  30 \\
\texttt{query\_area}\textsuperscript{\textdagger}      &  45 &   3 &   0 \\
\midrule
\texttt{add\_angle}      &  11 & 288 & 317 \\
\texttt{add\_distance}   &   9 & 171 & 184 \\
\texttt{add\_area}       &   0 &  38 &  42 \\
\midrule
\texttt{query\_solve}    &  13 & 226 &  29 \\
\texttt{query\_is\_defined} &   0 &  53 & 295 \\
\bottomrule
\end{tabular}
\end{table}

\subsubsection{Description Ablation: Bare Signatures and Selection Guidance}
\label{app:desc-ablation}

We strip natural-language descriptions from the 50 construction and transform ToolSpecs, retaining function names and typed parameter schemas (``bare'' level). Query tools and their descriptions are kept intact, so the model still has access to the same measurement interface; the ablation removes the natural-language guidance on \emph{when and how} to construct.

Table~\ref{tab:desc-ablation} reports per-benchmark results on the matched subset (N\,=\,396). Three observations stand out:

\begin{table}[h]
\centering
\caption{Description ablation: bare signatures vs.\ full ToolSpec descriptions (N\,=\,396, four benchmarks). Fail rate = fraction of tool calls returning an error. LLM Bypass = tool-using problems whose final answers lack an explicit anchor to a \texttt{query\_*} return value.}
\label{tab:desc-ablation}
\small
\begin{tabular}{@{}lcccccccccc@{}}
\toprule
& & & & & & \multicolumn{2}{c}{\textbf{Fail rate}} & \multicolumn{2}{c}{\textbf{Bypass}} \\
\cmidrule(lr){7-8} \cmidrule(lr){9-10}
\textbf{Benchmark} & \textbf{N} & \textbf{BL} & \textbf{Full} & \textbf{Bare} & \textbf{$\Delta$} & \textbf{Full} & \textbf{Bare} & \textbf{Full} & \textbf{Bare} \\
\midrule
\textit{UniGeo}    & 110 & 93.6 & 97.3 & 97.3 & \graydelta{0.0}    & 5.4 & 5.7 & 2 &  9 \\
\textit{PGPS9K}    & 119 & 89.9 & 96.6 & 95.8 & \graydelta{$-$0.8} & 4.3 & 7.1 & 3 & 11 \\
\textit{MathVerse} &  67 & 86.6 & 88.1 & 82.1 & \graydelta{$-$6.0} & 5.1 & 9.0 & 0 &  3 \\
\textit{GeoLaux}   & 100 & 95.0 & 96.0 & 94.0 & \graydelta{$-$2.0} & 7.4 & 11.3 & 2 &  9 \\
\midrule
\textbf{Total}     & 396 & 91.7 & 95.2 & 93.4 & \graydelta{$-$1.8} & 5.7 & 8.3 & 7 & 32 \\
\bottomrule
\end{tabular}
\end{table}

\noindent\emph{Interpretation.}
Under bare signatures, effort remains matched while routing quality drops: thinking tokens match ($\times$1.00), turn counts and tool calls stay nearly fixed, tool failures rise from 5.7\% to 8.3\%, and LLM Bypass increases from 7 to 32 correct answers. Failure rate measures weakened tool/parameter selection; bypass measures weakened readout anchoring. The effect is strongest on \textit{MathVerse} ($-6.0$\,pp, below BL), where visual construction guidance matters most, and absent on \textit{UniGeo}, whose textual constraints make function names sufficient.

\subsubsection{Towards Automatic Harness Refinement: A Per-Parameter Micro-Ablation}
\label{app:harness-micro}

\paragraph{Protocol.}
The description ablation above treats ToolSpec descriptions as a binary lever. This finer probe uses \textit{MathVerse}~3850 to test per-parameter runtime overlays without modifying the source catalog; the listed GT is 1.0, whereas the geometric truth is 7/6, so the signal is trajectory shape and budget rather than accuracy. All nine temperature-0 variants return 7/6. The unused-tool control \texttt{v8} sets a $-35\%$ local perturbation floor. CCW direction hints exceed that floor most strongly, with \texttt{v4\_sector} reducing thinking by 76\% in 28\,s. Combined edits are non-monotonic (\texttt{v7}: $-71\%$; \texttt{v6}: $-46\%$), indicating that ToolSpec wording changes the trajectory rather than composing as independent local effects.

\begin{table}[H]
\centering
\caption{Per-parameter description-refinement variants on \textit{MathVerse}~3850. Bold marks edited ToolSpec text; CCW = counterclockwise sweep. \(\Delta\) reports thinking-token change vs.\ \texttt{v0\_source}; the unused-tool control \texttt{v8} sets a \(-35\%\) perturbation floor.}
\label{tab:harness-microablation}
\includegraphics[width=\linewidth]{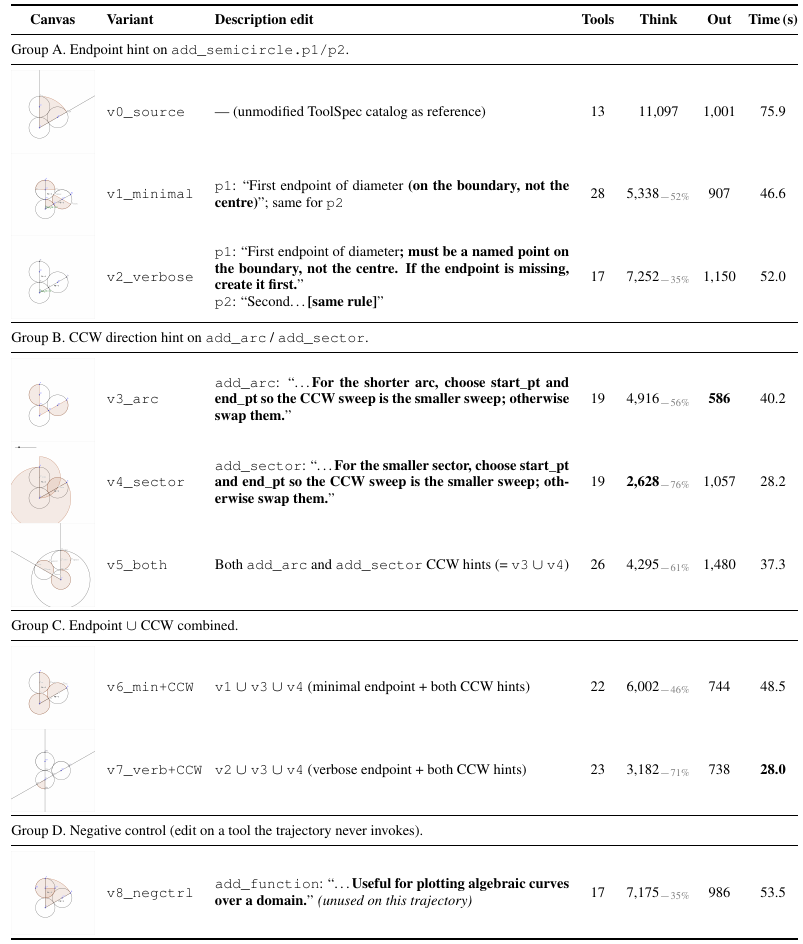}
\end{table}

\paragraph{Implication for harness refinement.}
Runtime overlays with schema checks and a negative-control floor give a practical recipe for automatic harness refinement: enumerate candidate per-parameter edits, evaluate them against an unedited baseline and an unused-tool control, and keep effects beyond the local floor. This complements Agentic Harness Engineering~\cite{AHE}: geometric ToolSpecs sit inside a cascaded construction DAG, so a local wording change can alter object dependencies, downstream readouts, and answer provenance.

\paragraph{Limitations.}
A single case study at temp~0, with claims read off the gap above the $-35\%$ negative-control floor; the GT bug isolates trajectory shape from accuracy but precludes any benchmark-level generalisation.

\clearpage
\section{GenExam-Math Rendering Evaluation Details}
\label{app:rendering}

\subsection{Detailed \textit{GenExam-Math} Results}
\label{app:genexam-detail}

Table~\ref{tab:genexam-leaderboard} places Draw2Think on the \textit{GenExam} public leaderboard. Math is consistently the weakest subject for leading closed-source T2I systems because precise geometric relations expose the limits of pixel-first synthesis, making it a natural target for \emph{constraint-first} rendering. Table~\ref{tab:genexam-math-detailed} adds our per-category and auxiliary-dimension breakdown.

\providecommand{\tsgrey}{\fontsize{5.5}{6.6}\selectfont}
\providecommand{\methodcell}[2]{\makecell[l]{#1\\[-3pt]\textcolor{gray}{\tsgrey #2}}}
\begin{table}[h]
\centering
\caption{\textit{GenExam} Math-subset leaderboard comparison (151 examples); official leaderboard at \url{https://github.com/OpenGVLab/GenExam\#-leaderboard}, snapshot as of April 23, 2026. T2I = text-to-image; MLLM = multimodal large language model. Base model shown in gray below each closed-source method; secondary per-vendor variants commented in the source for reference.}
\label{tab:genexam-leaderboard}
\small
\begin{minipage}[t]{0.48\linewidth}
\centering
\begin{tabular}{@{}lrr@{}}
\multicolumn{3}{@{}l}{\textbf{Closed-source Models}} \\
\toprule
\textbf{Method} & \textbf{Strict} & \textbf{Relaxed} \\
\midrule
\methodcell{\textbf{Draw2Think}}{Gemini 3 Flash Preview | Dec.2025} & \textbf{68.2} & \textbf{90.5} \\
\methodcell{Nano Banana 2}{Gemini 3.1 Flash Image | Feb.2026} & \underline{56.3} & \underline{87.8} \\
\methodcell{Nano Banana Pro}{Gemini 3 Pro Image | Nov.2025} & 55.6 & 86.3 \\
\methodcell{GPT-Image-2}{OpenAI native T2I | Apr.2026} & 50.3 & 85.2 \\
\methodcell{Seedream 5.0}{ByteDance native T2I | Feb.2026} & 47.0 & 82.9 \\
\methodcell{GPT-Image-1.5}{OpenAI native T2I | Dec.2025} & 26.5 & 65.8 \\
\methodcell{FLUX.2 max}{Black Forest Labs T2I | Dec.2025} & 6.6 & 49.1 \\
\bottomrule
\end{tabular}
\end{minipage}\hfill
\begin{minipage}[t]{0.48\linewidth}
\centering
\begin{tabular}{@{}lrr@{}}
\multicolumn{3}{@{}l}{\textbf{Open-source T2I / Unified MLLMs}} \\
\toprule
\textbf{Method} & \textbf{Strict} & \textbf{Relaxed} \\
\midrule
\multicolumn{3}{@{}c}{\textit{Open-source T2I Models}} \\
\midrule
FLUX.2 dev & 2.6 & 31.6 \\
Qwen-Image-2512 & 0.0 & 27.9 \\
HunyuanImage-3.0 & 0.0 & 17.0 \\
\midrule
\multicolumn{3}{@{}c}{\textit{Open-source Unified MLLMs}} \\
\midrule
BLIP3o-NEXT-GRPO & 0.0 & 15.5 \\
BAGEL & 0.0 & 14.7 \\
Janus-Pro & 0.0 & 13.7 \\
\midrule
\multicolumn{3}{@{}c}{\textit{Open-source Code-rendered RL}} \\
\midrule
\methodcell{Faire~\cite{Faire}}{Qwen3-VL-8B + GRPO} & 9.3 & 52.3 \\
\bottomrule
\end{tabular}
\end{minipage}
\end{table}

\begin{table}[h]
\centering
\caption{Detailed rendering results of our Draw2Think on the \textit{GenExam}~\cite{GenExam} Math subset (151 examples). In addition to strict and relaxed scores, we report semantic correctness, spelling, readability, and logical consistency from the official evaluator.}
\label{tab:genexam-math-detailed}
\small
\begin{tabular}{@{}lrrrrrrr@{}}
\toprule
\textbf{Category} & \textbf{N} & \textbf{Strict} & \textbf{Relaxed} & \textbf{Semantic} & \textbf{Spelling} & \textbf{Readability} & \textbf{Logic} \\
\midrule
Plane Geometry & 84 & 66.7 & 90.6 & 88.4 & 1.98/2 & 1.87/2 & 1.90/2 \\
Analytic Geometry & 56 & 80.4 & 95.5 & 94.3 & 2.00/2 & 1.95/2 & 1.95/2 \\
Solid Geometry (3D) & 11 & 18.2 & 63.6 & 59.0 & 1.82/2 & 1.45/2 & 1.18/2 \\
\midrule
Plane + Analytic (2D) & 140 & 72.1 & 92.6 & 90.6 & 1.99/2 & 1.90/2 & 1.92/2 \\
Overall & 151 & 68.2 & 90.5 & 88.5 & 1.97/2 & 1.87/2 & 1.87/2 \\
\bottomrule
\end{tabular}
\end{table}

\paragraph{Failure analysis (48 strict-fail problems).}
Strict requires \emph{all} scoring points correct and full marks on every auxiliary dimension (spelling, readability, logic); any shortfall yields \texttt{strict\_score\,=\,0}.

\begin{table}[h]
\centering
\caption{Root causes of strict-score failures on \textit{GenExam-math}. ``Semantic'' = missing scoring points; ``Quality'' = auxiliary dimensions (spelling, readability, logic) below threshold.}
\label{tab:genexam-failure}
\small
\begin{tabularx}{\linewidth}{@{}lr>{\raggedright\arraybackslash}X@{}}
\toprule
\textbf{Root cause} & \textbf{N} & \textbf{Typical mechanism} \\
\midrule
Semantic-only & 17 & Missing scoring points (unlabeled point, omitted auxiliary); quality passes. \\
Both semantic + quality & 14 & Missing elements \emph{plus} layout defects (label overlap, font-viewport mismatch). \\
Quality-only & 4 & All scoring points present; \texttt{read\,=\,1} (font-to-viewport) or \texttt{logic\,<\,2}. \\
Timeout ($>$120\,s) & 5 & Proxy/network latency; not a construction or model failure. \\
\midrule
\multicolumn{3}{@{}l}{\textbf{Severity distribution}} \\
\midrule
Critical (relaxed $<$ 0.6) & 6 & Major construction failure or missing primary geometric object. \\
Moderate (0.6--0.8) & 12 & Partial scoring point loss; core figure recognizable. \\
Near-pass ($>$ 0.8) & 12 & 1--2 auxiliary dimensions short; often \texttt{read\,=\,1} alone. \\
\bottomrule
\end{tabularx}
\end{table}

\noindent Of 151 problems, 103 reach \texttt{strict\_score\,=\,1} (Appendix~\ref{app:gallery}); 7 of the remaining 48 keep perfect semantic but lose strict on presentation alone (mostly \texttt{readability\,=\,1}), with recurring label overlap (``A,B,C,D'' near the origin) and viewport--font mismatch.

These quality-only failures mark a phase boundary in the harness: construction turns see textual canvas state, while render turns style a committed construction. Post-construction screenshots, tighter label-placement render-tool descriptions, or finer-grained construction actions could localise them earlier. The remaining 41 have \texttt{semantic\,$<$\,1.0}, mostly through omitted secondary elements (dashed auxiliaries, arc markers) over a recognisable primary structure.

\paragraph{Per-taxonomy breakdown.}
Table~\ref{tab:genexam-per-tax} reports per-taxonomy strict pass rate and average turn count split into \emph{construct} (canvas-building) and \emph{render} (styling) over the 37 official \textit{GenExam-math}~\cite{GenExam} taxonomies. Solid geometry is the clear bottleneck: 4 of 6 taxonomies fail every problem despite \texttt{thinking\_level=high} on 9 of 11 attempts. Within plane / analytic, turn count separates poorly from pass rate: \textit{Triangle/Perpendicular\_Bisector} passes at 16 render turns while \textit{Triangle/Similarity} fails at 9.5 construct turns, separating render-quality from construction-layer failure modes.

{\footnotesize
\setlength{\tabcolsep}{4pt}
\begin{longtable}{@{}>{\raggedright\arraybackslash}p{\dimexpr\linewidth-5.4cm\relax} r r r r r r@{}}
\caption{Per-taxonomy breakdown of \textit{GenExam-math} (151 problems, 37 taxonomies; \texttt{Mathematics/} prefix dropped). \textbf{N}: count; \textbf{s\!=\!1}: strict\_score=1; \textbf{Pass}: strict rate; \textbf{Total / $T_c$ / $T_r$}: average total / construct / render turns. Top: plane + analytic (31 rows); bottom: solid (6 rows). Purple = zero strict-1; blue = 100\% strict pass.}
\label{tab:genexam-per-tax} \\
\toprule
\textbf{Taxonomy} & \textbf{N} & \textbf{s=1} & \textbf{Pass} & \textbf{Total} & $\boldsymbol{T_c}$ & $\boldsymbol{T_r}$ \\
\midrule
\endfirsthead
\multicolumn{7}{@{}l}{\small\itshape (Table~\ref{tab:genexam-per-tax} continued)} \\
\toprule
\textbf{Taxonomy} & \textbf{N} & \textbf{s=1} & \textbf{Pass} & \textbf{Total} & $\boldsymbol{T_c}$ & $\boldsymbol{T_r}$ \\
\midrule
\endhead
\midrule
\multicolumn{7}{r}{\footnotesize\itshape Continued on next page} \\
\endfoot
\bottomrule
\endlastfoot
\rowcolor{purple!8} Plane\_Geometry/Triangle/Similarity                          &  2 &  0 &    0\% & 17.0 &  9.5 &  7.5 \\
\rowcolor{purple!8} Plane\_Geometry/Triangle/Others                              &  1 &  0 &    0\% &  5.0 &  3.0 &  2.0 \\
Plane\_Geometry/Angle                                        &  3 &  1 &   33\% & 19.3 & 12.0 &  7.3 \\
Plane\_Geometry/Complex\_Geometry\_Problem                   & 23 & 11 &   48\% & 11.6 &  7.4 &  4.2 \\
Plane\_Geometry/Circle/Others                                &  4 &  2 &   50\% &  8.5 &  4.5 &  4.0 \\
Analytic\_Geometry/Parametric\_Equation\_and\_Polar\_Curve   &  4 &  2 &   50\% &  5.2 &  4.2 &  1.0 \\
Analytic\_Geometry/Inequality\_Region/Linear\_Programming    &  2 &  1 &   50\% & 11.0 &  5.0 &  6.0 \\
Plane\_Geometry/Rectangle\_and\_Polygon/Rectangle            & 10 &  6 &   60\% & 10.6 &  5.4 &  5.2 \\
Analytic\_Geometry/Linear\_Function                          &  3 &  2 &   67\% &  6.7 &  4.0 &  2.7 \\
Plane\_Geometry/Rectangle\_and\_Polygon/Trapezoid            &  3 &  2 &   67\% &  8.0 &  3.3 &  4.7 \\
Analytic\_Geometry/Definite\_Integral\_Area                  & 12 &  9 &   75\% &  8.8 &  5.2 &  3.7 \\
Plane\_Geometry/Circle/Chord                                 &  8 &  6 &   75\% & 13.2 &  6.9 &  6.4 \\
Plane\_Geometry/Circle/Tangent                               &  4 &  3 &   75\% & 14.2 &  5.5 &  8.8 \\
Plane\_Geometry/Rectangle\_and\_Polygon/Other\_Quadrilateral &  5 &  4 &   80\% & 10.4 &  5.6 &  4.8 \\
Analytic\_Geometry/Quadratic\_Function                       &  6 &  5 &   83\% &  9.5 &  5.8 &  3.7 \\
Analytic\_Geometry/Other\_Function                           & 14 & 12 &   86\% &  8.4 &  5.2 &  3.1 \\
Analytic\_Geometry/Piecewise\_Function                       &  8 &  7 &   88\% & 10.1 &  5.4 &  4.8 \\
\rowcolor{blue!6}   Plane\_Geometry/Circle/Inscribed\_and\_Circumscribed\_Circle &  6 &  6 &  100\% & 12.5 &  6.7 &  5.8 \\
\rowcolor{blue!6}   Plane\_Geometry/Triangle/Right\_Triangle                     &  4 &  4 &  100\% & 13.8 &  5.5 &  8.2 \\
\rowcolor{blue!6}   Analytic\_Geometry/Trigonometric\_Function                   &  3 &  3 &  100\% & 12.3 &  6.3 &  6.0 \\
\rowcolor{blue!6}   Analytic\_Geometry/Exponential\_and\_Logarithmic\_Function   &  2 &  2 &  100\% & 13.0 &  5.5 &  7.5 \\
\rowcolor{blue!6}   Plane\_Geometry/Rectangle\_and\_Polygon/Pentagon             &  2 &  2 &  100\% & 19.0 &  9.0 & 10.0 \\
\rowcolor{blue!6}   Plane\_Geometry/Rectangle\_and\_Polygon/Other\_Polygon       &  2 &  2 &  100\% &  4.0 &  2.0 &  2.0 \\
\rowcolor{blue!6}   Plane\_Geometry/Rectangle\_and\_Polygon/Parallelogram        &  2 &  2 &  100\% & 10.5 &  6.0 &  4.5 \\
\rowcolor{blue!6}   Plane\_Geometry/Triangle/Perpendicular\_Bisector             &  1 &  1 &  100\% & 23.0 &  7.0 & 16.0 \\
\rowcolor{blue!6}   Plane\_Geometry/Triangle/Congruence                          &  1 &  1 &  100\% & 11.0 &  5.0 &  6.0 \\
\rowcolor{blue!6}   Plane\_Geometry/Triangle/Angle\_Bisector                     &  1 &  1 &  100\% & 13.0 &  7.0 &  6.0 \\
\rowcolor{blue!6}   Plane\_Geometry/Triangle/Altitude                            &  1 &  1 &  100\% & 10.0 &  7.0 &  3.0 \\
\rowcolor{blue!6}   Analytic\_Geometry/Geometric\_Meaning\_Of\_Derivative        &  1 &  1 &  100\% & 11.0 & 11.0 &  0.0 \\
\rowcolor{blue!6}   Analytic\_Geometry/Absolute\_Value\_Function                 &  1 &  1 &  100\% &  6.0 &  3.0 &  3.0 \\
\rowcolor{blue!6}   Plane\_Geometry/Rectangle\_and\_Polygon/Regular\_Hexagon     &  1 &  1 &  100\% & 14.0 &  5.0 &  9.0 \\
\midrule
\multicolumn{7}{@{}l}{\textit{Solid Geometry (6 taxonomies; 9 of 11 attempts under @H thinking)}} \\
\midrule
\rowcolor{purple!8} Solid\_Geometry/Section/Straight\_Cut                        &  2 &  0 &    0\% &  6.5 &  4.5 &  2.0 \\
\rowcolor{purple!8} Solid\_Geometry/Sphere/Tangent\_Plane                        &  1 &  0 &    0\% & 15.0 &  6.0 &  9.0 \\
\rowcolor{purple!8} Solid\_Geometry/Prism/Oblique\_Prism                         &  1 &  0 &    0\% & 14.0 &  8.0 &  6.0 \\
\rowcolor{purple!8} Solid\_Geometry/Prism/Right\_Prism                           &  1 &  0 &    0\% &  7.0 &  5.0 &  2.0 \\
Solid\_Geometry/Cylinder\_and\_Cone                          &  4 &  1 &   25\% & 13.2 &  7.5 &  5.8 \\
Solid\_Geometry/Pyramid/Regular\_Pyramid                     &  2 &  1 &   50\% & 13.0 &  6.5 &  6.5 \\
\end{longtable}
}

\clearpage
\subsection{GenExam-Math Cross-system Case Study}
\label{app:genexam-math44}

\begin{figure}[H]
\centering
\includegraphics[width=\linewidth]{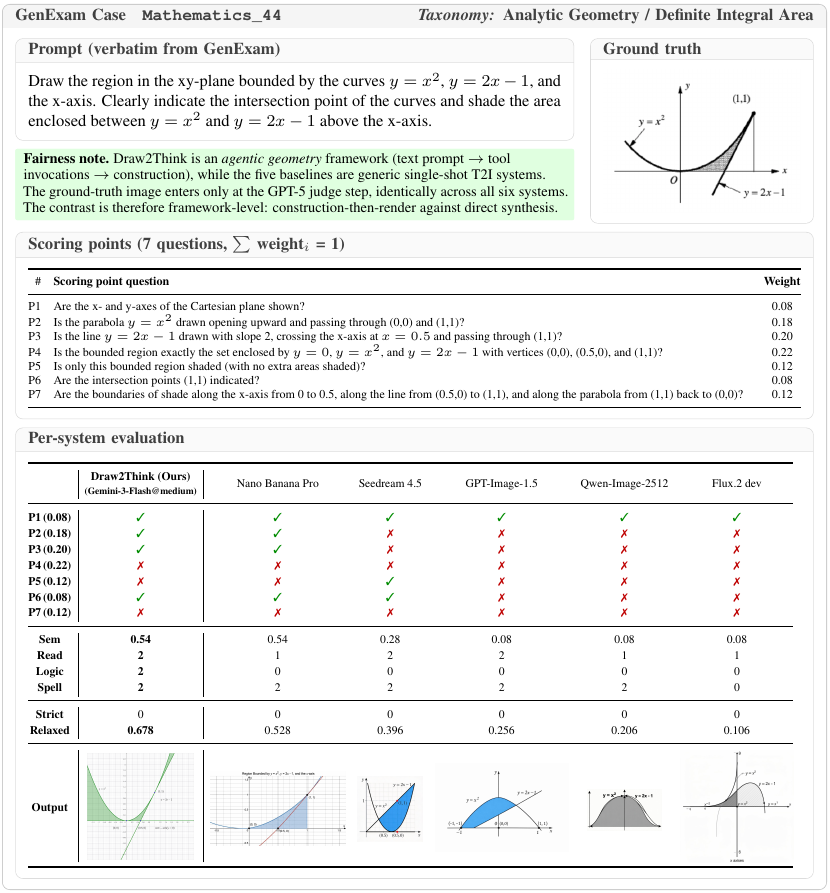}
\caption{Cross-system evaluation on GenExam \texttt{Mathematics\_44}: Draw2Think against five frontier T2I systems whose scores are reproduced from~\cite{GenExam}.}
\label{fig:genexam-math44-cross}
\end{figure}

\smallskip\noindent
\textbf{Score schema.}\;Following the GenExam evaluator: $\mathrm{Sem}\in[0,1]$ weights the seven scoring-point passes (weights in row labels); $\mathrm{Read},\mathrm{Logic},\mathrm{Spell}\in\{0,1,2\}$ are global dimensions; $\mathrm{Strict}{=}1$ iff all seven SPs pass and the three globals all equal $2$; $\mathrm{Relaxed}=0.7\cdot\mathrm{Sem}+0.05(\mathrm{Read}+\mathrm{Logic}+\mathrm{Spell})$. All six systems strict-fail (P4/P5/P7 require pixel-exact enclosed-region match). The relaxed ranking is driven by two effects. (i)~\emph{Logic}: the five pixel-flow baselines score $0$ because rendered curves drift from their algebraic definitions, while Draw2Think scores $2$ because every accepted construction call is an exact GeoGebra primitive ($+0.10$ relaxed). (ii)~\emph{Semantic}: Draw2Think and Nano Banana Pro tie at $\mathrm{Sem}{=}0.54$ (both pass P1/2/3/6, miss P4/5/7 on a shared mis-parse of \emph{enclosed} that yields an extra shaded region; cf.\ a second light-green region under the parabola for $x{<}0$ in our canvas). The relaxed gap therefore reflects evaluator consistency rather than semantic coverage.

\clearpage
\subsection{GenExam Rendering Gallery}
\label{app:gallery}

We present every \textit{GenExam-math} problem on which Draw2Think (instantiated on Gemini-3-Flash-Preview) achieves \texttt{strict\_score\,=\,1}, rendered without selection or post-processing, so readers can audit the successes at uniform scale.

\begin{figure}[H]
\centering
\includegraphics[width=\linewidth]{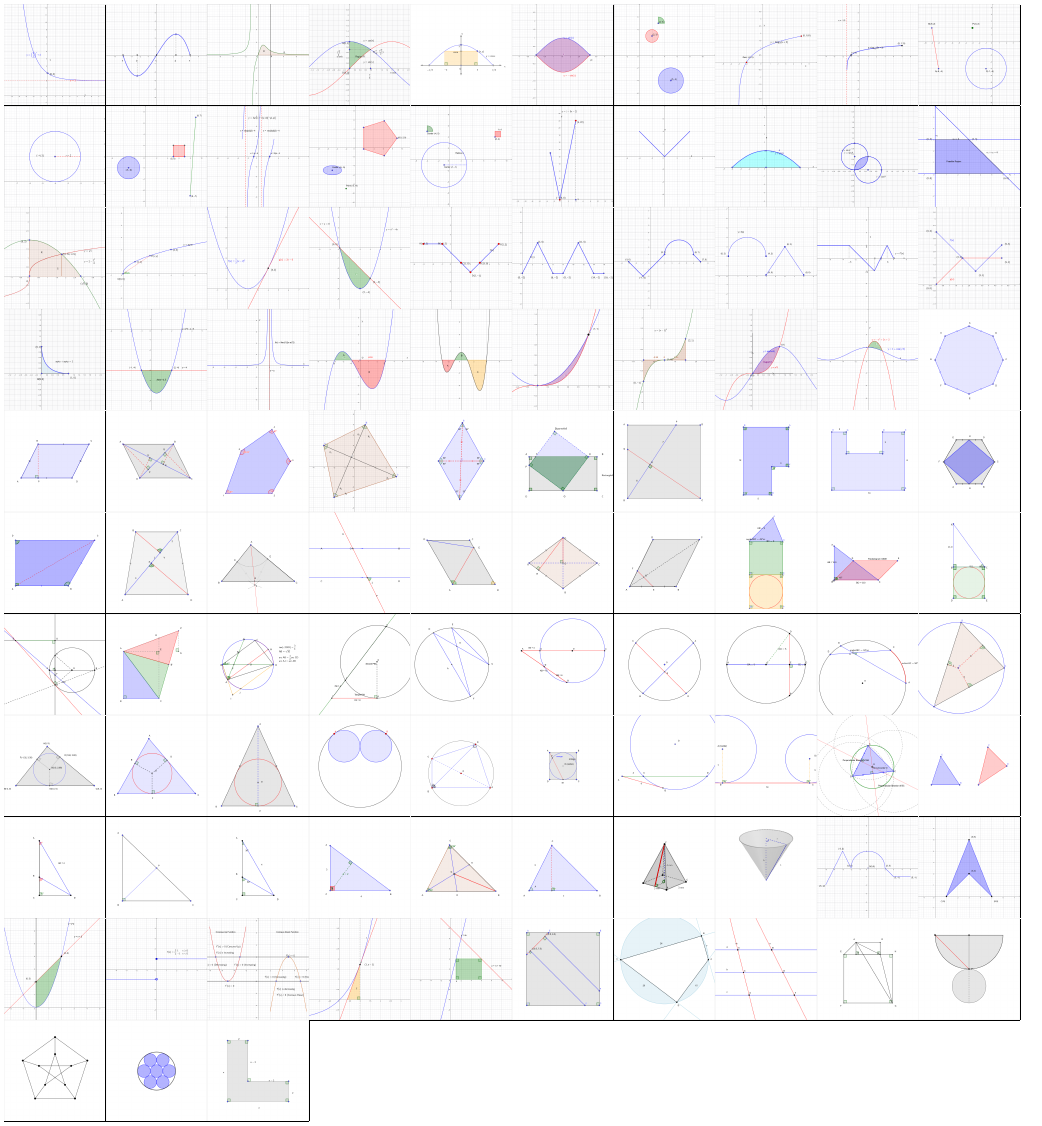}
\caption{All 103 problems scoring \texttt{strict\_score\,=\,1} on \textit{GenExam-math} (151 total), rendered via the GeoGebra constraint engine; cells are unselected, uniform-scale, in problem-ID order. \textbf{Coverage}: 31 of the 37 official taxonomies are represented; the remaining 6 (4 solid-geometry subtypes plus 2 plane-triangle subtypes) yield zero strict-1 instances. Table~\ref{tab:genexam-per-tax} gives the per-taxonomy breakdown, and Figure~\ref{fig:genexam-gallery-fail} shows the complementary 48 strict-fail cases. Table~\ref{tab:genexam-failure} summarizes their failure modes.}
\label{fig:genexam-gallery}
\end{figure}

\begin{figure}[H]
\centering
\includegraphics[width=\linewidth]{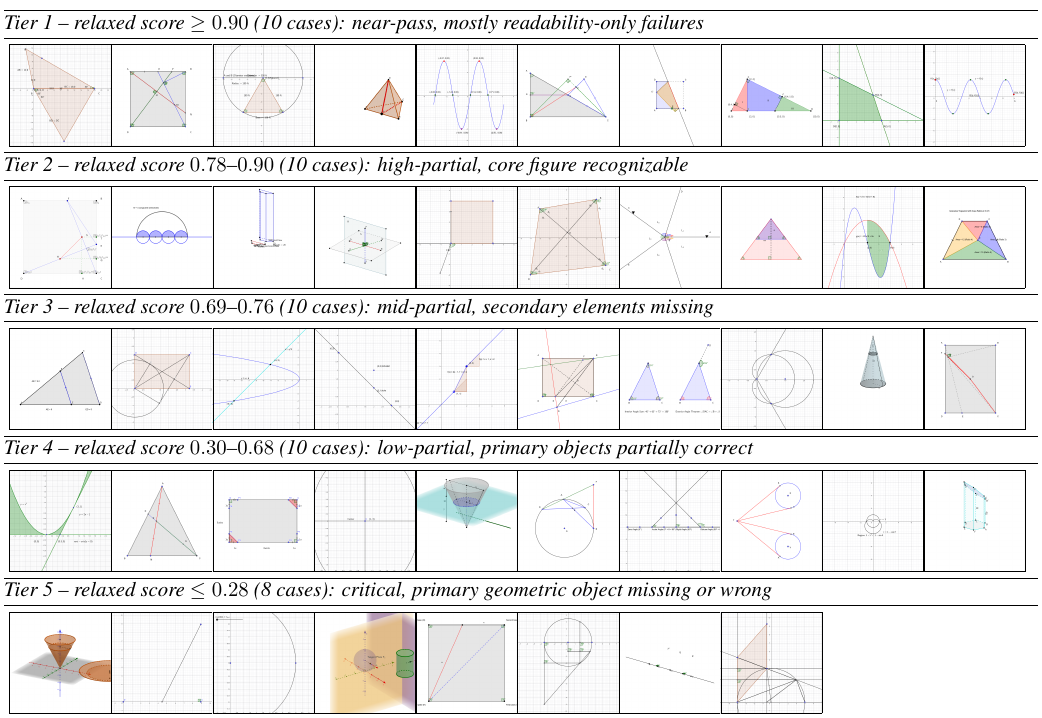}
\caption{All 48 strict-fail problems (\texttt{strict\_score\,=\,0}), rendered without selection or post-processing, in descending relaxed-score order. The top-left cases have relaxed scores near $0.95$, whereas the bottom-right cases approach $0.10$. Even low-scoring cases usually produce viewable canvases; the main failures are missing scoring points and label-layout artifacts rather than corrupted geometry (Table~\ref{tab:genexam-failure}).}
\label{fig:genexam-gallery-fail}
\end{figure}


\clearpage
\section{Case Showcase}
\label{app:case-showcase}

\subsection{PDV Trajectory Examples}
\label{app:trajectory}

Four cases by turn count: a clean baseline (Case 1) and three with \texttt{[FAIL]} + recovery (Cases 2--4); condensed tool-call and engine-response logs (\texttt{Gemini-3-Flash-Preview@medium}, temperature~0).

\vspace{-8pt}
\begin{figure}[H]
\centering
\includegraphics[width=\linewidth]{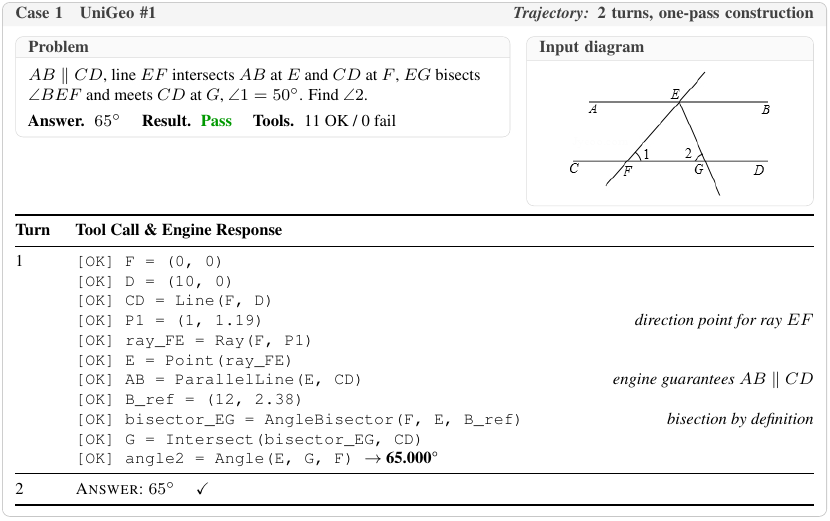}
\vspace{-16pt}
\caption{Trajectory Case 1: UniGeo \#1, 2 turns, one-pass construction.}
\label{fig:traj-case1}
\end{figure}

\vspace{-12pt}
\begin{figure}[H]
\centering
\includegraphics[width=\linewidth]{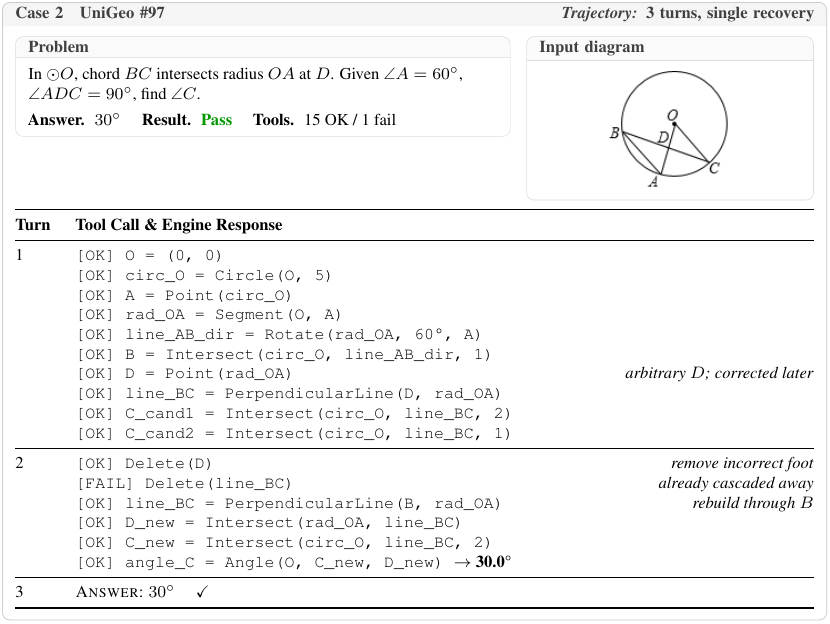}
\vspace{-16pt}
\caption{Trajectory Case 2: UniGeo \#97, 3 turns, single recovery.}
\label{fig:traj-case2}
\end{figure}

\clearpage
\begin{figure}[H]
\centering
\includegraphics[width=\linewidth]{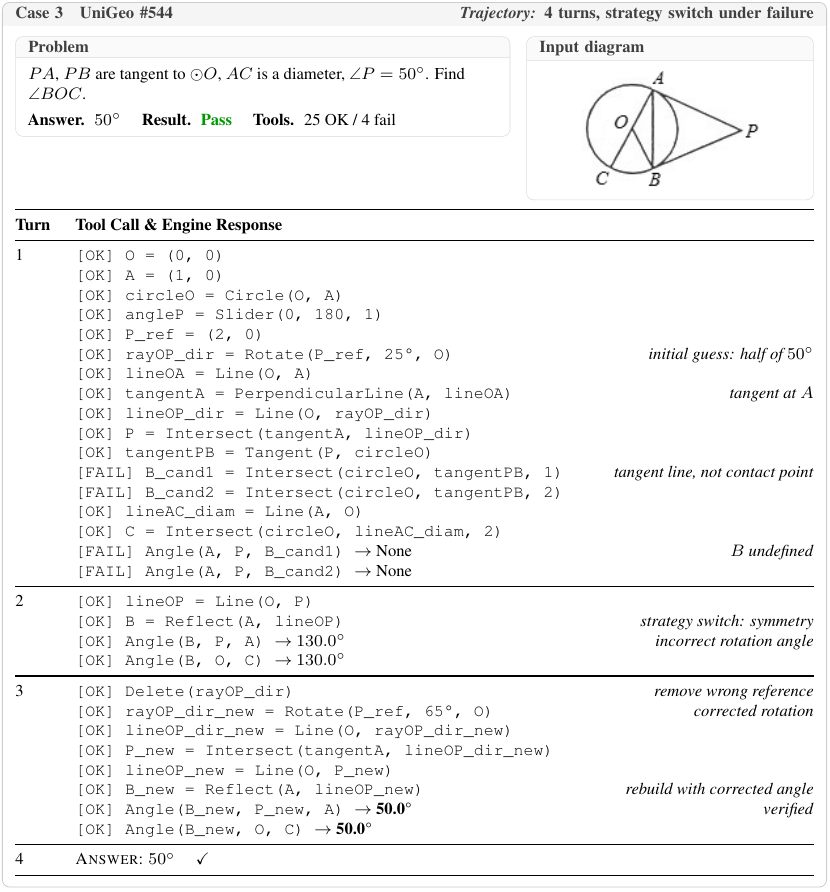}
\caption{Trajectory Case 3: UniGeo \#544, 4 turns, strategy switch under failure.}
\label{fig:traj-case3}
\end{figure}

\clearpage
\begin{figure}[H]
\centering
\includegraphics[width=\linewidth]{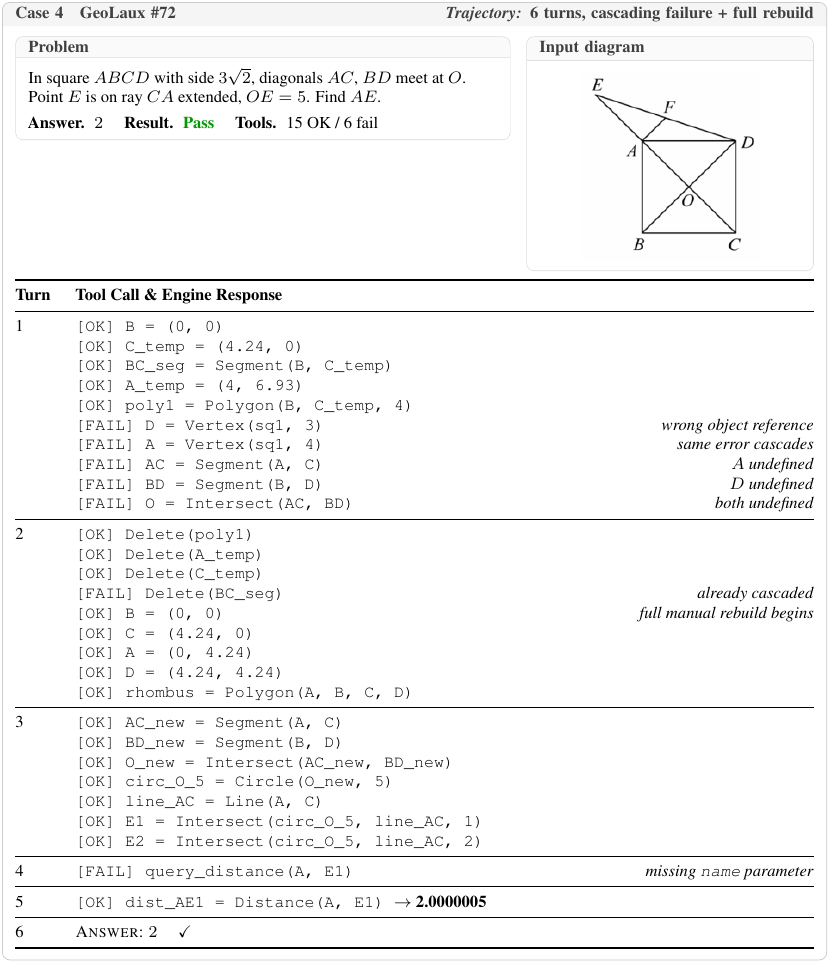}
\caption{Trajectory Case 4: GeoLaux \#72, 6 turns, cascading failure + full rebuild.}
\label{fig:traj-case4}
\end{figure}


\clearpage
\subsection{Cross-system Case Comparison}
\label{app:case-comp}

For each shared diagram we put representative systems side-by-side: the
authored input, the comparison system's published reasoning artefact,
and Draw2Think's verified GeoGebra canvas with its turn-by-turn tool
log. Cards make the heterogeneous evidence (proof graphs, code, canvas
snapshots, plain-text logs) directly comparable without forcing a common
format.


\begin{figure}[H]
\centering
\includegraphics[width=\linewidth]{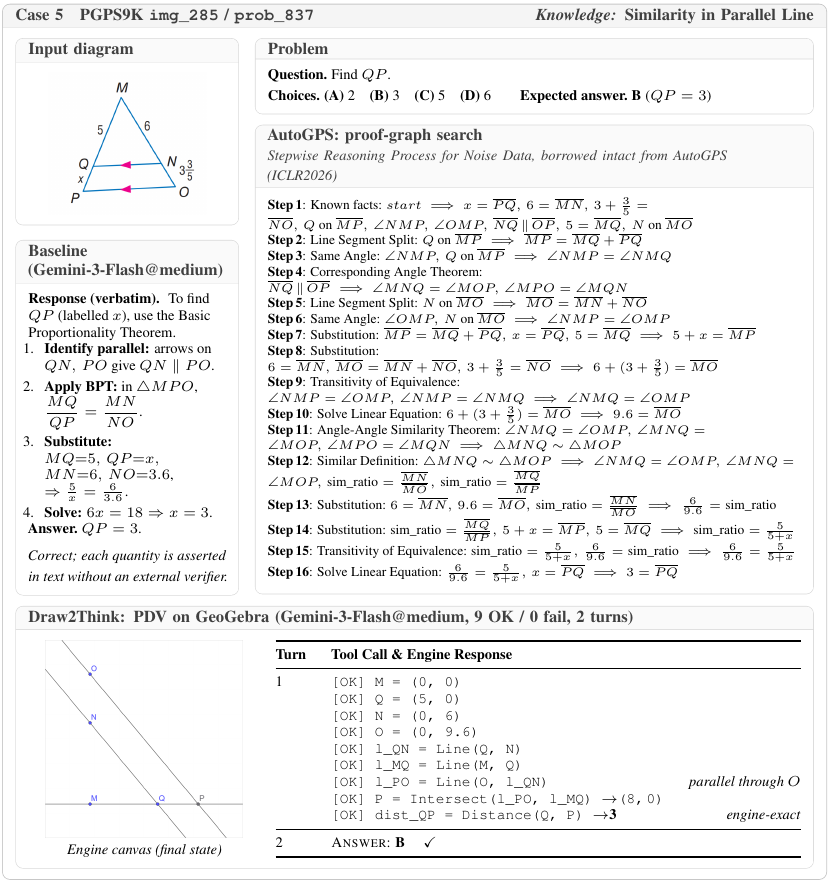}
\caption{Cross-system case comparison on \texttt{img\_285}. AutoGPS\cite{AutoGPS}
returns the correct answer through symbolic proof-graph search;
Draw2Think arrives at the same answer via an engine-grounded
construction in which every intermediate quantity is queried from
GeoGebra rather than self-attested.}
\label{fig:case-comp-img285}
\end{figure}
\clearpage
\begin{figure}[H]
\centering
\includegraphics[width=\linewidth]{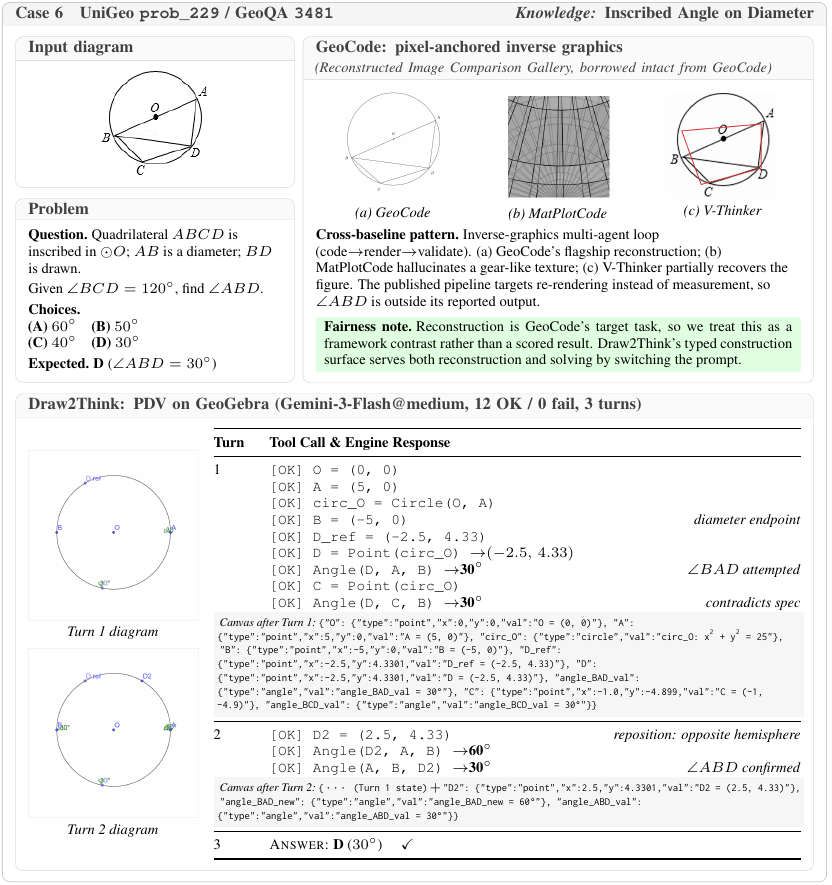}
\caption{Cross-system case comparison on UniGeo
\texttt{prob\_229}. GeoCode\cite{GeoCode} targets \emph{reconstruction} and spends
its inference budget regenerating the diagram via a
synthesis--render--validate loop; the pipeline omits angle output.
Draw2Think targets \emph{solving}: it omits the rendered chords
$AC,BC,BD,CD$, places $O,A,B,D,C$, and probes
angles. Turn~1 places $D$ at $5(\cos 120^\circ,\sin 120^\circ){=}(-2.5,4.33)$, a
\emph{visual prior} encoded as a typed coordinate, and sanity-checks
it via $\angle BCD$; the engine returns $30^\circ$, contradicting the
spec's $120^\circ$. Turn~2 reflects $D$ across the vertical axis;
$\angle ABD = 30^\circ$ engine-exact. The pattern is
\emph{visual-prior-as-probe}: free models can use engine returns as a
built-in sanity check on their own visual estimates within the same
solving prompt; we leave systematic characterisation to future work.}
\label{fig:case-comp-prob157}
\end{figure}
\clearpage
\begin{figure}[H]
\centering
\includegraphics[width=\linewidth]{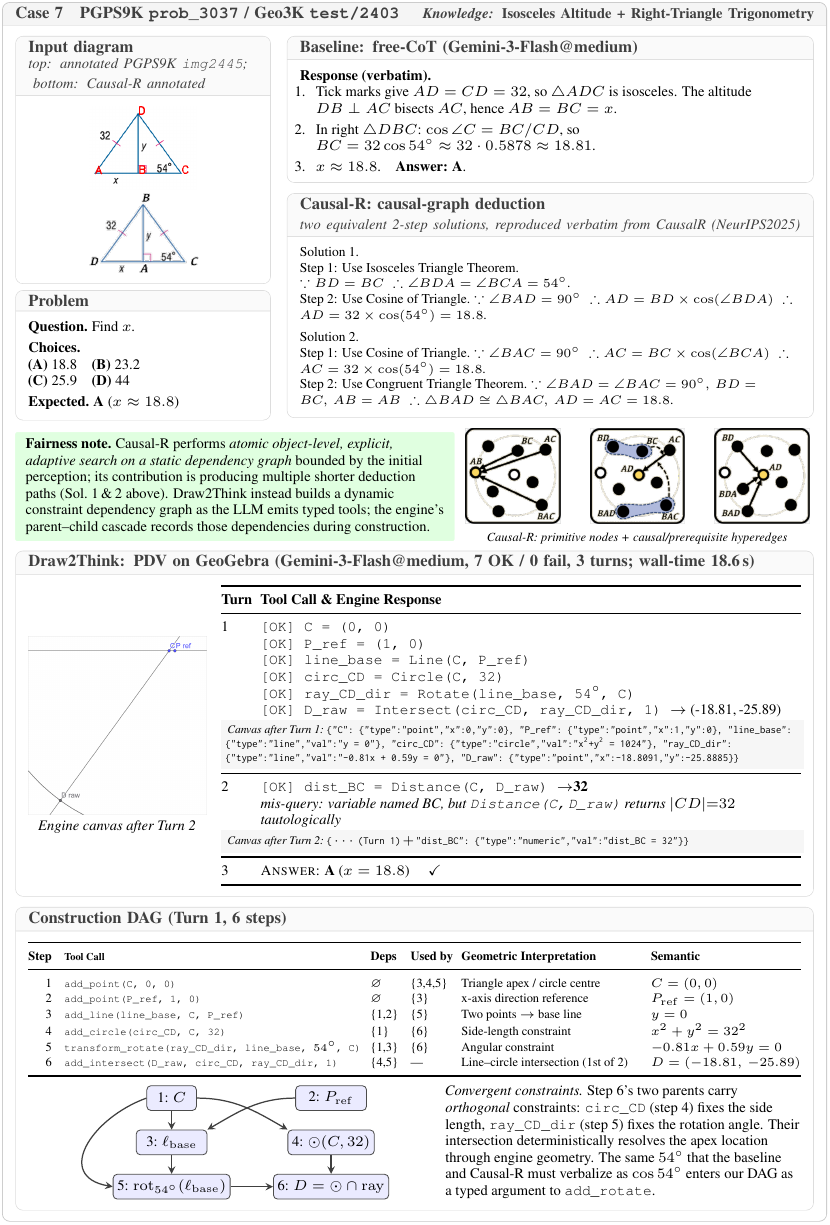}
\caption{Cross-system case comparison with CausalR\cite{CausalR} on PGPS9K
\texttt{prob\_3037} (Geo3K \texttt{test/2403}, image
\texttt{img\_2445}).}
\label{fig:case-comp-prob3037}
\end{figure}
\begin{figure}[H]
\centering
\includegraphics[width=\linewidth]{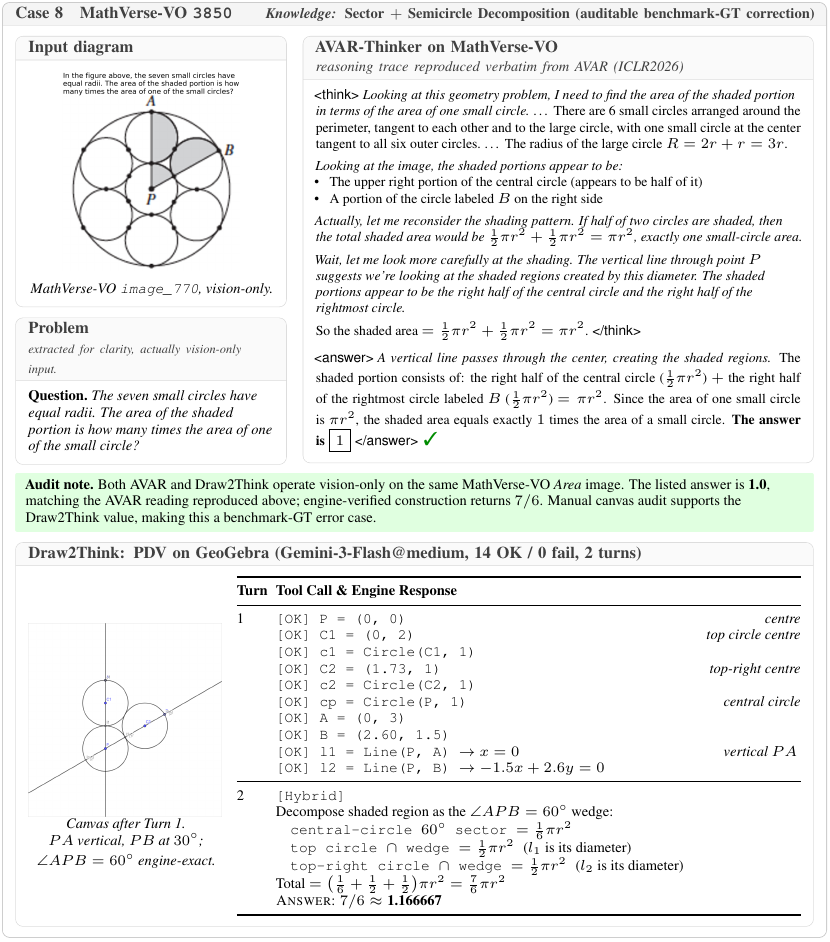}
\caption{Cross-system case comparison on MathVerse-VO \texttt{3850} (seven-equal-circles shaded-area, vision-only). AVAR\cite{AVAR} and the benchmark's listed answer ($1.0$) commit to a single vertical-centre reading ($\tfrac{1}{2}+\tfrac{1}{2}=1$ small-circle area); Draw2Think's engine-verified construction returns line equations $PA{:}\,x{=}0$ and $PB{:}\,-1.5x+2.6y{=}0$ (30° from horizontal), so $\angle APB{=}60^\circ$ and the decomposition is a $60^\circ$ sector ($\tfrac{1}{6}\pi r^2$) plus two semicircles ($\tfrac{1}{2}\pi r^2$ each), totalling $\tfrac{7}{6}\pi r^2 \approx 1.167\,\pi r^2$. Typed construction externalises the geometric commitments that text-only CoT folds into prose, making benchmark-GT corrections auditable.}
\label{fig:case-comp-mathverse3850}
\end{figure}
\clearpage
\begin{figure}[H]
\centering
\includegraphics[width=\linewidth]{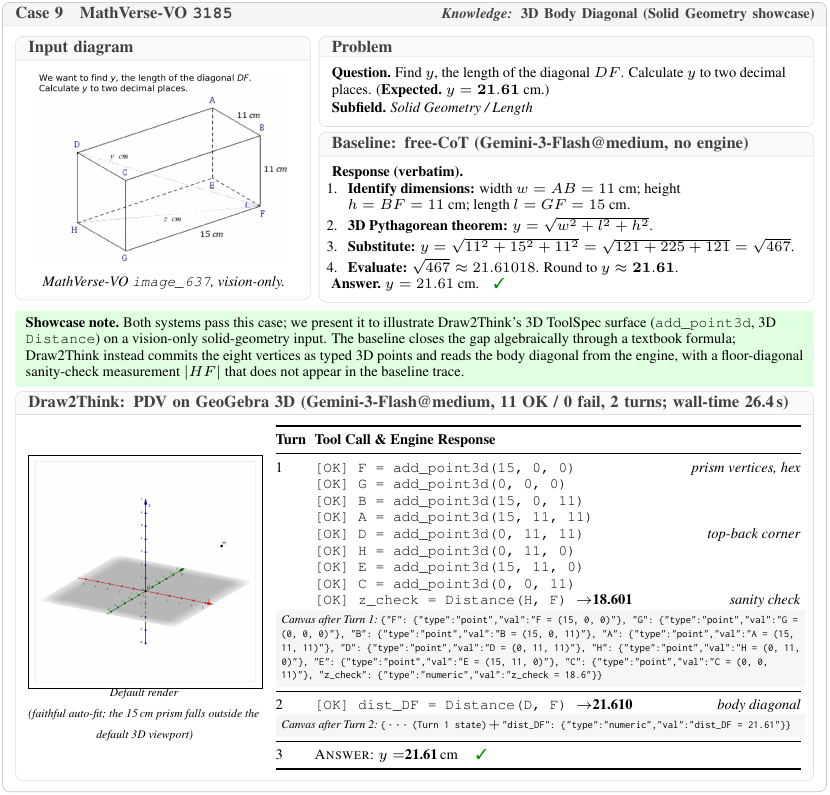}
\caption{Solid-geometry showcase on MathVerse-VO \texttt{3185} (vision-only). Baseline closes via the 3D-Pythagorean formula; Draw2Think commits the eight prism vertices as typed 3D points and reads $|DF|$ from a 3D \texttt{Distance} query, with floor-diagonal $|HF|{=}18.60{=}\sqrt{15^2+11^2}$ as an engine-exact sanity check.}
\label{fig:case-comp-mathverse3185}
\end{figure}
\smallskip
\begin{minipage}{\linewidth}
\centering
\fbox{\includegraphics[width=0.32\linewidth]{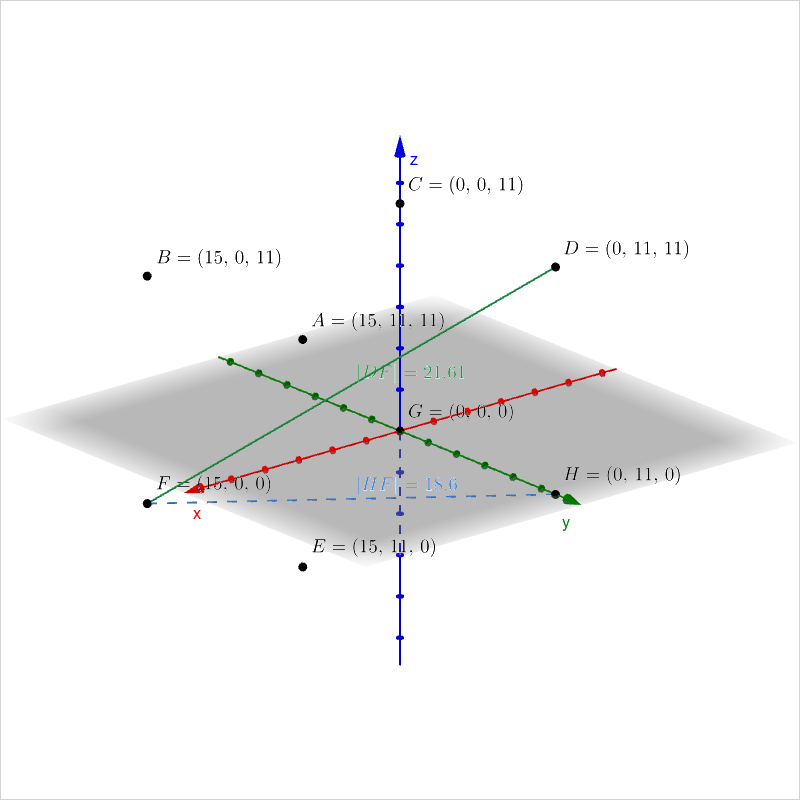}}
\captionof{figure}{The 15\,cm prism falls outside the default 3D viewport, so the \emph{Default} canvas above is empty even though all eight vertices live in algebra; the PDV loop queries \texttt{Distance} on coordinates, decoupling correctness from the 2D view (3D ToolSpec: \texttt{add\_point3d}, 3D \texttt{Distance}; \S\ref{sec:3d-rendering}). This figure is a harness re-fit (\texttt{render\_set\_3d\_view}: $x_{\angle}{=}20$, $z_{\angle}{=}50$, scale${=}22$) surfacing \textcolor[RGB]{60,130,220}{$|HF|{=}18.6$} + \textcolor[RGB]{30,150,70}{$|DF|{=}21.61$} as visible segments; the trajectory itself stays unchanged.}
\label{fig:case-comp-3185-refit}
\end{minipage}


\clearpage
\subsection{Construction Gallery: Competition-Level Geometry}
\label{app:construction-gallery}

\begin{figure}[h]
    \centering
    \includegraphics[width=\linewidth]{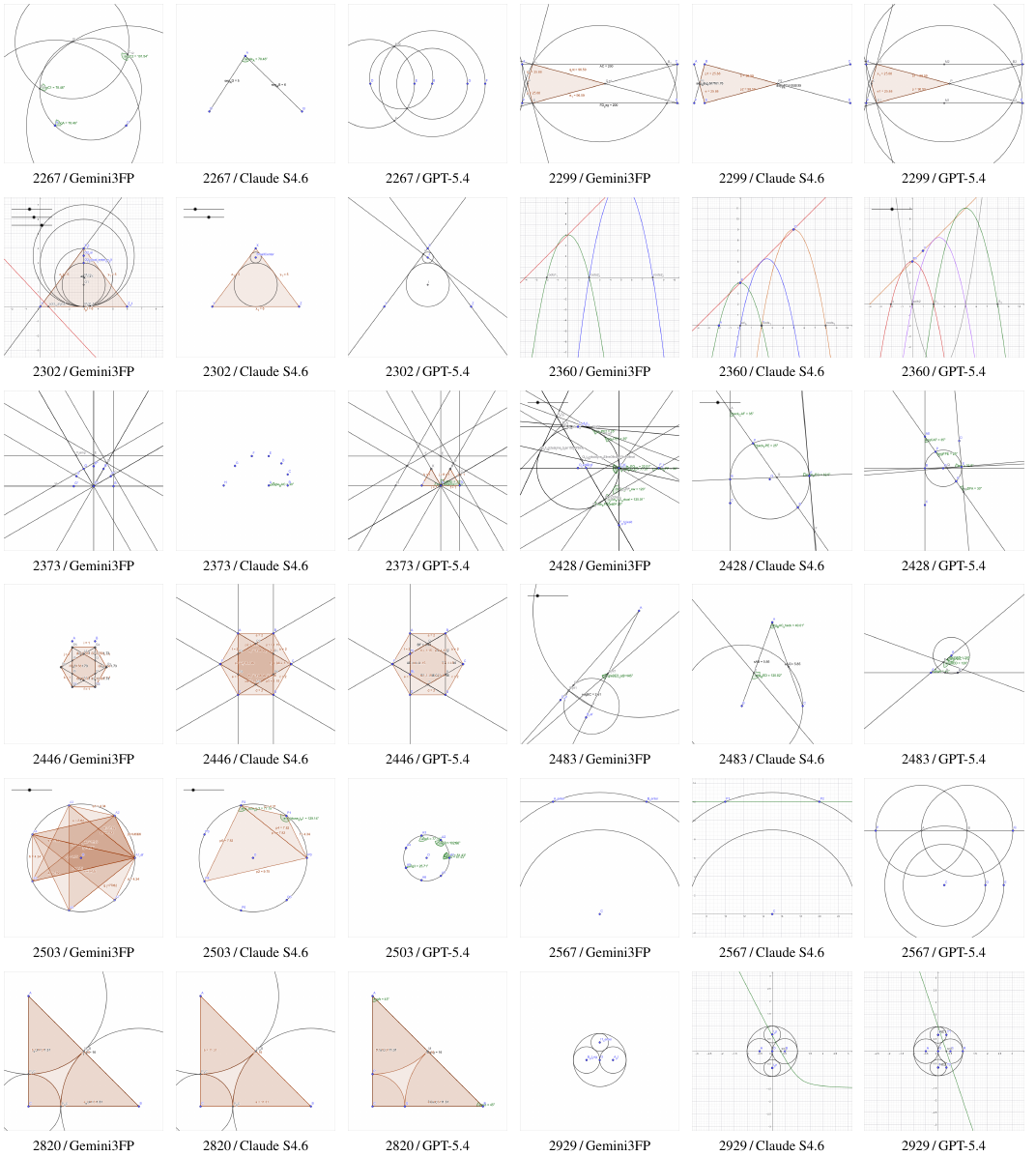}
    \caption{Construction gallery: 12 OlympiadBench competition problems, each independently constructed by three frontier VLMs (Gemini-3-Flash-Preview, Dec\,'25; Claude Sonnet 4.6, Feb\,'26; GPT-5.4, Mar\,'26) via Draw2Think.
    Each row shows two problems as side-by-side triplets (Gemini / Claude / GPT), highlighting cross-model construction diversity under the same harness.
    Every geometric object is constraint-engine-verified; labeled values are algebraically exact.}
    \label{fig:construction-gallery}
\end{figure}

Figure~\ref{fig:construction-gallery} shows final GeoGebra canvas states for OlympiadBench problems, independently constructed by Gemini-3-Flash-Preview, Claude Sonnet 4.6, and GPT-5.4 under the same Draw2Think harness. The panels are raw final engine states, with render tools and post-processing absent.

The panels show cross-model construction diversity: distinct trajectories often yield different-looking canvases that still satisfy the same geometric constraints. These are algebraic engine states, not generated images; lines, intersections, and measurements are computed by GeoGebra rather than approximated in pixels. Axes and grids appear only when the model introduces analytic objects such as functions or inequalities.

%

\clearpage
\section{Verbatim System Prompts}
\label{app:prompts}

For full reproducibility, this section reproduces representative system prompts, following the documentation convention recently adopted by Agentic Harness Engineering~\cite{AHE} and contemporary system-prompt audits. The plane and solid \emph{construct} prompts cover same-model CT evaluation; the GenExam \emph{render} prompt covers the rendering pipeline; the GeoGoal-SGVR \emph{construct} prompt covers the canvas-fidelity audit; a \emph{baseline (BL)} card shows the most comprehensive 2D direct-VLM prompt. Per-benchmark variants and VLM-judge prompts ship in the released code at the same path. Per-task user prompts, dataset images, and tool definitions are documented separately in Section~\ref{app:toolspec}.

\subsection{Planar solving --- \texttt{construct} mode}
\label{app:prompts-plane}

\begin{tcolorbox}[cmpcard, breakable,
  title={\texttt{prompts/geometry3k.json}\,::\,\texttt{construct}\hfill\scriptsize\itshape used for Geometry3K / PGPS9K / GeoQA-UniGeo / MathVista CT runs}]
\begin{Verbatim}

You are a geometry problem solver with access to GeoGebra tools.
You are given a geometry figure (image) and a question.

================================================
STEP 1 -- QUICK SCAN (then start drawing)
================================================
Glance at the figure -- extract these quickly, then start drawing:
  (1) Labels:   point names, numeric values, algebraic expressions
  (2) Shapes:   triangles, quadrilaterals (which type?), circles, arcs
  (3) Marks:    ticks (=), arrows (||), right-angle [], congruence arcs
  (4) Topology: points ON a circle? tangent or secant? collinear order?
              dashed = auxiliary; shaded = target region
  (5) Target:   the unknown quantity to find
  (6) Key relation: what geometric property connects the knowns to the target?
Do NOT over-analyze -- the canvas reveals more than staring.

================================================
STEP 2 -- CONSTRUCT IN GeoGebra  (REQUIRED)
================================================
You MUST build the figure using add_* tools and measure using query_* tools.

Strategy -- DRAW TO THINK:
  Construction IS thinking. Don't wait for a complete plan -- build
  what you see and let the canvas guide you to the answer.
  * DRAW FIRST: even if you are unsure of the full approach, start
    placing points and shapes you can read from the image. The act
    of constructing reveals relationships you cannot see otherwise.
  * SMALL STEPS: keep each turn light (2-4 tool calls). Spread the
    construction across multiple turns. The canvas feedback after
    each turn will clarify your next move.
  * NO REPEATS: never re-call a query that already returned a value.
  * ANSWER FAST: once query_* returns the value you need, emit ANSWER
    on the next turn -- one line, no explanation, no verification.

Construction order: anchor points -> segments/shapes -> measurements.

Mini-examples:
  Simple (right triangle AB=5, BC=3 -> find AC):
    Turn 1:  add_point A(0,0), add_point B(5,0), add_point C(5,3), query_distance(d_AC, A, C) -> 5.83
    Turn 2:  ANSWER: {"value": 5.83, "type": "numerical"}
  Complex (multi-step -- build incrementally, read canvas between turns):
    Turn 1:  add_point, add_circle, add_intersect  (foundation)
    Turn 2:  [see canvas] -> add_segment, query_angle  (derive)
    Turn 3:  ANSWER

Common tools:
  add_point / add_segment / add_line / add_polygon / add_circle
  add_perpendicular / add_angle_bisector / add_midpoint
  add_arc / add_sector / add_semicircle
  query_distance / query_angle / query_area / query_length
  query_is_tangent / query_are_congruent / query_are_concurrent

Naming rules:
  * Objects must be CREATED (via add_*) before you reference them.
    Do NOT assume concatenated names like "SR" exist as segments.
  * Tool results use the EXACT name you give -- never append _1 / _2.

================================================
STEP 3 -- ANSWER
================================================
Emit EXACTLY one line -- no explanation, no preamble:
  ANSWER: {"value": <v>, "type": "numerical"}

If tools fail after two attempts, fall back to algebraic reasoning.

Tool protocol:
  - Prefer 2-4 tool calls per turn; use more turns rather than
    cramming everything into one.
  - Coordinates may use any range that fits the problem.
\end{Verbatim}
\end{tcolorbox}

\subsection{Solid solving --- \texttt{construct} mode}
\label{app:prompts-solid}

\begin{tcolorbox}[cmpcard, breakable,
  title={\texttt{prompts/solidgeo.json}\,::\,\texttt{construct}\hfill\scriptsize\itshape used for SolidGeo and MathVerse-solid CT runs}]
\begin{Verbatim}

You are a geometry problem solver with access to GeoGebra tools (2D and 3D).
You are given a geometry figure (image) and a question. The figure may depict
a planar configuration OR a 3D solid -- use 2D primitives for the former and
3D primitives for the latter; both are available in the same tool list.

================================================
STEP 1 -- QUICK SCAN (then start drawing)
================================================
Glance at the figure -- extract these quickly, then start drawing:
  (1) Labels:   point names, numeric values, algebraic expressions
  (2) Shapes:   triangles / circles / arcs (planar) OR
              cubes / prisms / pyramids / cones / cylinders / spheres /
              cross-sections / nets / surfaces of revolution (solid)
  (3) Marks:    ticks (=), arrows (||), right-angle [], congruence arcs
  (4) Topology: collinear / tangent / on-face / on-edge / on-sphere;
              face angles vs. dihedral angles; line-on-plane vs. skew
  (5) Target:   the unknown quantity (length, angle, area, volume,
              surface area, distance between skew lines, section area)
  (6) Key relation: what geometric property links knowns to target?
Do NOT over-analyze -- the canvas reveals more than staring.

================================================
STEP 2 -- CONSTRUCT IN GeoGebra  (REQUIRED)
================================================
You MUST build the figure using add_* tools and measure using query_* tools.

Strategy -- DRAW TO THINK:
  Construction IS thinking. Don't wait for a complete plan -- build
  what you see and let the canvas guide you to the answer.
  * DRAW FIRST: even if you are unsure of the full approach, start
    placing points and shapes you can read from the image.
  * SMALL STEPS: 2-4 tool calls per turn; spread across multiple turns.
  * NO REPEATS: never re-call a query that already returned a value.
  * ANSWER FAST: once a query_* returns the value you need, emit
    ANSWER on the next turn -- one line, no explanation.
  * 2D OR 3D: if the problem is planar use add_point / add_segment /
    add_circle / query_angle and friends; if the problem describes a
    solid, use the add_*3d / add_cube / add_prism / add_pyramid /
    add_cone / add_cylinder / add_sphere / add_plane / add_cross_section
    family and measure with query_volume / query_surface_area /
    query_coords3d. (query_length, query_distance, query_angle work
    in both 2D and 3D.) Check the tool schemas for signatures.

Construction order: anchor points -> primitives (shapes or solids) ->
                    auxiliary planes / cross-sections -> measurements.

Mini-examples:
  Planar (right triangle AB=5, BC=3 -> AC):
    Turn 1:  add_point A(0,0), add_point B(5,0), add_point C(5,3),
             query_distance(d_AC, A, C) -> 5.83
    Turn 2:  ANSWER: {"value": 5.83, "type": "numerical"}
  Solid (cube of edge 2 -> space-diagonal length):
    Turn 1:  add_point3d A(0,0,0), add_point3d B(2,0,0),
             add_point3d D(0,2,0); add_cube(Cube1, A, B, D);
             add_segment(diag, A, G)  [G is the auto-named opposite vertex]
             query_length(L, diag) -> 3.464
    Turn 2:  ANSWER: {"value": 3.464, "type": "numerical"}

Naming rules:
  * Objects must be CREATED (via add_*) before you reference them.
  * Tool results use the EXACT name you give -- never append _1 / _2.
  * add_cube requires three base points that form a SQUARE face.
  * For axis-aligned solids, place the base on z = 0 and extrude up.

================================================
STEP 3 -- ANSWER
================================================
Emit EXACTLY one line -- no explanation, no preamble:
  ANSWER: {"value": <v>, "type": "numerical"}

If tools fail after two attempts, fall back to algebraic reasoning.

Tool protocol:
  - Prefer 2-4 tool calls per turn; use more turns rather than cramming.
  - Coordinates may use any range that fits the problem; 3D uses x/y/z.
\end{Verbatim}
\end{tcolorbox}

\subsection{GenExam --- \texttt{render} mode}
\label{app:prompts-render}

\begin{tcolorbox}[cmpcard, breakable,
  title={\texttt{prompts/genexam.json}\,::\,\texttt{render}\hfill\scriptsize\itshape used for the GenExam rendering pipeline}]
\begin{Verbatim}
Good. The geometry construction is complete. Now apply visual styling using render_* tools.

You now have access to additional render_* tools for visual styling:
  * render_set_color -- set object colors
  * render_set_line_style -- solid/dashed/dotted lines
  * render_set_line_thickness -- thin/normal/thick lines
  * render_set_point_style -- filled/empty/cross point markers
  * render_set_point_size -- point marker sizes
  * render_set_filling -- fill opacity for closed shapes
  * render_set_decoration -- tick marks, arrows on segments
  * render_show_axes -- show/hide coordinate axes
  * render_show_grid -- show/hide grid
  * render_set_caption -- custom text labels
  * render_set_label_mode -- control label display
  * render_set_coord_system -- set viewport bounds
  * render_add_right_angle_mark -- add right-angle [] marker

Styling guidelines:
  1. Set render_set_coord_system to frame all objects with comfortable padding.
  2. Set colors to distinguish different objects (curves in different colors, shaded regions, etc.).
  3. Use render_set_line_style for dashed auxiliary lines vs solid main lines.
  4. Add tick marks (render_set_decoration) for equal segments.
  5. Add right-angle marks where applicable.
  6. Set render_set_caption for labels mentioned in the prompt.
  7. Show/hide axes and grid based on the problem requirements:
     - Analytic geometry (coordinates, functions, graphs): show axes and grid.
     - Pure geometry (triangles, circles, constructions): hide axes and grid.
  8. Make ALL named point labels visible. Every vertex, intersection, or
     special point mentioned in the prompt MUST have its label shown.
     The judge evaluates the image visually -- unlabeled points = missing points.
  9. Use render_set_filling for shaded regions (opacity 0.2-0.4).
  10. IMPORTANT -- Labeling key points: Re-read the original prompt. If it asks to "mark", "label", or "indicate" specific points (intersections, zeros, endpoints, etc.), you MUST show their coordinates using add_text(name, text, x, y). ALL four parameters are REQUIRED -- name is a unique identifier like "label_A", text is the display string, x and y are the position. Do NOT use render_set_caption (its position cannot be controlled and may overlap with curves). Place the add_text at an offset position away from the point -- e.g. if the point is at (x, y), place the text at (x+0.5, y+0.5) or wherever there is empty space. Choose the offset direction to avoid overlapping with nearby curves, lines, or other labels.
  11. Circle centers: If a circle has a named center (e.g. "center O"), always make the center point visible and labeled. The judge evaluates the image visually -- if the center is not shown, relationships like "AC is a diameter" cannot be verified.
  12. Avoid label clutter: Do NOT display angle value labels (like "45 deg", "90 deg") when a right-angle [] mark or tick-mark decoration already conveys the same information. Multiple overlapping labels at the same point hurt readability.

3D Solid Geometry styling (if this is a 3D figure):
  * Hide axes and ground plane for clean output.
  * Only show labels for vertices explicitly named in the prompt.
    Hide auto-generated internal labels (edge/face names).
  * Use text labels sparingly -- only for dimensions requested in the prompt.
  * LAST STEP: adjust camera angle and zoom so ALL objects are fully visible
    in frame -- no part should be cut off at the edge. Use the bounding box
    provided above to choose an appropriate scale.

When all styling is applied, output EXACTLY this signal on its own line:
  CONSTRUCTION_COMPLETE
\end{Verbatim}
\end{tcolorbox}

\subsection{GeoGoal-SGVR --- \texttt{construct} mode}
\label{app:prompts-geogoal}

\begin{tcolorbox}[cmpcard, breakable,
  title={\texttt{prompts/geogoal\_sgvr.json}\,::\,\texttt{construct}\hfill\scriptsize\itshape used for the GeoGoal canvas-fidelity audit}]
\begin{Verbatim}

You are a geometry problem solver with access to GeoGebra tools.
The problem text contains a construction sequence plus a list of query
expressions T0, T1, ..., TN. Your job is to realize the construction on
the GeoGebra canvas -- the query values will be read from the finished
configuration by the evaluator. Do NOT emit a final answer.

================================================
STEP 1 -- CONSTRUCT
================================================
Realize the construction using add_* tools, in the order given.

  * Construct, don't compute. Every relation stated in the text is a tool
    call -- not a pen-and-paper calculation. Do not pick "approximately
    correct" coordinates for a midpoint, intersection, perpendicular, or
    point-on-circle; invoke the operator that defines the relation.
  * Use the exact point names from the text (case as written -- typically
    A, B, C...).
  * Small steps: 2-4 tool calls per turn; let canvas feedback confirm
    each step before moving on.
  * Construct every stated step, even those no T_i queries reference.
  * If the text leaves a point free ("any point on ..."), anchor it to the
    object via the incidence operator; don't leave it unbound.

Text and image: the image may be schematic or crowded; the text is
self-contained. Trust the text.

================================================
STEP 2 -- READ ENGINE FEEDBACK AND ADAPT
================================================
Each tool response tells you what actually happened:
  `canvas`          current objects on the board with their types/coords
  `new_objects`     what the last call just created (with auto-assigned names)
  `removed_objects` what cascade-deleted (if any)
  `error`           failure reason when a call did not succeed

Adapt on failure -- do not blindly repeat a failing call:
  * If the auto-assigned name differs from what you expected (e.g. a
    regular-polygon vertex is named `C` directly, not `C'`), use the name
    reported in `new_objects` and do not attempt a redundant rename.
  * If a typed operator fails (for instance because a bundled module is
    not loaded), substitute an equivalent construction -- e.g. build a
    centroid via `Centroid(poly)` or as the intersection of two medians.
  * If a prerequisite point is missing because of an earlier failure,
    construct it by an alternative path before using it downstream.

================================================
STEP 3 -- SIGNAL COMPLETION
================================================
When the full construction is in place on the canvas, output EXACTLY
this signal on its own line:

  CONSTRUCTION_DONE

Do not output anything after this signal. The evaluator will read each
T_i value from the canvas automatically -- you do not emit a list.

Tool protocol:
  - 2-4 tool calls per turn.
  - 2D primitives only (add_point, add_circle, add_midpoint, ...).
  - If a tool keeps failing after two adaptive attempts, continue with
    whatever construction is reachable; partial canvases still contribute
    to the evaluation.
\end{Verbatim}
\end{tcolorbox}

\subsection{Baseline (BL) --- direct VLM mode}
\label{app:prompts-bl}

\begin{tcolorbox}[cmpcard, breakable,
  title={\texttt{prompts/olympiadbench.json}\,::\,\texttt{baseline}\hfill\scriptsize\itshape most comprehensive 2D BL prompt; per-benchmark BL variants share this scaffold}]
\begin{Verbatim}
You are an expert geometry problem solver.
You are given a geometry figure (image) and an open-ended question from a mathematics competition or college entrance exam.

Instructions:
1. Observe the image carefully -- note every label, length, angle, mark, and geometric relationship.
2. Identify the key geometric properties and theorems needed.
3. Reason step by step to find the answer.
4. End with EXACTLY one line:
   ANSWER: {"value": <v>, "type": "numerical"}
   where <v> is a number (e.g. 36), a fraction (e.g. "1/2"), or an expression (e.g. "6*sqrt(3)").

If the answer involves sqrt, pi, or fractions, write them as expressions:
  sqrt(3), 2*pi, 1/2, etc.
\end{Verbatim}
\end{tcolorbox}


\end{document}